\documentclass[12pt]{iopart}
\usepackage{booktabs}
\usepackage{comment,mfirstuc}
\usepackage{graphicx}% Include figure files
\usepackage{grffile}
\usepackage{algorithm}% http://ctan.org/pkg/algorithm
\usepackage{algpseudocode}% http://ctan.org/pkg/algorithmicx
\usepackage{hyperref}% add hypertext capabilities
\usepackage{soul}
\usepackage{listings}
\usepackage{multirow}
\usepackage{dcolumn}% Align table columns on decimal point
\usepackage{bm}% bold math
\usepackage{placeins}
\expandafter\let\csname equation*\endcsname\relax
\expandafter\let\csname endequation*\endcsname\relax
\usepackage{amsmath}
\usepackage{amssymb}
\usepackage{mathtools}
\usepackage[bottom=1.0in]{geometry}
\usepackage{cite}
\usepackage{lineno}
\RequirePackage{xcolor}
\usepackage[misc]{ifsym}
\usepackage{lipsum} 
\usepackage{subcaption}
\usepackage{tcolorbox} % Create coloured boxes (e.g. the one for the key-words)
\usepackage{stfloats} % Correct position of the tables
\usepackage{siunitx}
\DeclareSIUnit\eVperc{\eV\per\clight}
\DeclareSIUnit\clight{\text{\ensuremath{c}}}

\usepackage{ulem}

\usepackage{xspace}
\usepackage{colortbl}
\usepackage{array}
\newcolumntype{P}[1]{>{\centering\arraybackslash}p{#1}}
\newcolumntype{M}[1]{>{\centering\arraybackslash}m{#1}}
\hypersetup{ 
    pdfnewwindow=true,      % links in new window
    colorlinks=true,       % false: boxed links; true: colored links
    linkcolor=blue,         % color of internal links
    citecolor=blue,        % color of links to bibliography
    filecolor=blue,      % color of file links
    urlcolor=blue        % color of external links
}  

\algblock{Input}{EndInput}
\algnotext{EndInput}
\algblock{Output}{EndOutput}
\algnotext{EndOutput}

\def\be{\begin{eqnarray} &&} 
 
\def\ee{\end{eqnarray}}

\makeatletter
\newcommand{\mainmatter}{%
  \setcounter{footnote}{0}%
  \patchcmd{\@makefntext}{\fnsymbol}{\arabic}{}{}%
  \patchcmd{\@thefnmark}{\fnsymbol}{\arabic}{}{}%
  \def\@makefnmark{\textsuperscript{\arabic{footnote}}}
  \long\def\@makefntext##1{\parindent 1em\noindent
        \hb@xt@1.8em{%
            \hss\@textsuperscript{\normalfont\@thefnmark}}##1}%
}
\makeatother

\newcommand{\addComment}[2]{
  \expandafter\newcommand\csname #1\endcsname[1]{{\bf \color{#2} \capitalisewords{#1}:\,##1}}
  \expandafter\newcommand\csname #1cor\endcsname[2]{{\color{#2} \capitalisewords{#1}:\,\st{##1}{\bf ##2}}}
  \expandafter\newcommand\csname #1color\endcsname{#2}
}
\addComment{cris}{blue} 
\addComment{james}{red}
\addComment{mike}{magenta}

\newcommand{\gluex}{\textsc{GlueX}\xspace} 
\newcommand{\geant}{\textsc{Geant4}\xspace}

\begin{document}

% Keywords command
\providecommand{\keywords}[1]
{
  \small
  \textbf{Keywords:}  {\color{blue}#1
  }
}

%\title[\scriptsize{Fast Simulation for the hpDIRC at the Electron Ion Collider}]{Fast Simulation for the hpDIRC at the Electron Ion Collider} 

%\title[\scriptsize{Towards Unified Spatiotemporal Understanding in Pixelated Readout Systems with Foundation Models}]{Towards Unified Spatiotemporal Understanding in Pixelated Readout Systems with Foundation Models} 

\title[\scriptsize{Towards Foundation Models for Experimental Readout Systems Combining Discrete and Continuous Data}]{Towards Foundation Models for Experimental Readout Systems Combining Discrete and Continuous Data} 

\author{J. Giroux$^{1,\star}$, C. Fanelli$^{1,\star}$} 

\address{
$^{1}$ William \& Mary, Department of Data Science, Williamsburg, VA 23185, USA\\
$^{\star}$ Author to whom any correspondence should be addressed.
}

\ead{{\color{blue}
jgiroux@wm.edu,
cfanelli@wm.edu
}}

\vspace{10pt}
\begin{indented}
\item[]\today
\end{indented}

%\linenumbers
\begin{abstract}

We present a (proto) Foundation Model for Nuclear Physics, capable of operating on low-level detector inputs from Imaging Cherenkov Detectors at the future Electron Ion Collider. Building upon established next-token prediction approaches, we aim to address potential challenges such as resolution loss from existing tokenization schemes and limited support for conditional generation. We propose four key innovations: (i) separate vocabularies for discrete and continuous variates, combined via Causal Multi-Head Cross-Attention (CMHCA), (ii) continuous kinematic conditioning through prepended context embeddings, (iii) scalable and simple, high-resolution continuous variate tokenization without joint vocabulary inflation, and (iv) class conditional generation through a Mixture of Experts. Our model enables fast, high-fidelity generation of pixel and time sequences for Cherenkov photons, validated through closure tests in the High Performance DIRC. We also show our model generalizes to reconstruction tasks such as pion/kaon identification, and noise filtering, in which we show its ability to leverage fine-tuning under specific objectives.

\end{abstract}


%\maketitle

\mainmatter

\section{Introduction}\label{sec:intro}

Foundation Models (FM) are beginning to emerge as powerful tools within the field of both Nuclear Physics (NP) and High Energy Physics (HEP), supporting the inclusion of multiple downstream tasks \cite{finke2023learning,birk2024omnijet,mikuni2024omnilearn,bardhan2025hepjepa,leigh2025tokenization} and fine-tuning \cite{vigl2024finetuning,Golling_2024,wildridge2024bumblebee,harris2024re}.\footnote{In \cite{birk2024omnijet}, the authors coin the term proto-Foundation Models, representing model development along the path to true, large-scale Foundation Models.} Seminal works, such as those in \cite{birk2024omnijet,mikuni2024omnilearn}, show that relatively modest sized transformer backbones are easily generalizable to both reconstruction (particle identification (PID), jet tagging, etc) and simulation tasks. In which fine tuning from one task to another possess inherent reduction in computational overhead in comparison to models trained from scratch. Moreover, they show that different underlying mechanisms, \textit{i.e.,} next token prediction, masked prediction (akin to traditional decoder only, or encoder-decoder style language models) \cite{finke2023learning,huang2024language,Golling_2024,birk2024omnijet,Butter2025} or iterative refinement through Diffusion Transformers \cite{peebles2023scalable} can be equally as generalizable. 
In which the latter method \cite{mikuni2024omnilearn} promotes reconstruction tasks through selective time restrictions in the diffusion processes (\textit{i.e.,} $t=0$).
In general, both of these approaches work with relatively high-level reconstructed quantities such as 4-vectors. Recently, Birk et al.~\cite{birk2025omnijet} demonstrated that next-token prediction can be effectively applied to simulate calorimeter response, particularly for modeling photon-induced shower development using low-level features such as pixel positions and energy depositions. While the results are promising and represent an important step forward, there are two potential limitations in the current approach that warrant further consideration. First, given the requirements of next token prediction to (generally) operate in a discrete space, quantization methods must be employed upstream in terms of a learnable tokenization process, generally in the form of a Vector Quantized-Variational AutoEncoder (VQ-VAE) \cite{van2017neural}. VQ-VAE models provide discrete latent encodings, therefore allowing tokenization of continuous and multidimensional spaces such as location (pixel) and energy in calorimeters. While powerful, such methodologies do not come without significant tradeoffs in terms of resolutions - a potential degrading effect in physics simulations of detectors through loss of information.\footnote{Decoded tokens inherently posses some uncontrolled smearing effect proportional to the codebook size.} Second, their approach does not encompass conditional generation, and therefore generative tasks that exhibit explicit conditional dependence on external parameters (\textit{e.g.}, Cherenkov production in DIRC detectors) is not possible. As a result, we propose two improvements for next token prediction models to circumvent these potential issues in physics, in which we deploy our techniques to Imaging Cherenkov Detectors at the future Electron Ion Collider (EIC) - pixelated detectors that require both precise location, and timing information of individual hits.\footnote{We leave translation to calorimeters (see, \textit{e.g.}, \cite{birk2025omnijet}) for future works.} 
The main contributions of this work are as follows:
\begin{itemize}
    \item We introduce a (proto) Foundation Model in Nuclear Physics, operating on low level detector inputs. Its performance is demonstrated through a series of closure tests using Cherenkov detectors at the future EIC.
    \item We introduce learning through split vocabularies in space and time, in which discretization (tokenization) in the continuous time space can be done at a fraction of the resolution of the detector. These modalities are combined through Causal Multi-Head Cross-Attention (CMHCA) \cite{lin2022cat,fei2024video} to produce spatio-temporally aware embeddings, in which we query the pixel space given times.
    \item We introduce continuous kinematic dependencies through fixed context, in which prepending of kinematic embeddings drives sequence generation forward in time. Providing kinematic aware next token prediction.
    \item We introduce class conditional generation through a Mixture of Experts (MoE), removing the need for class-dependent generative models as seen in previous works \cite{fanelli2024deep,giroux2025generativemodelsfastsimulation}.
\end{itemize}

The use of split vocabularies allows our model to retain high resolution time generations, without the need for excessive vocabulary size or learnable tokenization. In fact, the resolutions capable of being deployed in our study would comprise a joint-vocabulary (space and time) of roughly 36 million tokens (5920 possible time values, 6144 possible pixel values). We also note that our approximation through split vocabularies remains valid under most pixelated detector systems, given the one-to-many mappings that occur between space and the continuous variate such as time, energy or both \cite{ameli2022streaming}. In this scenario, one could envision the combination of space, time, and energy through sequential CMHCA blocks, where, \textit{e.g.}, time queries energy to generate a temporal-energy informed embedding, which is then used to query spatial indices.

In what follows, we demonstrate the ability of our methods to produce high fidelity fast simulation of pions and kaons in the High Performance DIRC (hpDIRC) through a series of closure tests, in which we compare generations from class-wise independent models with those stemming from a singular model with multiple experts. We also show the model's ability to operate under a classification scheme, both at the sequence (PID) and token (noise filtering) levels, with and without fine tuning.

\section{Dataset}\label{sec:data}

\begin{figure}[h]
    \centering 
    \includegraphics[width=0.5\textwidth]{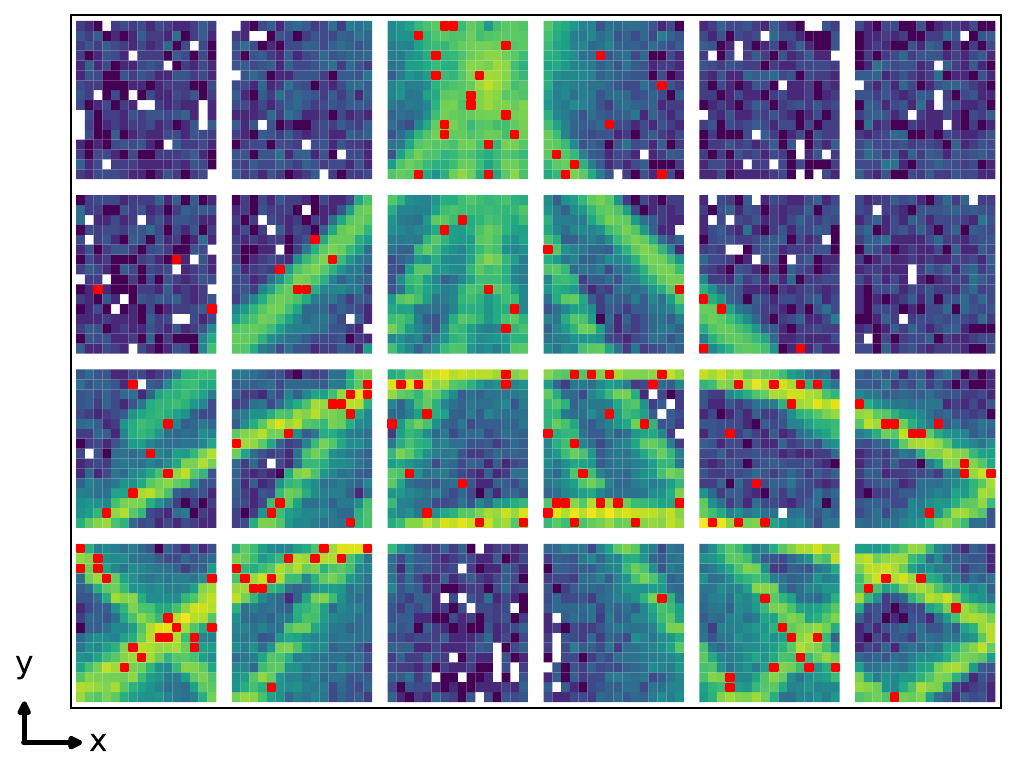} \\
    \caption{\textbf{Optical box output:} Individual tracks leave sparse hit patterns (red points) integrated over time on the hpDIRC readout plane. 
    The denser hit pattern is obtained by accumulating multiple tracks with the same kinematics. Figure and caption taken from \cite{giroux2025generativemodelsfastsimulation}.}
    \label{fig:sparse_hits}
\end{figure}

We utilize the dataset developed by Giroux et al. \cite{giroux2025generativemodelsfastsimulation}, which records Cherenkov photon hit patterns captured by the hpDIRC readout system comprised of photomultiplier tube (PMT) arrays. Each hit pattern is associated with a single charged particle track, which is inherently characterized by kinematic parameters. As depicted in Figure \ref{fig:sparse_hits}, an individual track produces a sparse and inherently stochastic subset of hits (highlighted in red). However, when aggregating hits from multiple tracks with identical kinematics—namely, momentum ($|\vec{p}|$) and polar angle ($\theta$)—a well-defined probability density function (PDF) emerges. Importantly, the number of hits per track is not fixed, and dependent on the kinematics, resulting in a combinatorially large space of possible hit configurations. This variability presents a significant modeling challenge for autoregressive (AR) architectures.

As stated in \cite{giroux2025generativemodelsfastsimulation}, the generated charged tracks (both pions and kaons) are distributed approximately uniformly over the phase-space, spanning $1<|\vec{p}|<10$~GeV/c, $25<\theta<160^\circ$, providing $\sim$ 5 million tracks for each particle type. We limit ourselves to a max sequence length of 250, corresponding to a large range in photon yields across the phase-space. In contrast to prior works learning information at the photon-level, in which tracks can be formed through aggregation, we instead directly learn to generate individual tracks (sequences of hits). %

\section{Methods}\label{sec:methods} 

\subsection*{Tokenization of Space and Time}

Given the pixelated nature of the readout system, no additional tokenization is required along the spatial dimension; each discrete pixel index corresponds uniquely to a fixed $(x, y)$ coordinate in the PMT array, yielding a spatial vocabulary size of 6144. In regards to time, which is a continuous variate, we employ a simple linear binning strategy in which we discretize on the order of $1/4$ the time resolution yielding a vocabulary of size 5920. While we employ uniform binning in this work, more sophisticated schemes—such as non-linear binning or function-based discretizations—could be used to capture known nonlinearities in the readout system. We also observed that further reducing the temporal resolution to one-tenth of the intrinsic resolution had negligible impact on model performance. Therefore, our current choice of 1/4 the timing resolution provides similar vocabulary sizes for both spaces

\subsection*{Time Queries Space}

We first obtain time and spatial embeddings through learnable, independent projections from their vocabularies, in which we add positional embeddings along each dimension. We also embed both the momentum and angular values through independent linear projections (continuous spaces), and prepend our kinematics as initial context along both time and spatial projections, an example of which is given in Eq. \ref{eq:tokens}. Each embedding dimension is set to 256 in our experiments. 
We therefore define the following sequences:\footnote{Here, \texttt{SOS} and \texttt{EOS} denote the \textit{Start of Sequence} and \textit{End of Sequence} tokens, respectively.}

\begin{align}\label{eq:tokens}
    \text{spatial} &\rightarrow \{|\vec{p}|,\theta,\text{SOS}_p,p_1, \ldots , p_{n}, \text{EOS}_p\} \notag \\
    \text{time} &\rightarrow \{|\vec{p}|,\theta,\text{SOS}_t,t_1, \ldots , t_{n}, \text{EOS}_t\}
\end{align}

We then combine information through a transformer block using CMHCA, in which we obtain our Query projection (Q) from the time embeddings, and the Key (K) and Value (V) projections from the spatial embeddings.\footnote{For a given time, query the possible pixel locations of valid Cherenkov photons.} The result is a spatio-temporally aware embedding to be further processed by transformer blocks utilizing traditional Multi-Head Self-Attention (MHSA) blocks \cite{vaswani2017attention}. Each transformer block consists of its respective attention type with 8 heads, a linear projection layer, and a standard feed forward neural network (FFNN) with 2 linear layers and GeLU activation \cite{hendrycks2016gaussian}. Given the large combinatorial space of plausible tracks at a given kinematics, careful consideration must be taken within each attention head. Specifically, we must enforce the model to learn more focused attention maps given an initial context. We do not want the attention maps to simply collapse to the most probable configurations as aggregations over the entire phase-space. Therefore, we employ a pre-normalization scheme in the transformer blocks \cite{ba2016layer,xiong2020layer}. We further augment this through normalization of the Q,K matrices \cite{henry2020query}, in which $\ell_2$ normalization is applied with a learnable scale factor. This makes attention scores invariant to their scale - promoting stability, increased variability in our output space and direction focused. This being a key feature given the large combinatorial space of plausible track configurations mentioned prior.

\begin{figure}[!]
    \centering
    \includegraphics[width=0.55\textwidth,angle=270]{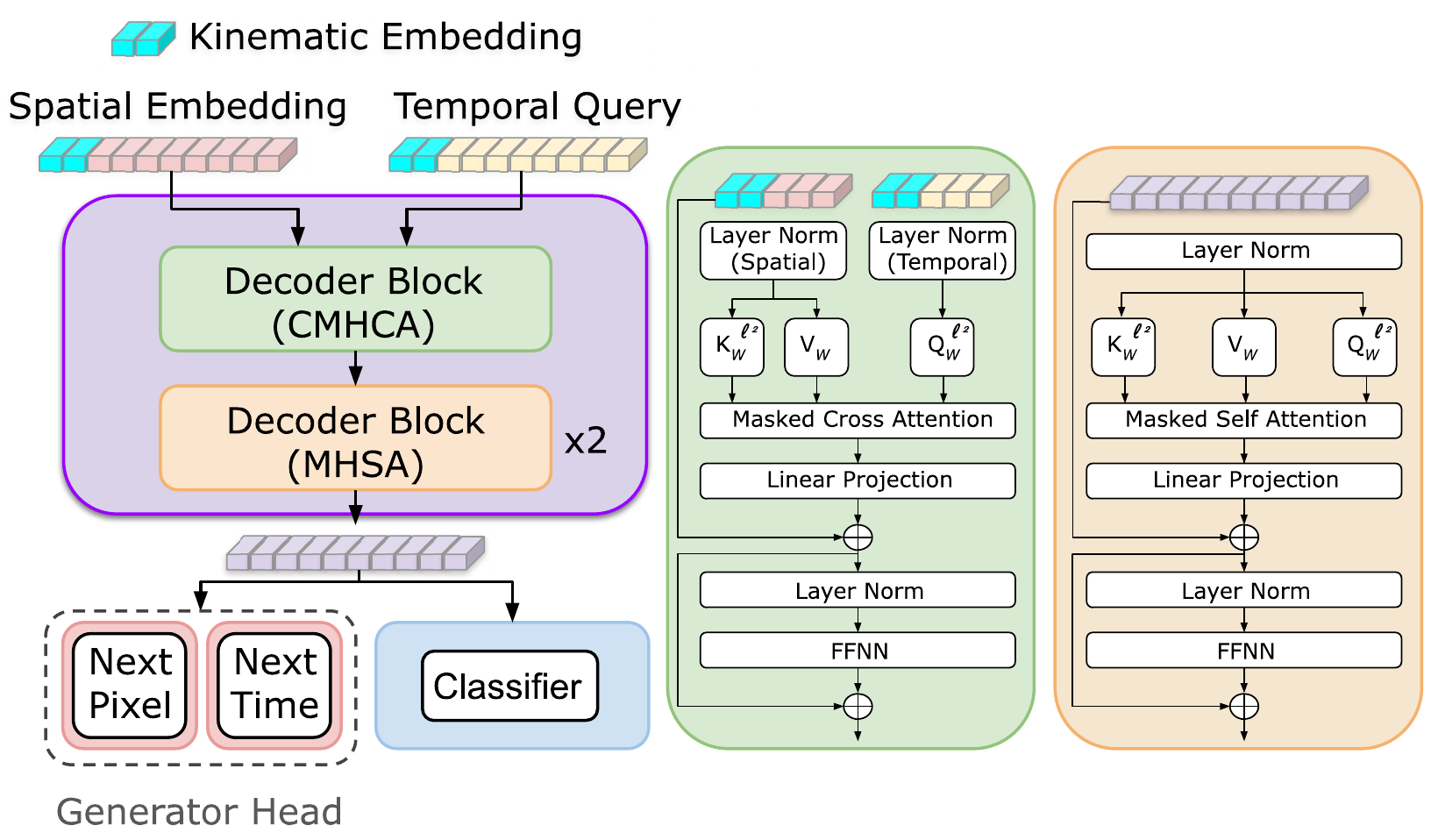}
    \caption{\textbf{Architecture:} Spatial and temporal sequences are first independently embedded via learnable projections from their respective vocabularies, with positional encodings added along each dimension. Kinematic information (momentum magnitude and polar angle) is projected from continuous space and prepended as contextual tokens to both sequences. The time embeddings serve as queries (Q), while the spatial embeddings provide keys (K) and values (V) within a Causal Multi-Head Cross Attention (CMHCA) mechanism. The resulting spatio-temporal representations are passed through transformer blocks utilizing Multi-Head Self Attention (MHSA). We apply a pre-normalization scheme, and include $\ell_2$-normalized Q and K projections with learnable scaling. For generation, the model outputs next-token predictions over space and time vocabularies through two linear heads. In the case of class conditional generation with a Mixture of Experts, the experts replace the Feed Forward Neural Networks (FFNNs). For sequence level classification (particle identification), a \textit{CLS} token is used to summarize the sequence and predict class scores via a final linear layer. Token level classification (noise filter) uses a single linear layer to produce per token latent distributions for classification.}
    \label{fig:model}
\end{figure}

In the case of generation, the final output from the transformer blocks feeds two independent next token prediction heads operating over the space and time vocabularies. Each of which are represented by single linear layers. The loss is then the linear combination of the Cross Entropy (CE) over both vocabularies.
In the case of classification (sequence level prediction), we represent our output latent distributions through a single linear layer. This layer is fed through a traditional \textit{CLS} token format, in which we place this token at the start of each sequence. The loss is then given by traditional Binary Cross Entropy (BCE).
Similarly, for filtering tasks (token level prediction), we represent the per-token latent distributions through a single linear layer operating element-wise. Given that the the signal to noise ratio is high, the class of noise is generally underrepresented. As a result, we use a focal-loss \cite{lin2017focal} with $\alpha = 0.75$ and $\gamma = 2$. A diagrammatic representation of our model is shown in Fig. \ref{fig:model}.

\subsection*{Class Conditional Generation with a Mixture of Experts and Fixed Routing}

A Mixture of Experts \cite{shazeer2017outrageously} replaces the traditional FFNN layers of a Transformer with a set of \textit{experts}---multiple FFNNs designed to specialize on different aspects of data and potentially increase interpretability \cite{genovese2025mixture}. Rather than all tokens being passed through the same shared FFNN, each token is dynamically routed to a subset of experts based on a gating mechanism, often referred to as a \textit{Router}. Let $R(x)$ and $E_i(x)$ denote the router (a simple linear matrix operation), and the $i-$th expert. The output of a MoE layer is then given by the weighted sum of the router and the $N_{E}$ experts, Eq. \ref{eq:MoE}, where $R_i(x)$ is the softmax of the $i-$th expert.\footnote{In Eq. \eqref{eq:MoE}, $s$ and $d$ stand for the sequence length and the embedding dimension, respectively.}

\begin{equation}\label{eq:MoE}
    z = \sum_{i=1}^{N_{E}} R_i(x) E_i(x), \; \; R_i(x) \in \mathbb{R}^{s \times 1}, \; \; E_i(x) \in \mathbb{R}^{s \times d}
\end{equation}

To increase computational efficiency, Shazeer et al. \cite{shazeer2017outrageously} employ a Top-$k$ selection criteria, sparsifying the space of experts by allowing only the most probable $k$ experts to contribute. In our work, we instead deploy a class conditional selection criteria, closely resembling that of Top-k selection (see Fig. 1 in \cite{shazeer2017outrageously} for a visual depiction of Top-k selection).
Specifically, given an input sequence and its associated class label (\textit{e.g.}, pion or kaon), we fix the routing to a subset of experts through masking of all other expert contributions. Each class (particle type) is then routed through its own subset of experts, allowing class-specific generation while sharing the bulk of the Transformer layers, \textit{i.e.}, simulation of pions and kaons through a unified model. This approach mimics the sparsity of Top-$k$ while introducing explicit expert-class alignment. 

We explore configurations with both two, and four total experts. In the two expert case, where each class is exclusively routed through a singular expert, no load balancing is required.
However, in the four (or more) expert case, each class is routed to $k=2$ experts
where output for each class is the weighted sum of the remaining $k$ experts. This introduces the potential for expert collapse, where only one expert dominates. To mitigate this, we include a load balancing loss that encourages the routing probabilities to be approximately uniform across the selected experts, as shown in Eq.~\ref{eq:load_balance}:

\begin{equation}\label{eq:load_balance}
    \mathcal{L}(x)_{\text{balance}} =  \sum_{i=1}^k\left(R_i(x) - \frac{1}{k}\right)^2
\end{equation}

For other tasks, such as sequence (PID) or token (filtering) level classification, MoE is not used by default. In what follows we show that such a model can still be used for fine-tuning by setting the weights of the standard FFNN as the average of the experts.

\section{Analysis and Results}\label{sec:results}

\subsection*{Generative Model Evaluation}

We follow the procedure in Giroux et al. \cite{giroux2025generativemodelsfastsimulation}, validating our fast simulation approach at the central region in momentum ($\SI[per-mode=symbol]{6}{\giga\eVperc}$) through a series of histograms and ratios, in which we will also compare the learned sequence length, \textit{i.e.}, ability to learn correct photon yield as a function of the phase-space. Plots for additional kinematics (\textit{e.g.}, 3 GeV/c and 9 GeV/c) can be found in \ref{app:3GeV} and \ref{app:9GeV}. Note that while our model directly provides pixel and timing location, we convert the spatial indices into $x,y$ coordinates to provide more informative visualizations. This will then be followed by a  closure test using classifier metrics, namely the same Kernel Density Estimation (KDE) derived from FastDIRC \cite{hardin2016fastdirc} at fixed momentum. In which we ultimately translate the fidelity of our generations into a more interpretable metric of separation power. We also conduct closure tests of generative models both with (class conditional generation), and without a MoE, in which their performance should be equivalent given effective class conditioning. In the case of the former, we compare models with four and two experts, resulting in two or a singular expert per class.

\subsubsection*{Histogram Level Evaluations}

We begin through visual comparisons of generations produced by our architecture to the ground truth of \geant at fixed momentum. We show generations from our model near the two ends of the bar (large, or small polar angle), along with the central region of the bar. 
As noted in \cite{giroux2025generativemodelsfastsimulation}, charged particles entering the bar at polar angles near the expansion volume \cite{Kalicy_2020} produce a clear temporal separation between direct and indirect photons. In this case, indirect photons travel the full bar length, reflect from the far end, and return to the readout with significantly delayed arrival times. For particles traversing the central region of the bar, indirect photons reflect after approximately half the bar length, resulting in intermediate delays. Conversely, particles near the far end of the bar generate both direct and indirect photons that arrive nearly simultaneously, making their separation more difficult.
These variations in photon propagation paths across polar angles result in a range of timing topologies that serve as a rigorous test of the model’s ability to capture geometric and temporal features in Cherenkov photon detection.
While not quantitative in nature, fixed point generation comparisons are extreme stress tests of our model given its training on a fully continuous phase space. By forcing the model to generate many samples at isolated points it may have rarely---or even never---seen during training, we directly test its ability to generalize smoothly and coherently across the conditioning space. Moreover, these simple projections provide immediate cues into potential mode collapse or blending of kinematic dependencies, ultimately resulting in deviation from correct rings structures. 

\begin{figure}[h]
    \centering
    \includegraphics[width=0.48\textwidth]{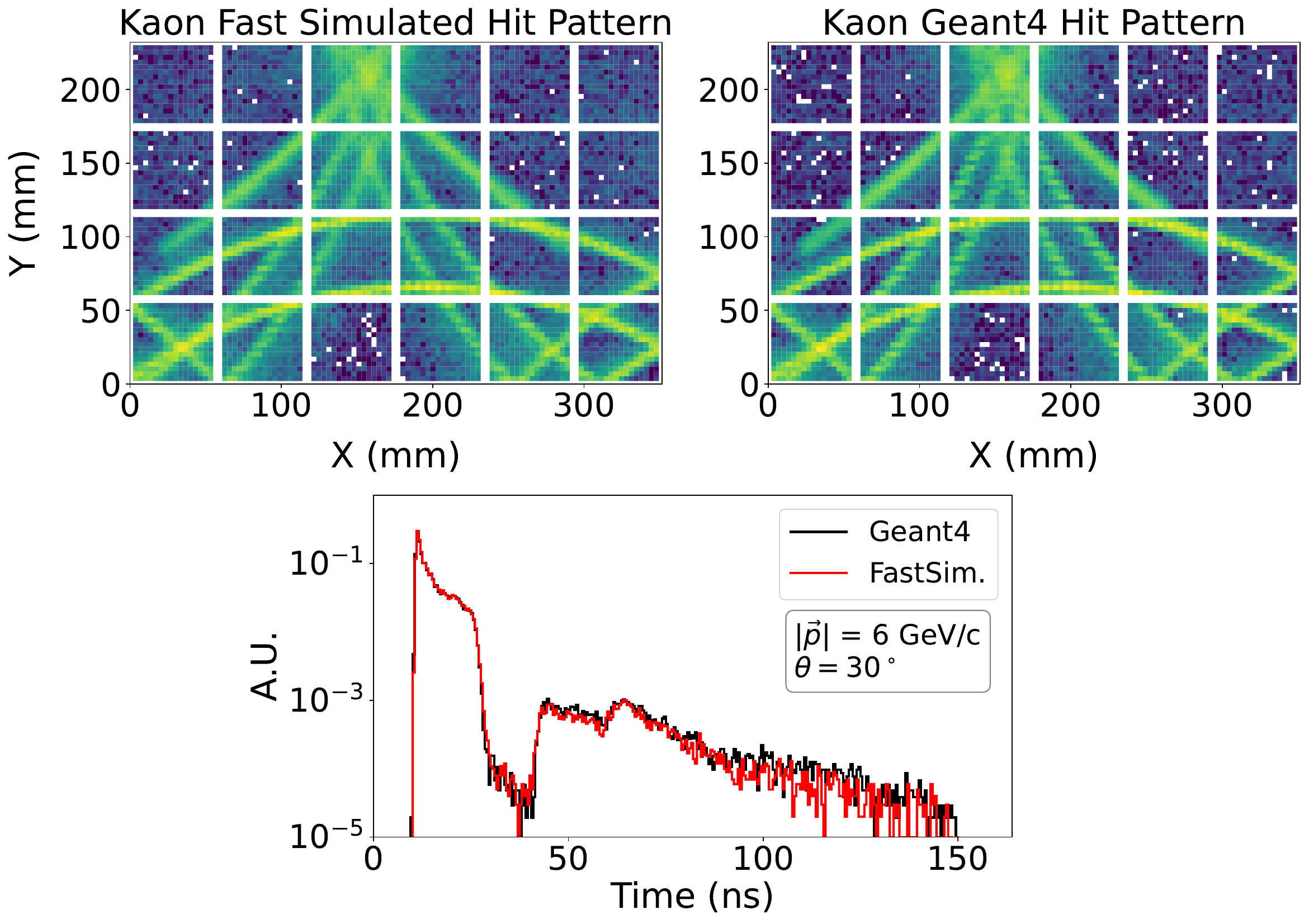}% 
   \includegraphics[width=0.48\textwidth]{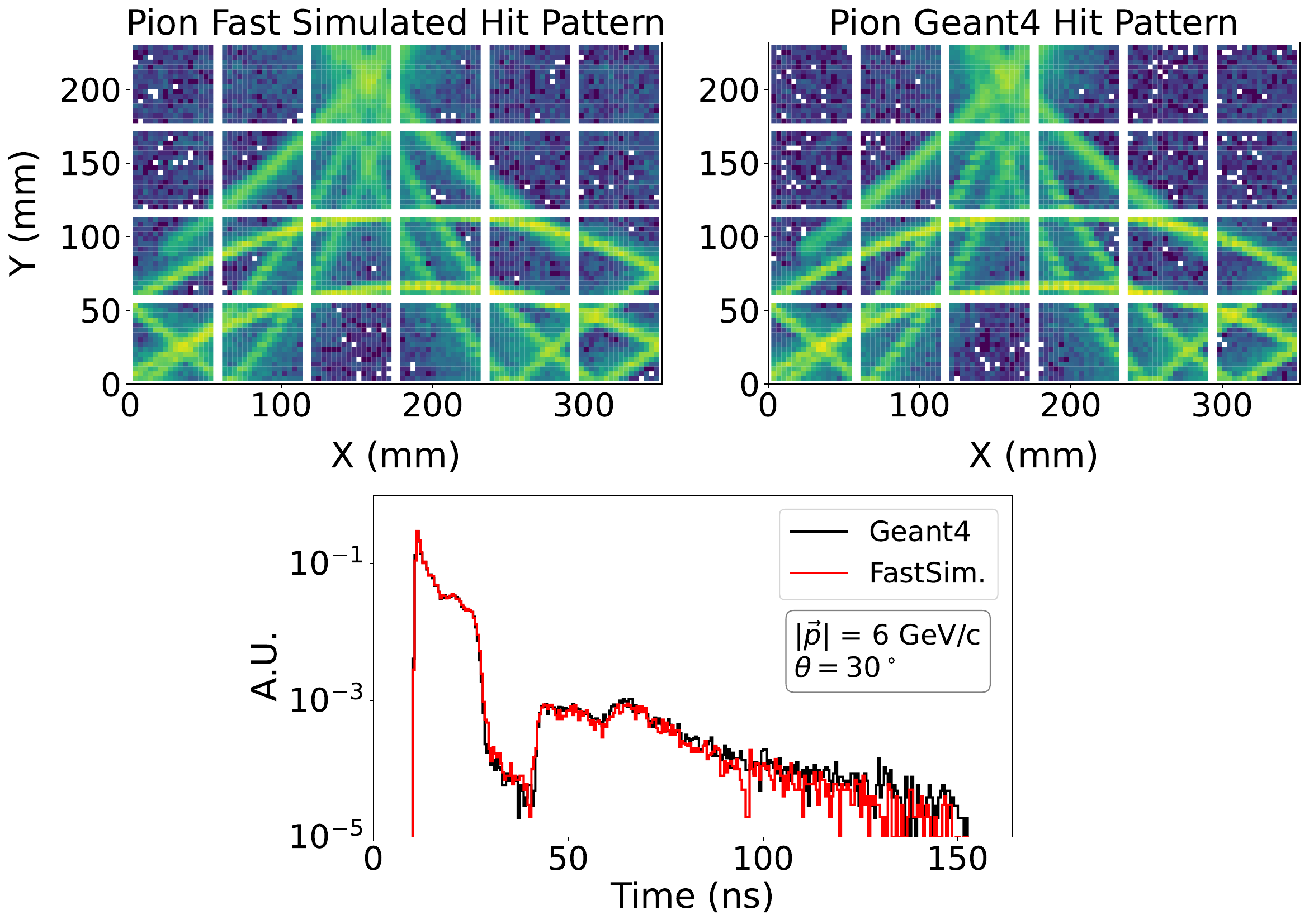} \\
    \includegraphics[width=0.48\textwidth]{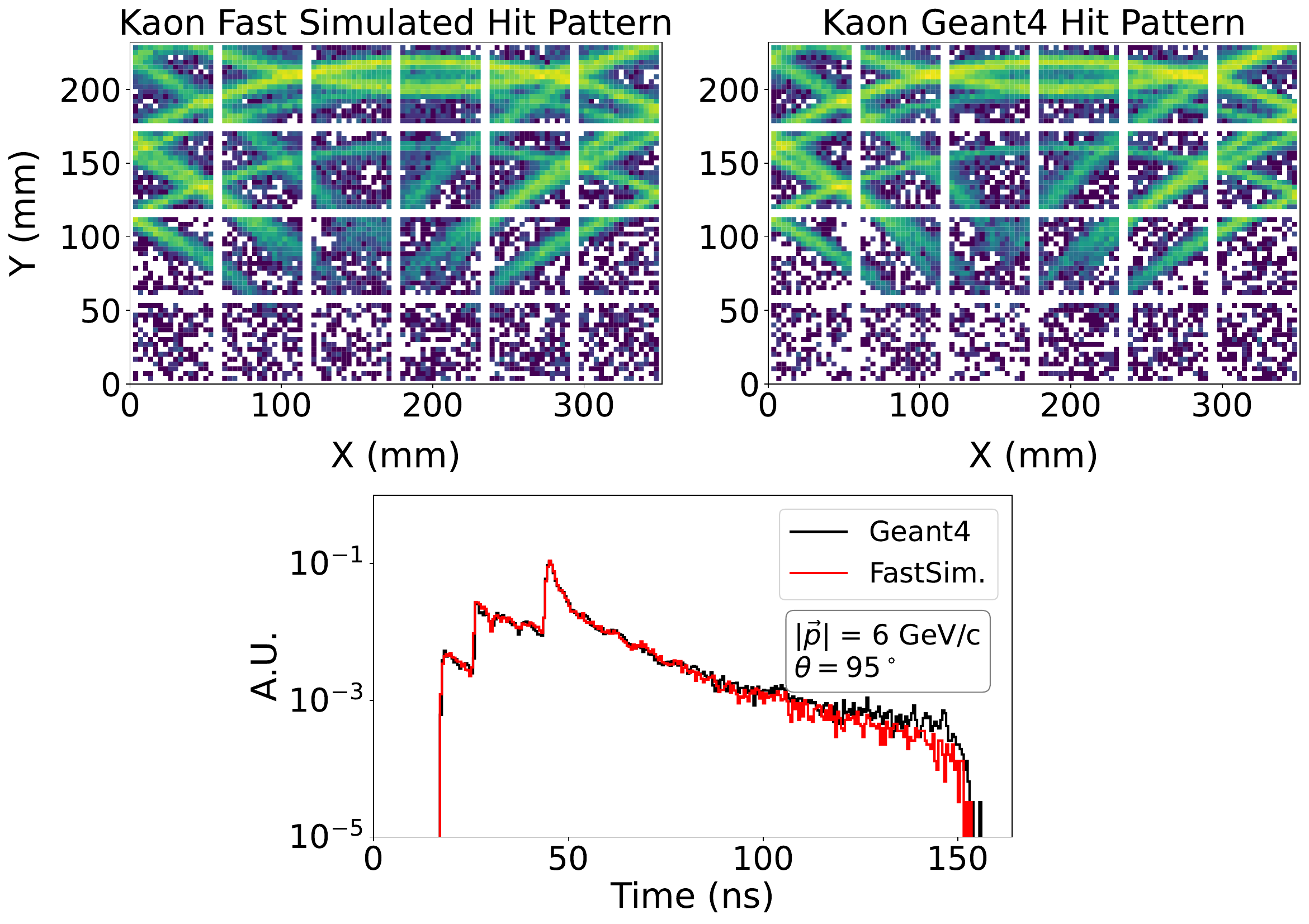} %
    \includegraphics[width=0.48\textwidth]{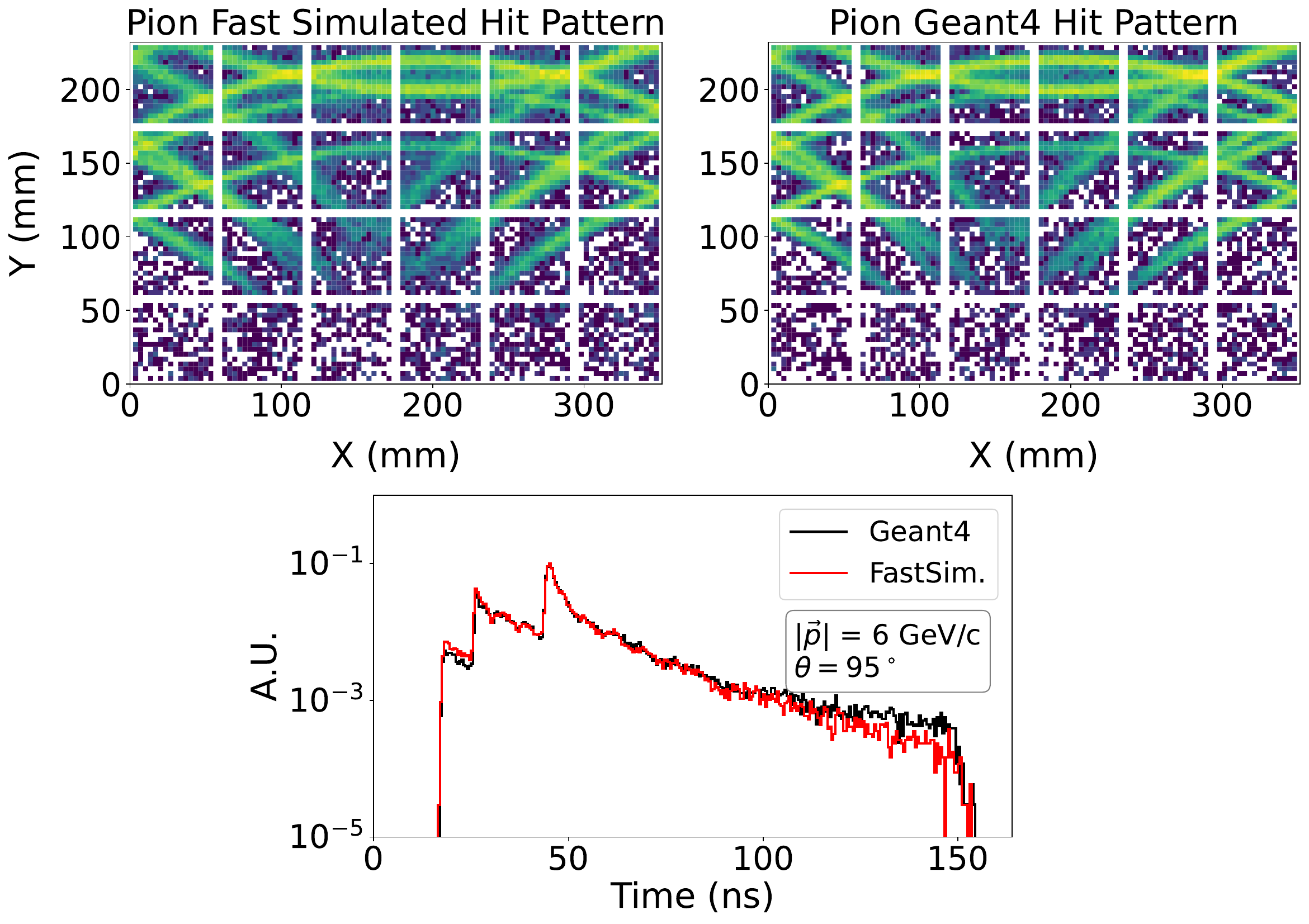} \\
    \includegraphics[width=0.48\textwidth]{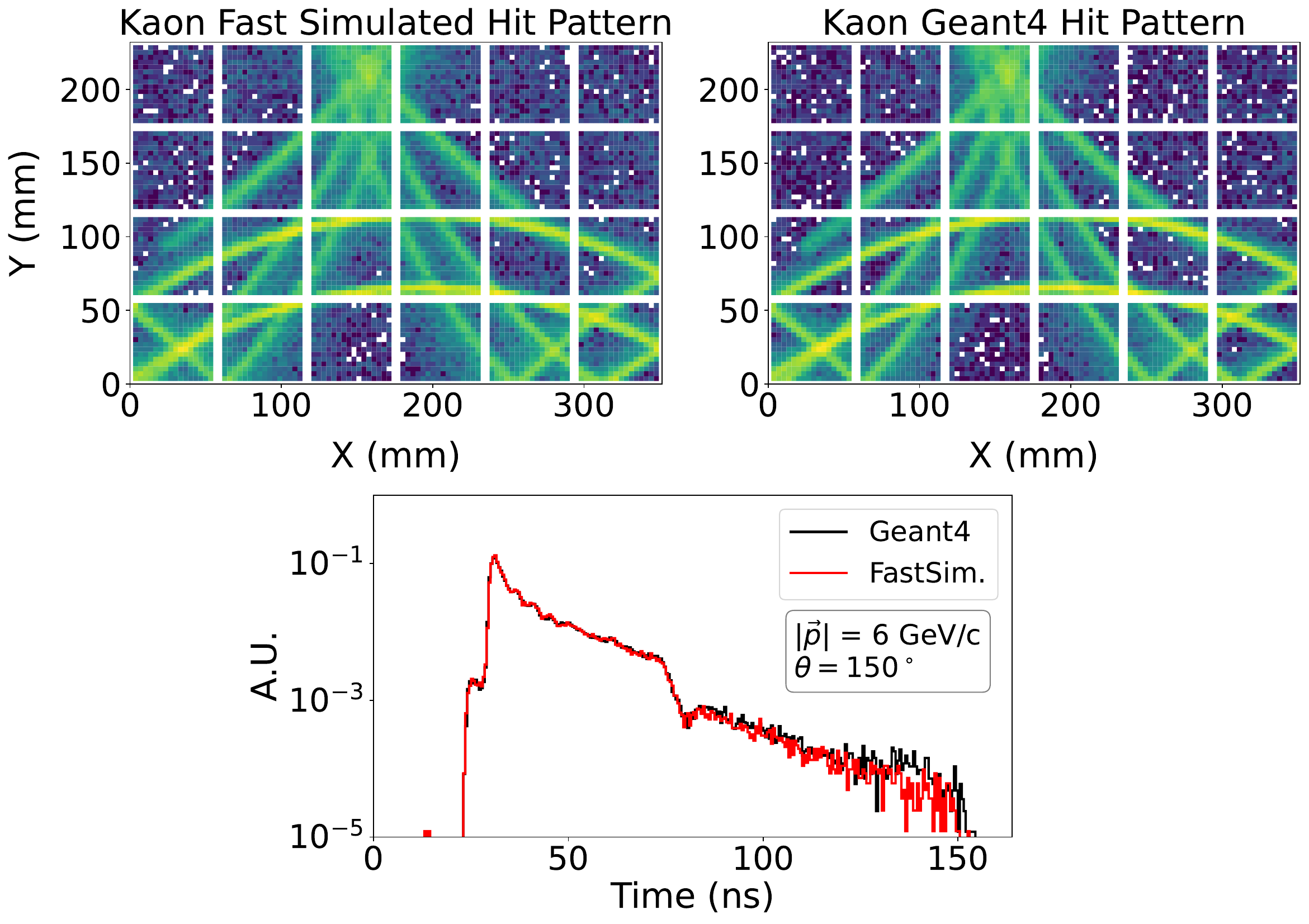} %
    \includegraphics[width=0.48\textwidth]{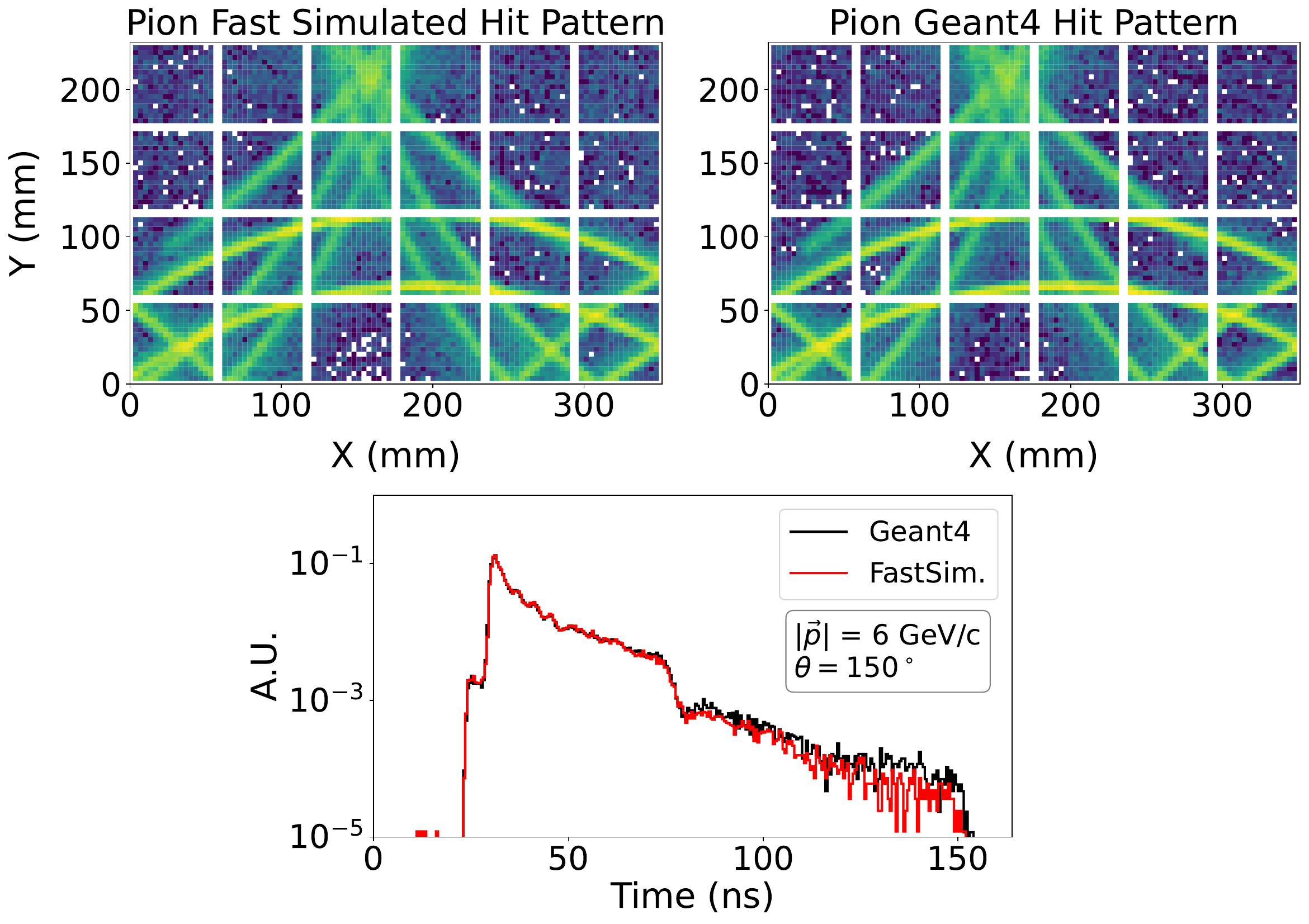} %
    \caption{
    \textbf{Fast Simulation at 6 GeV/c:} Fast Simulation of kaons (left column of plots), and pions (right column of plots) at 6 GeV/c at various polar angles, using Nucleus sampling, fixed temperature and a mixture of four experts (class conditional routing).}
    \label{fig:Generations_6GeV}
\end{figure}

In the generations that follow, we employ a Nucleus Sampling \cite{holtzman2019curious} technique with $p=0.995$, and a fixed temperature of $T=1.05$. These generations schemes have shown to provide the most consistent cumulative distributions across the phase-space. It should also be noted that given the high degree of stochasticity within individual sequences, small changes in these parameters can result in significantly different outputs. In fact, the idea of ``next token'' prediction of Cherenkov photons is highly ambiguous, as shown by previous works in which high fidelity track-level generations can be produced through aggregations of individually generated photons \cite{fanelli2024deep,giroux2025generativemodelsfastsimulation}. 
These works demonstrate that modeling individual photons as independent entities does not result in a loss of coherence in the generated distributions.
Fig. \ref{fig:Generations_6GeV} shows simulated kaon (left column) and pion (right column) events at three representative polar angles: \SI{30}{\degree} (top row), \SI{95}{\degree} (middle row), and \SI{150}{\degree} (bottom row), generated using a mixture of four experts.

\begin{figure}[h]
    \centering
      \begin{subfigure}[b]{0.49\textwidth}
        \centering
        \includegraphics[width=\textwidth]{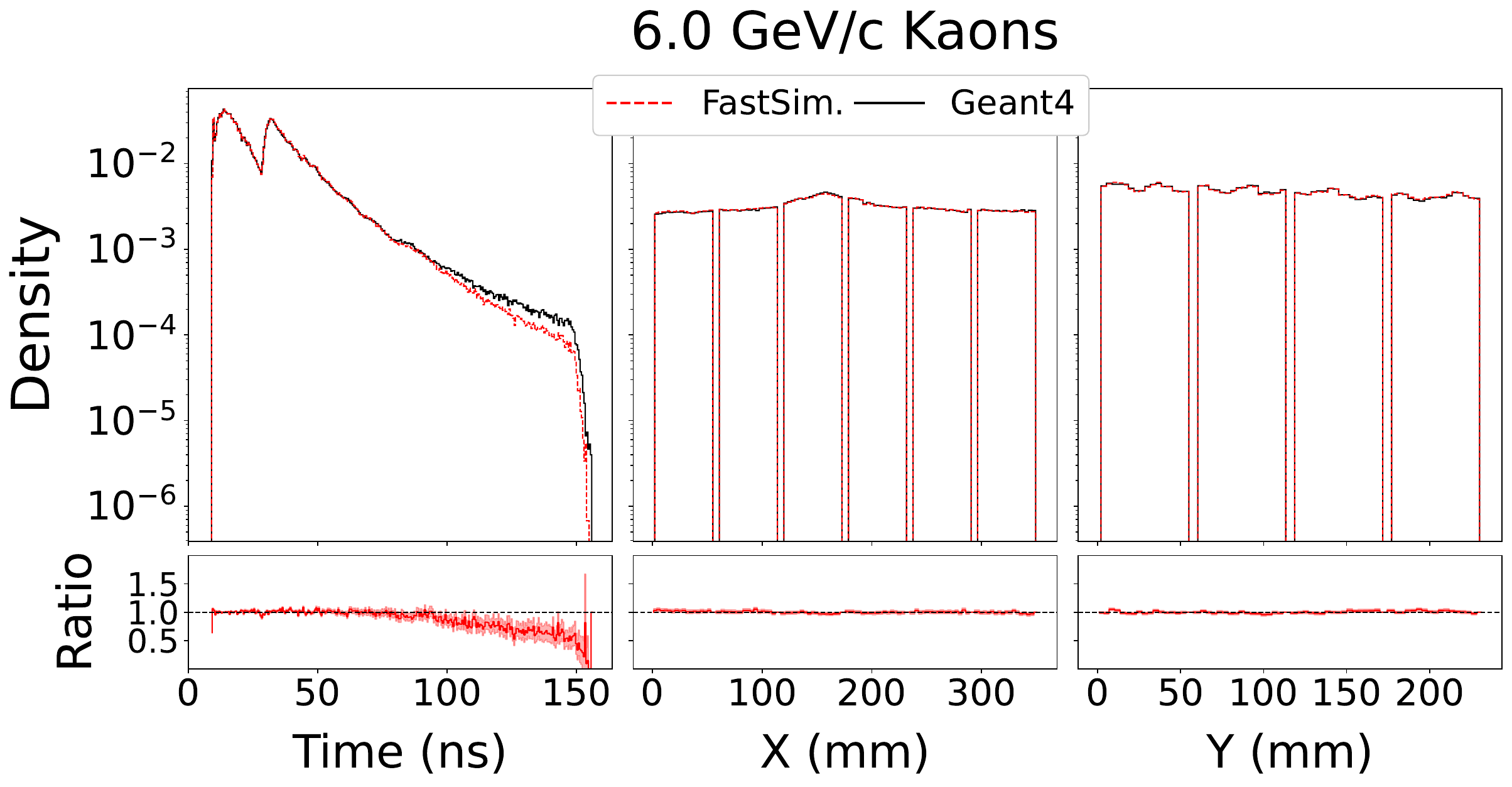}
        \caption{Kaons - Independent Model}
    \end{subfigure}
    \begin{subfigure}[b]{0.49\textwidth}
        \centering
        \includegraphics[width=\textwidth]{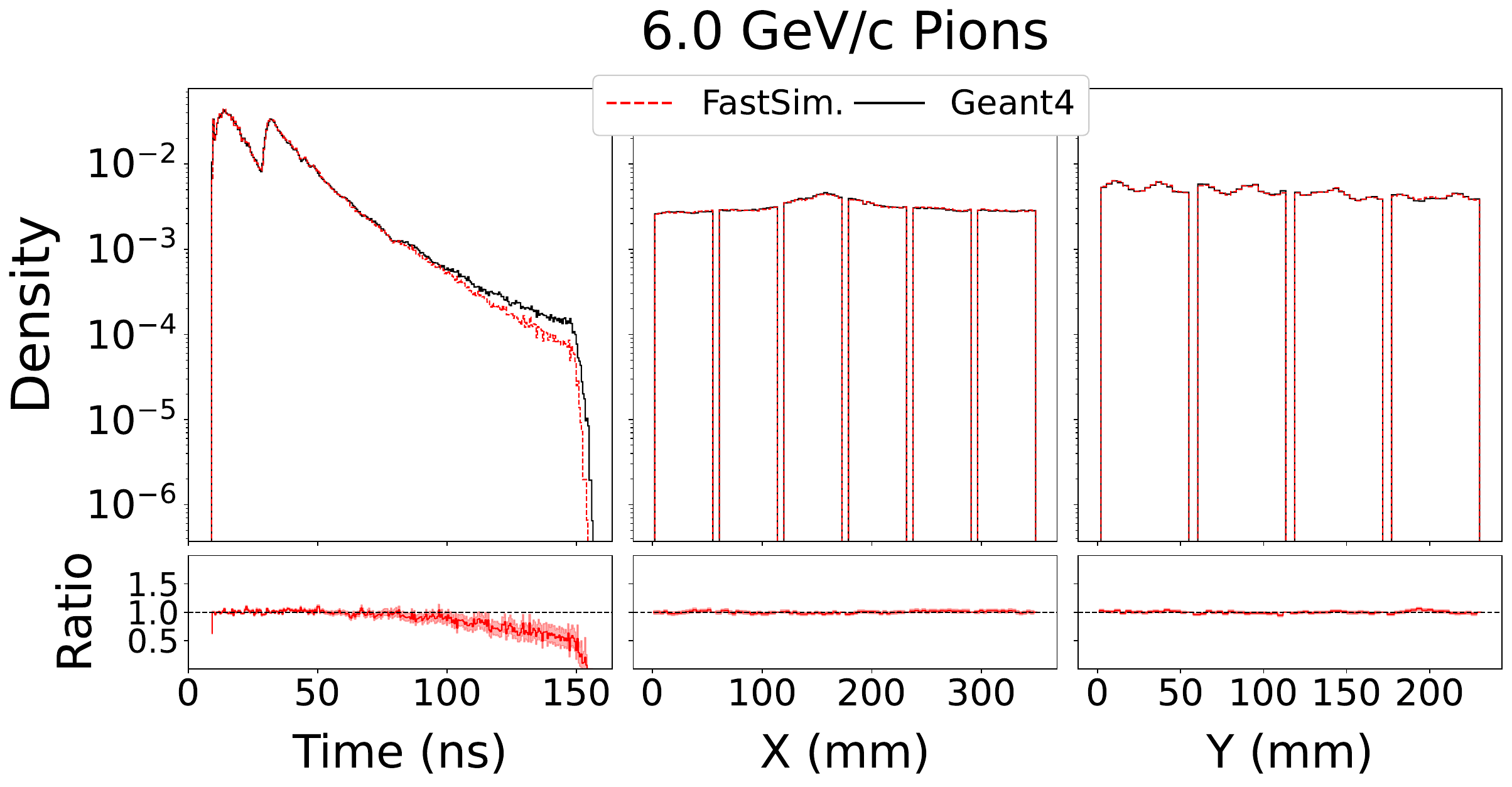}
        \caption{Pions - Independent Model}
    \end{subfigure} \\
    \begin{subfigure}[b]{0.49\textwidth}
        \centering
        \includegraphics[width=\textwidth]{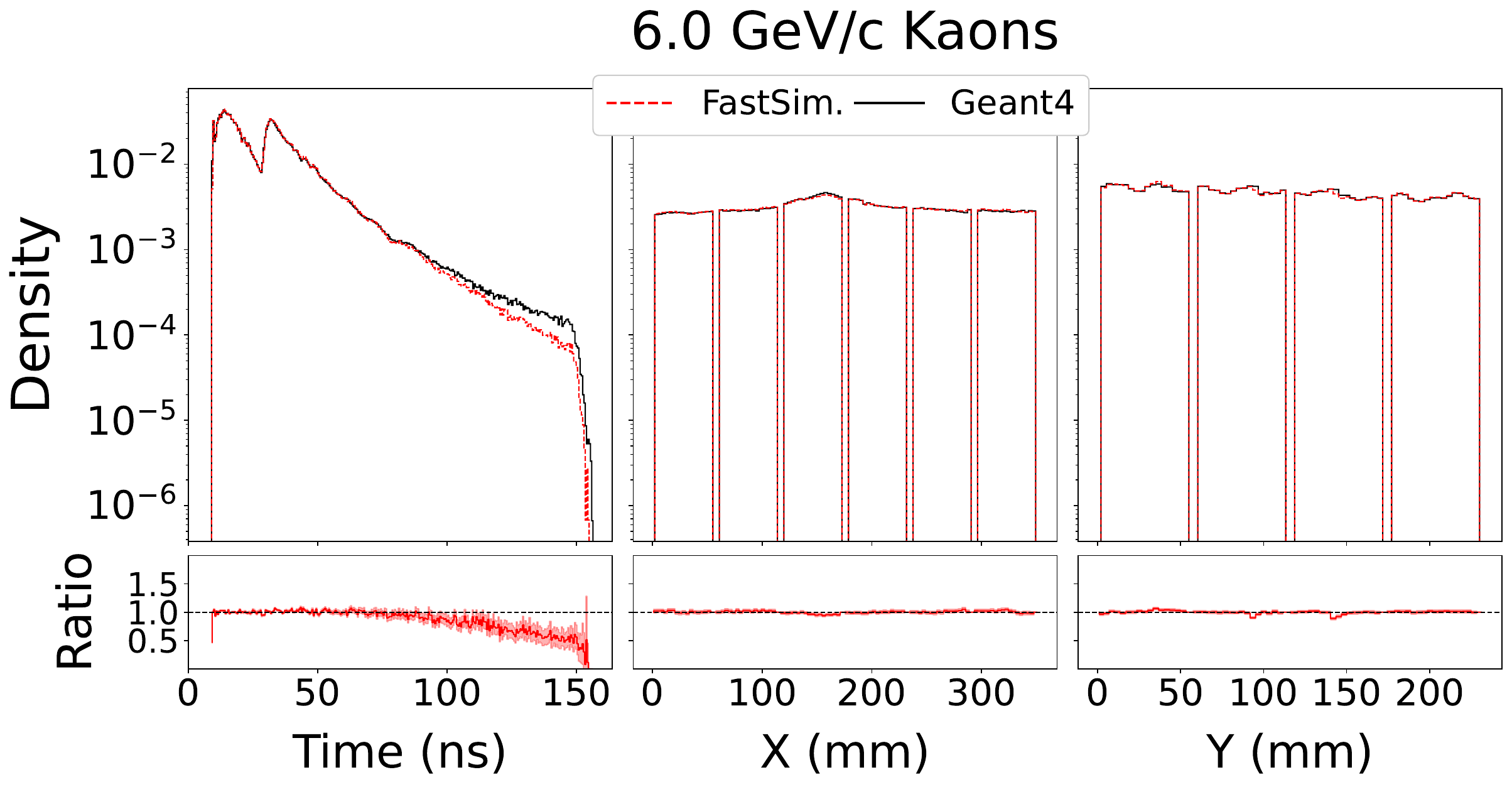}
        \caption{Kaons - Two Experts}
    \end{subfigure} 
    \begin{subfigure}[b]{0.49\textwidth}
        \centering
        \includegraphics[width=\textwidth]{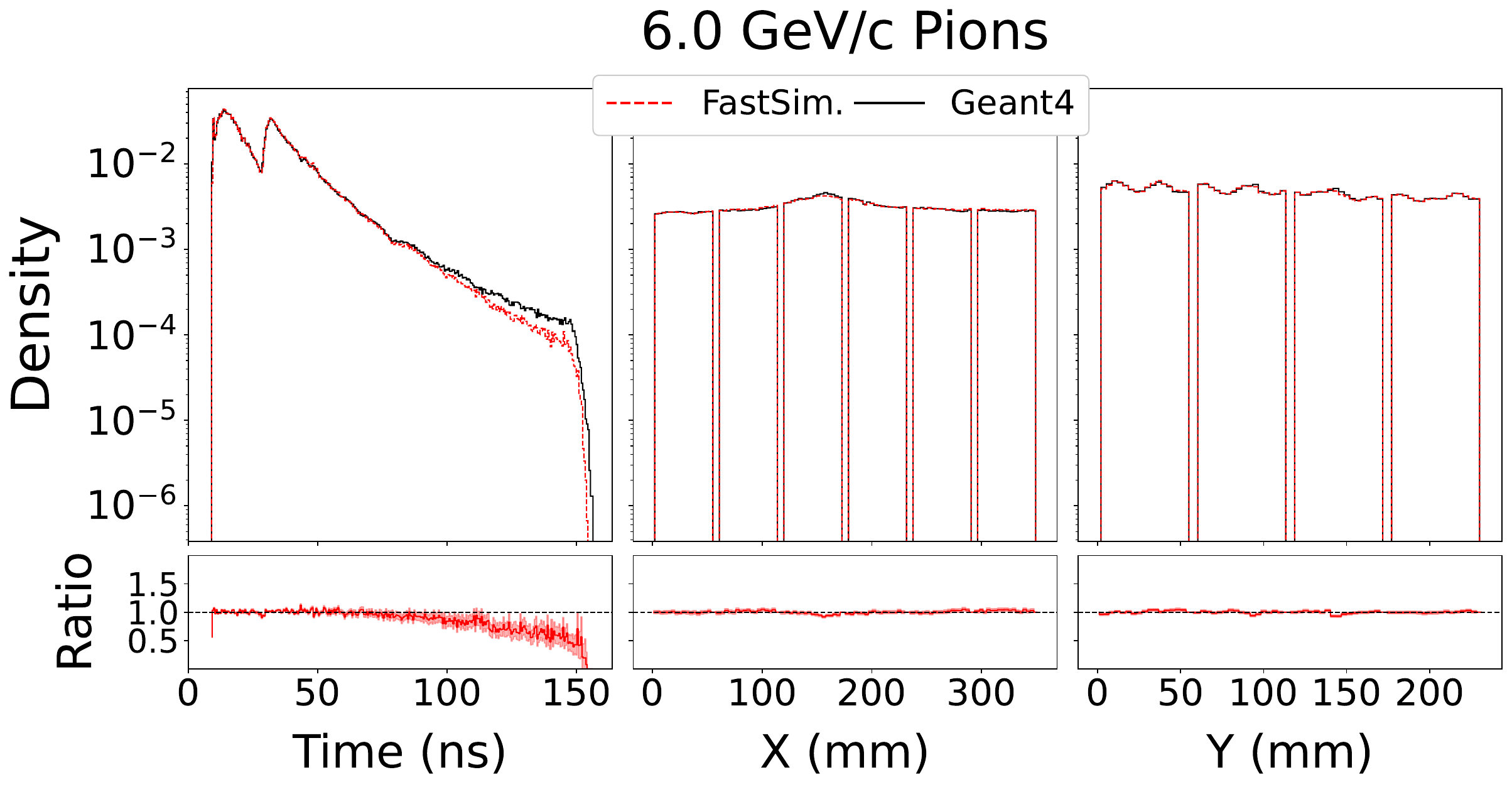}
        \caption{Pions - Two Experts}
    \end{subfigure} 
    \begin{subfigure}[b]{0.49\textwidth}
        \centering
        \includegraphics[width=\textwidth]{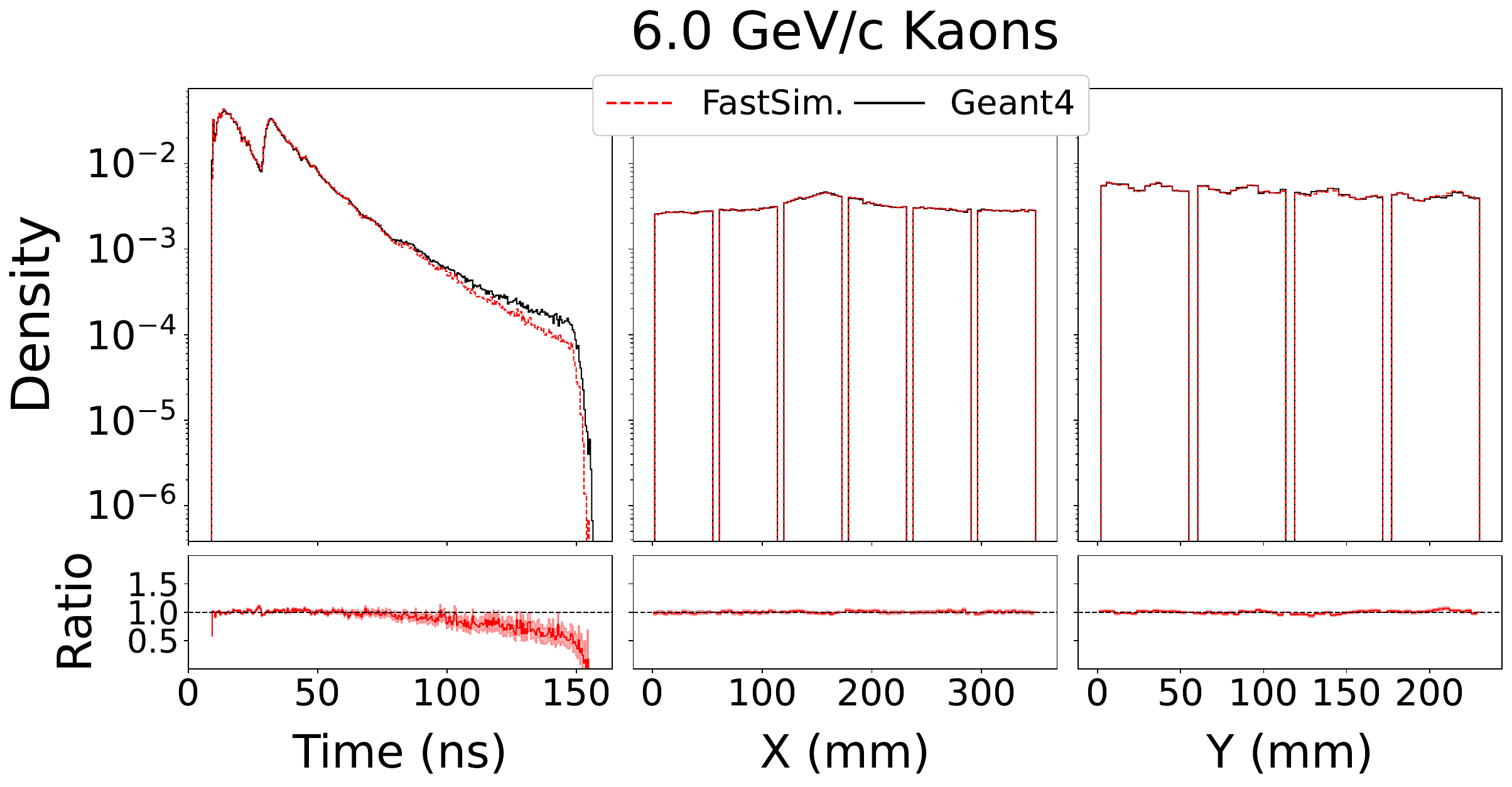}
        \caption{Kaons - Four Experts}
    \end{subfigure} 
    \begin{subfigure}[b]{0.49\textwidth}
        \centering
        \includegraphics[width=\textwidth]{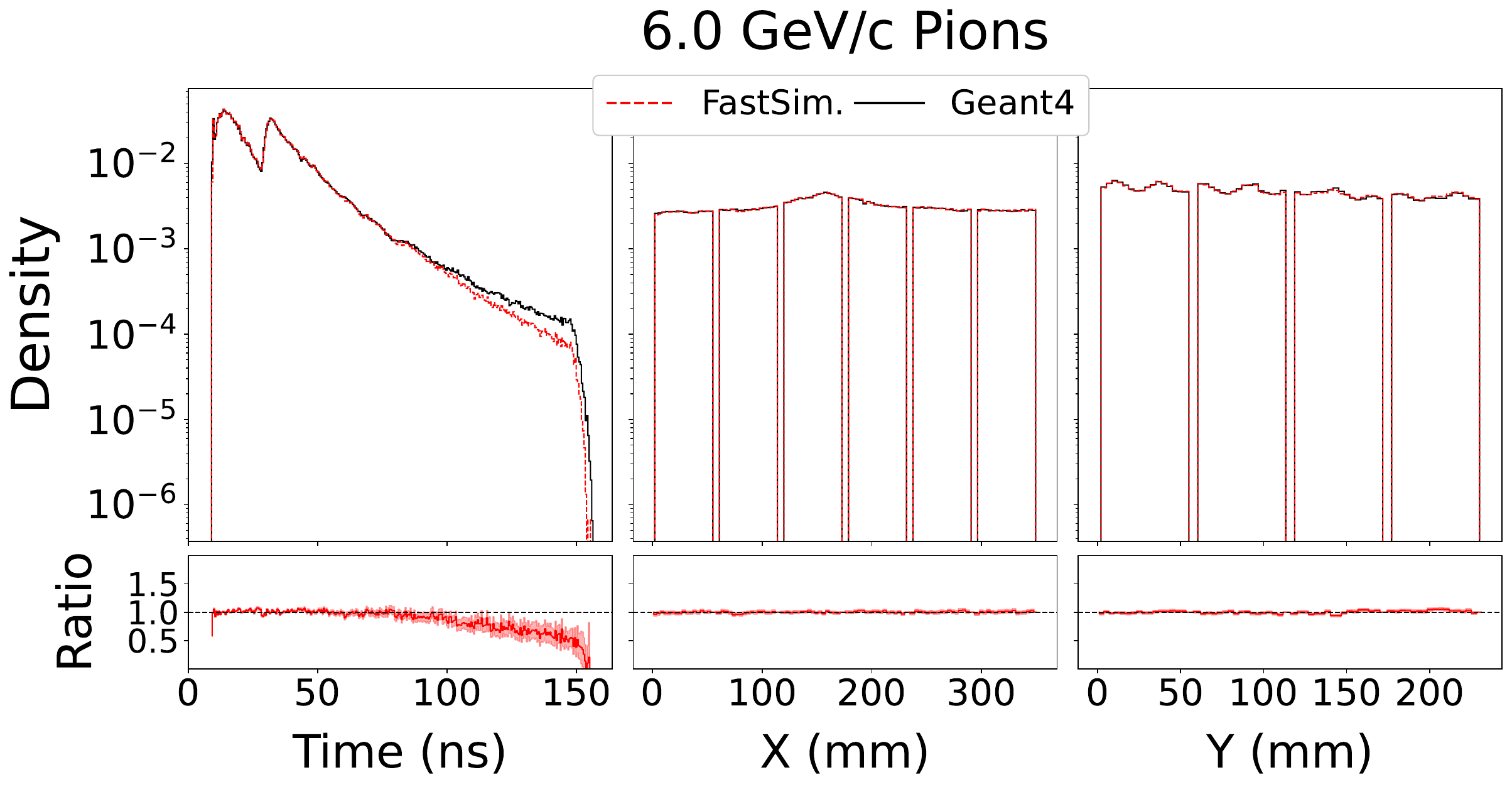}
        \caption{Pions - Four Experts}
    \end{subfigure} 
    \caption{\textbf{Ratio of synthetic to original data distributions at 6~GeV/\textit{c}:} Ratio plots for kaons (left column) and pions (right column), comparing generations to original data. The top row shows results from independent models using nucleus sampling with fixed temperature; the middle and bottom rows show results from combined models with two and four experts, respectively.
    }
    \label{fig:ratios}
\end{figure}

From inspection of the figure, we can qualitatively assess that our model correctly captures kinematic dependencies in both space and time.\footnote{In this work, all models---including those incorporating an MoE formulation---produce generations that are visually comparable.}
Furthermore, we are able to obtain hints of geometrical effects, such as the \textit{kaleidoscopic effect} \cite{dey2015design} at angles closer to the expansion volumes (\textit{e.g.}, $\SI{30}{\degree}$). %
To provide a more quantitative assessment of our generations, we provide cumulative distributions in Fig. \ref{fig:ratios}, for kaons (left column) and pions (right column) for independent models (top row), a mixture of two experts (middle row), and a mixture of 4 experts (bottom row).

\begin{figure}[!]
    \centering
      \begin{subfigure}[b]{0.49\textwidth}
        \centering
        \includegraphics[width=\textwidth]{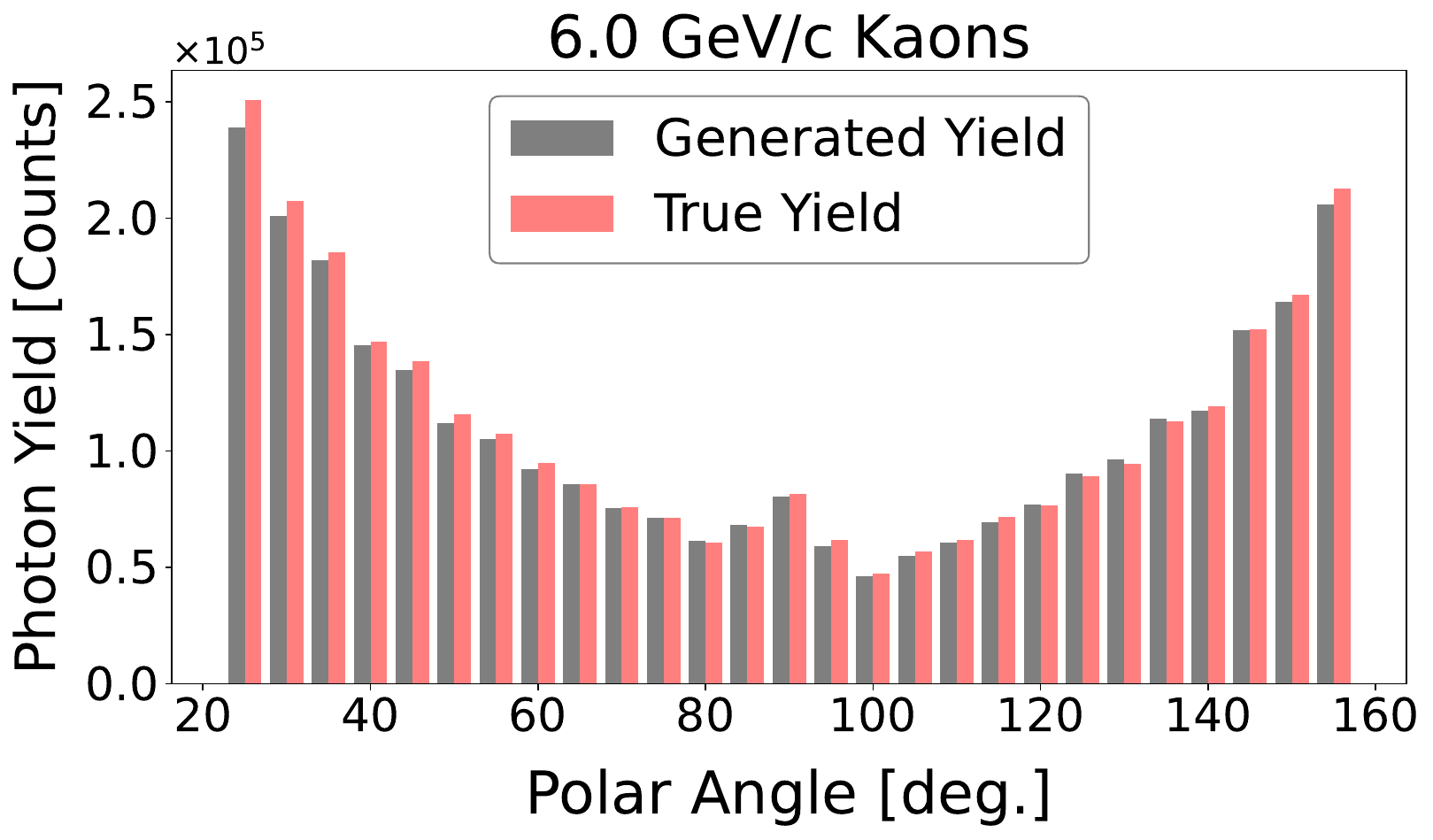}
        \caption{Kaons  - Independent Model}
    \end{subfigure}
    \begin{subfigure}[b]{0.49\textwidth}
        \centering
        \includegraphics[width=\textwidth]{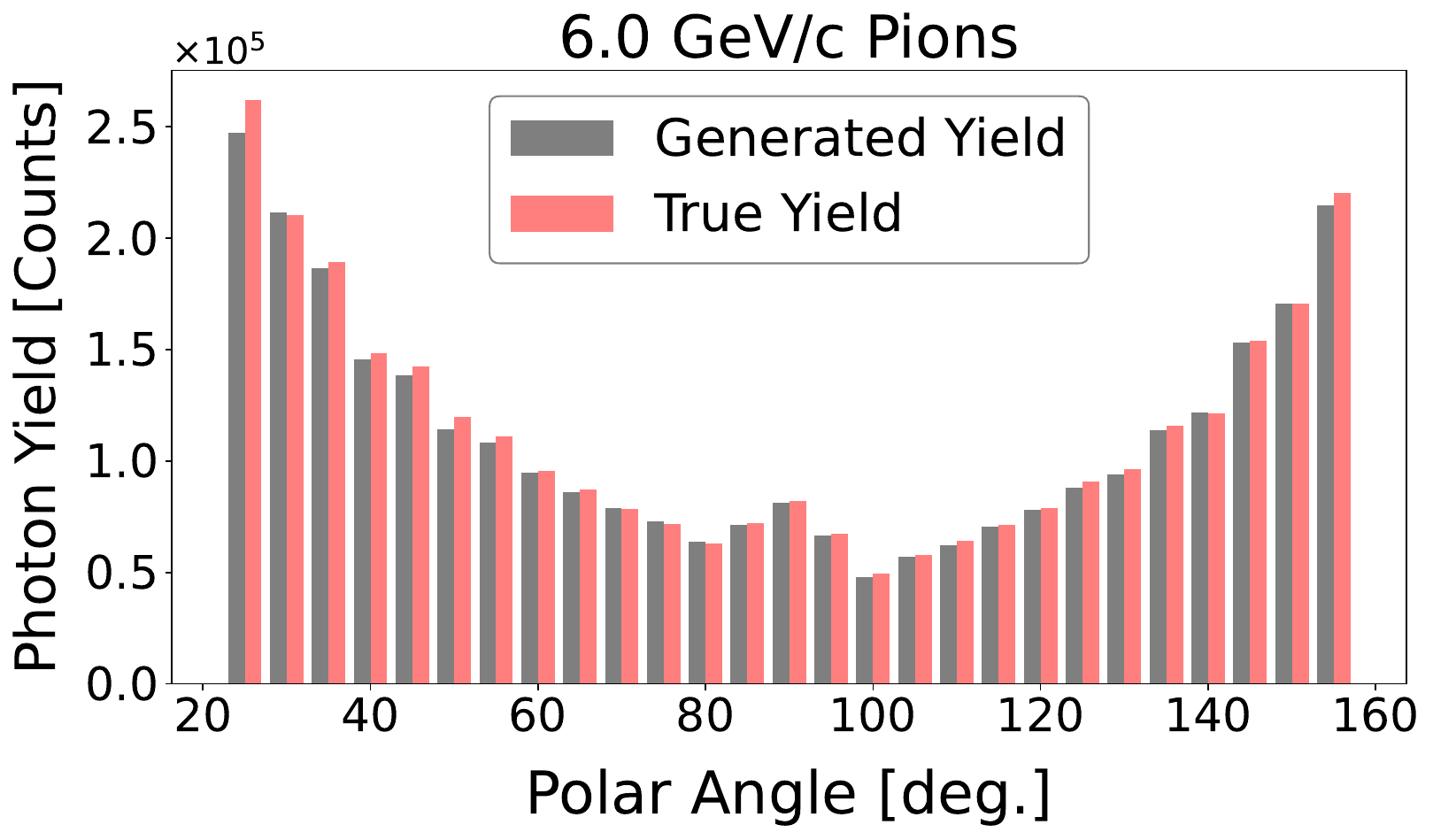}
        \caption{Pions  - Independent Model}
    \end{subfigure} \\
      \begin{subfigure}[b]{0.49\textwidth}
        \centering
        \includegraphics[width=\textwidth]{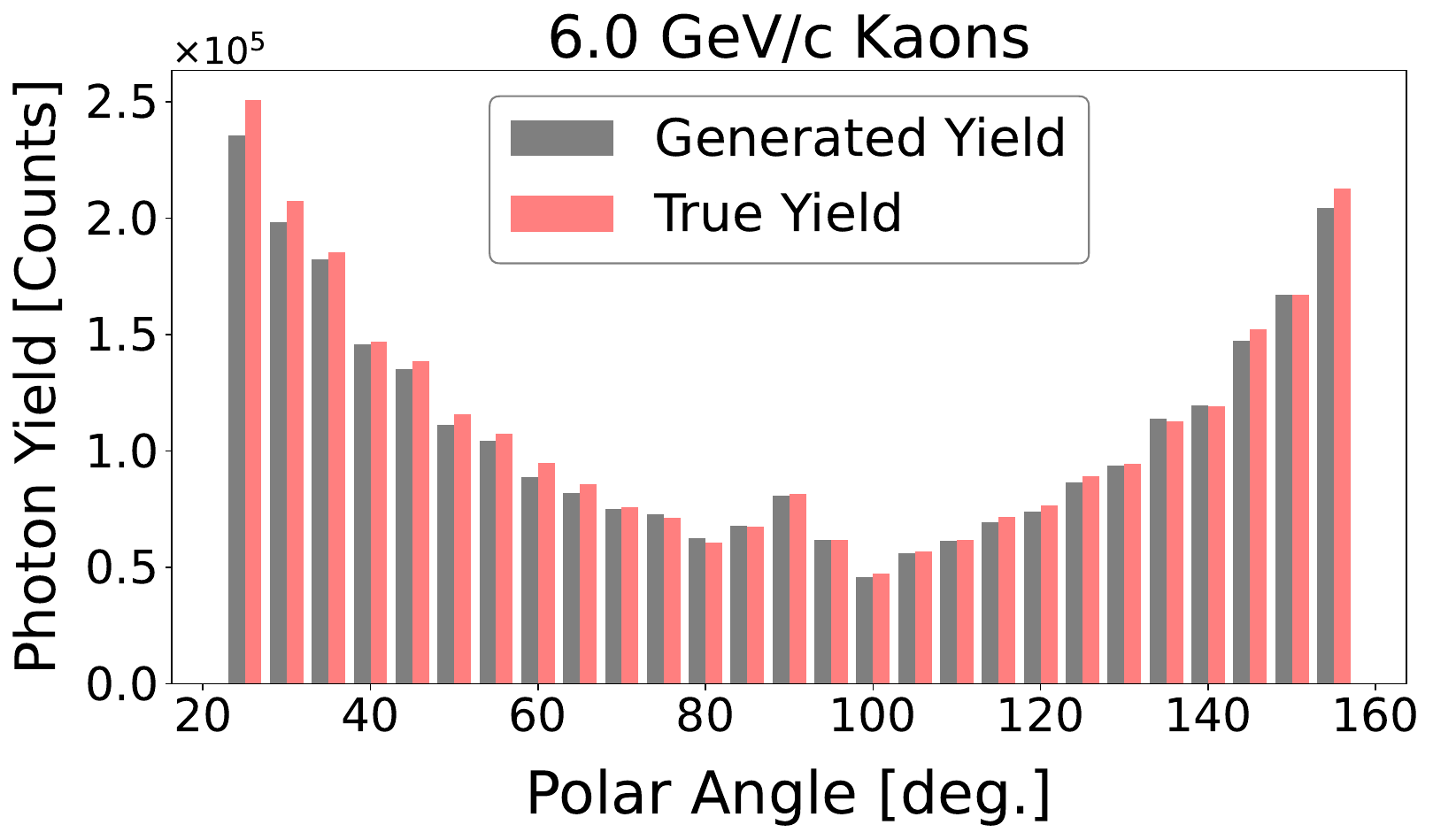}
        \caption{Kaons - Two Experts}
    \end{subfigure}
    \begin{subfigure}[b]{0.49\textwidth}
        \centering
        \includegraphics[width=\textwidth]{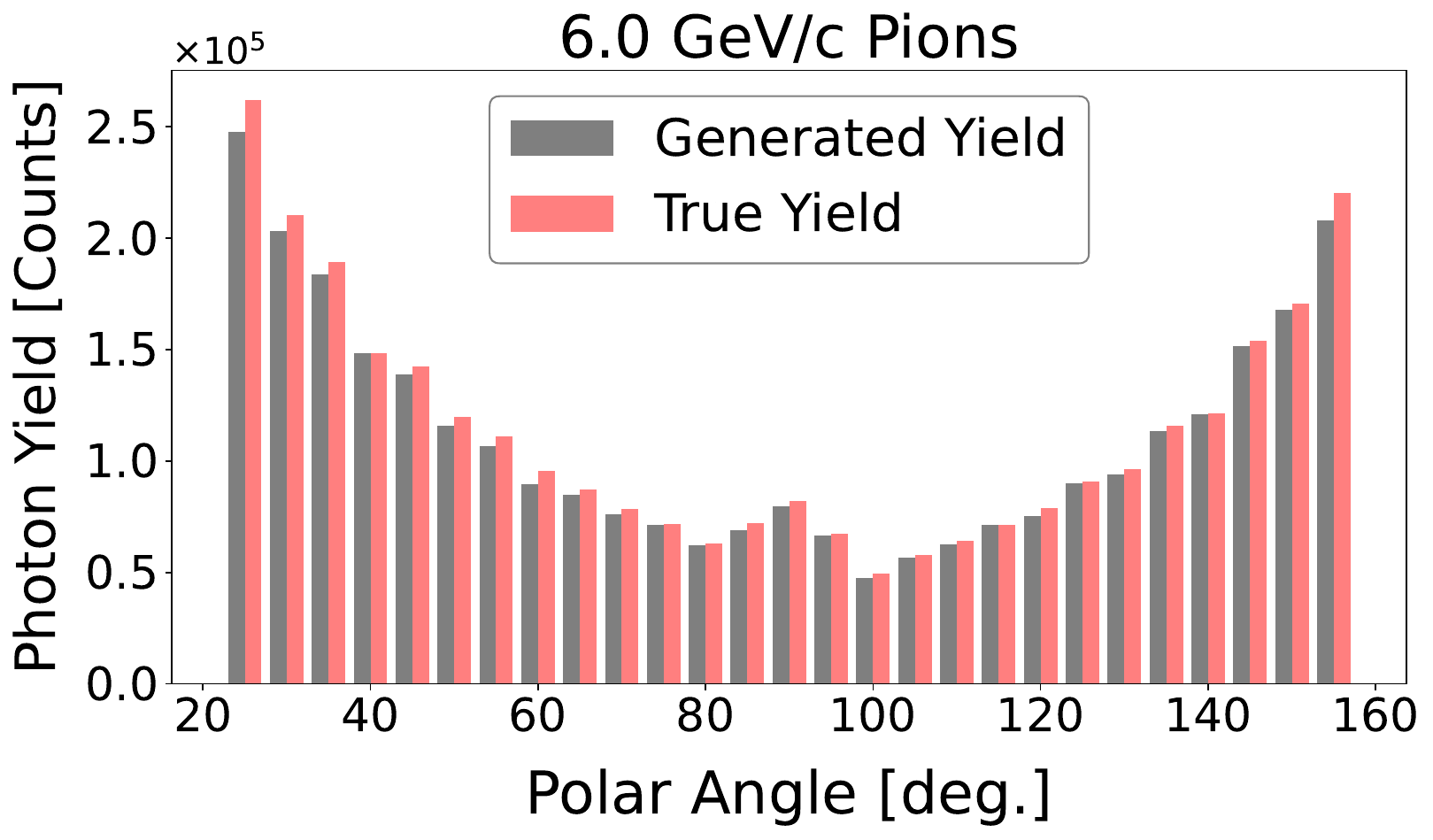}
        \caption{Pions - Two Experts}
    \end{subfigure} \\
      \begin{subfigure}[b]{0.49\textwidth}
        \centering
        \includegraphics[width=\textwidth]{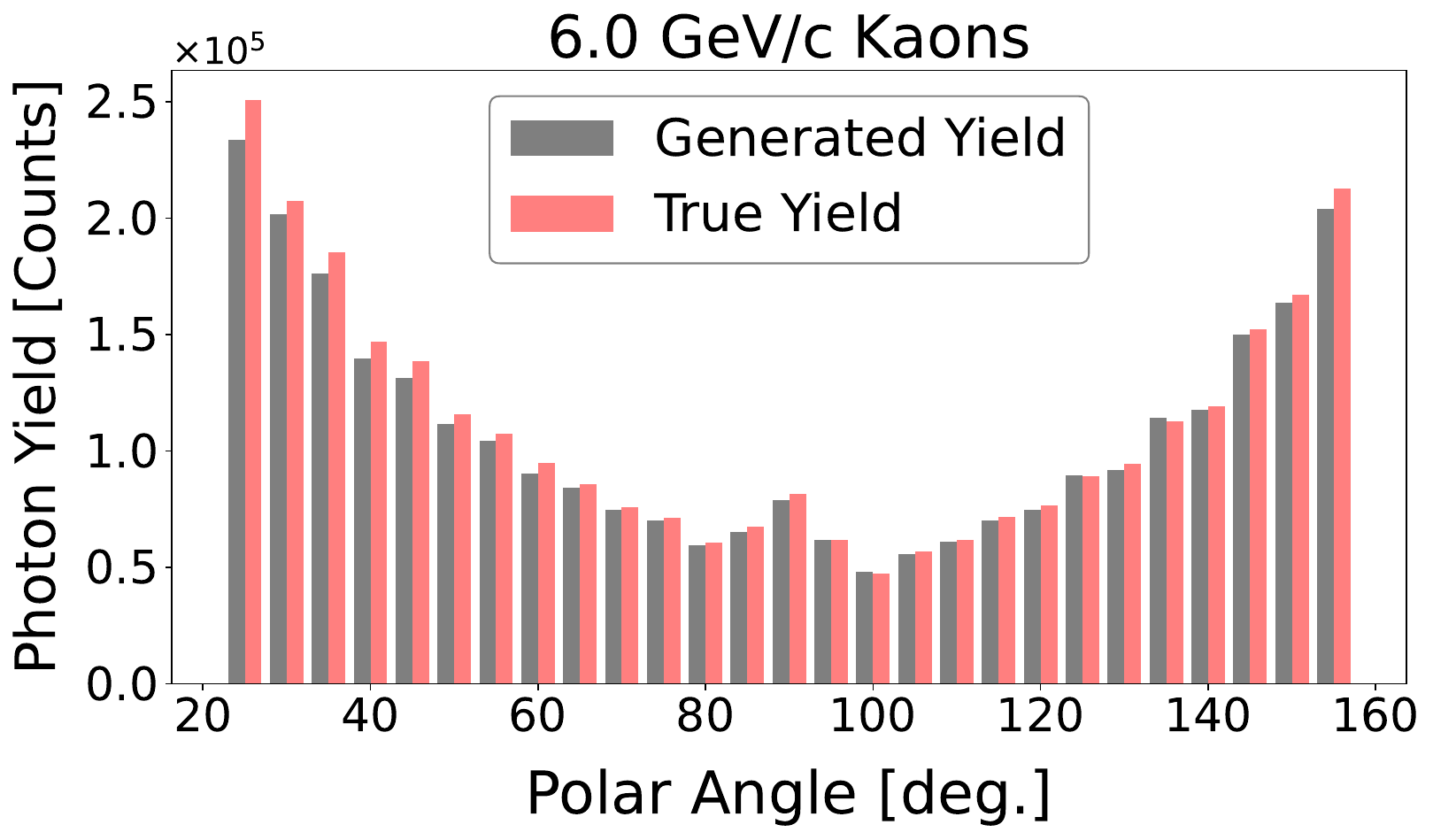}
        \caption{Kaons - Four Experts}
    \end{subfigure}
    \begin{subfigure}[b]{0.49\textwidth}
        \centering
        \includegraphics[width=\textwidth]{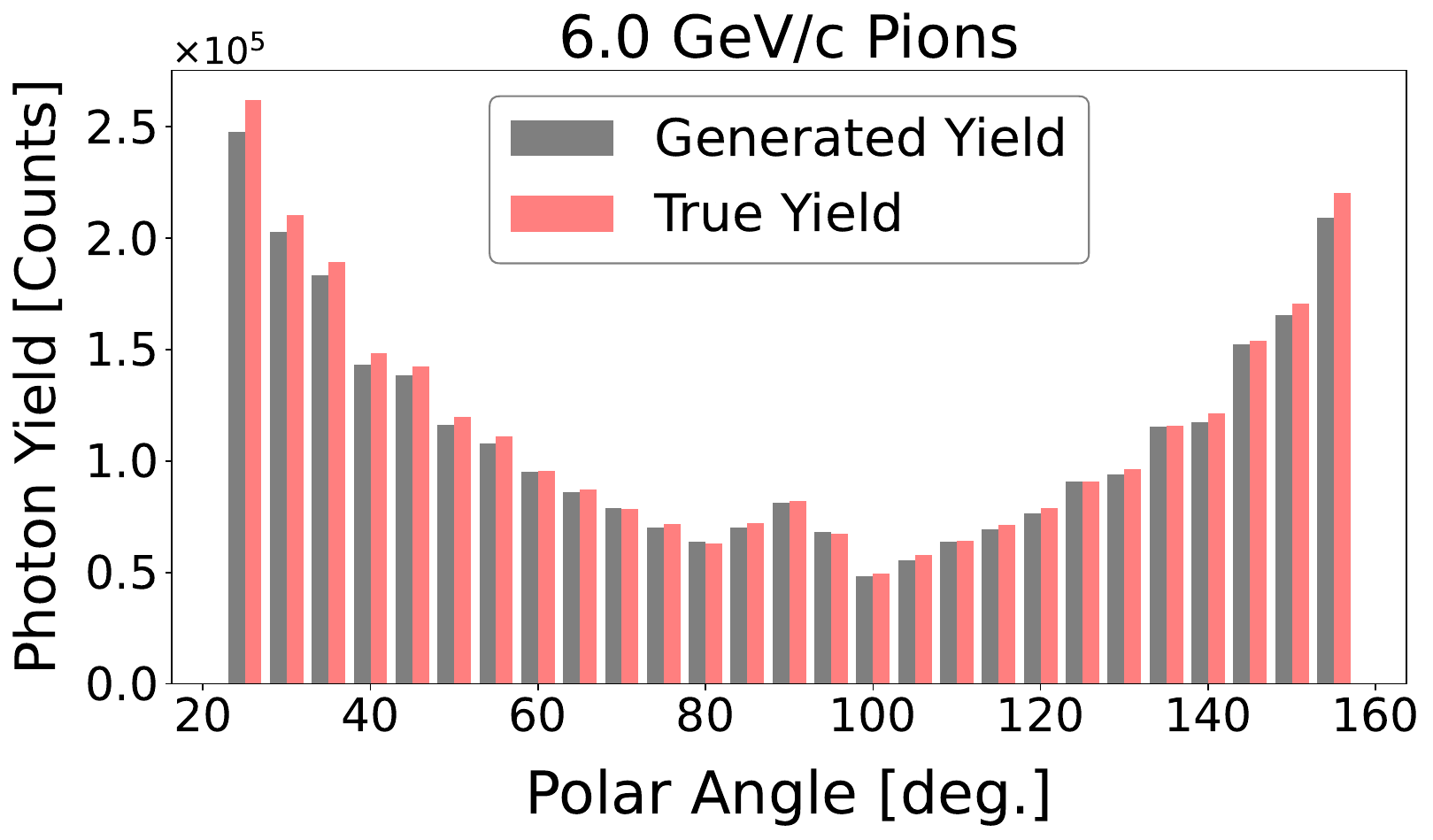}
        \caption{Pions - Four Experts}
    \end{subfigure} 
    \caption{\textbf{Photon Yield Comparison at 6 GeV/c:} Comparison of generated photon yield for kaons (left column) and pions (right column), using Nucleus sampling and fixed temperature for independent models (top row), a combined model with two experts (middle row) and a combined model with four experts (bottom row).}
    \label{fig:yields}
\end{figure}

We note exceptional agreement in the spatial dimensions, in which our ratios are $\sim$ 1 across all values. 
Our time distribution closely matches the ground truth over the main photon population, with most Cherenkov hits occurring within the first $\SI{100}{\nano\second}$. Discrepancies appear beyond $\SI{100}{\nano\second}$, where photon densities fall well below $\mathcal{O}(10^{-3})$; as a result, their impact on the overall distribution is negligible.
This is an expected outcome in the context of modern generative models in general.

Unlike previous fast simulation methods for Cherenkov detectors, in which post-hoc modeling of the photon yield must be employed, our method directly learns this variability in sequence length as a function of the phase-space. Fig. \ref{fig:yields} depicts a comparison between the generated and ground truth photon yields at 6 GeV/c, in $\SI{5}{\degree}$ bins of the polar angle for kaons (left column) and pions (right column) for independent models (top row), a mixture of two experts (middle row), and a mixture of 4 experts (bottom row).

We note good agreement for both generative models (pions and kaons) in terms of photon yield, with slight under estimation at lower values of the polar angle (\textit{e.g.}, $\theta < \SI{100}{\degree}$). These results are consistent at all regions of the phase-space as indicated by plots in \ref{app:3GeV} and \ref{app:9GeV}.

%\clearpage
%\newpage

\subsubsection*{KDE based Evaluation}

\begin{figure}[h]
    \centering
    \includegraphics[width=0.6\textwidth]{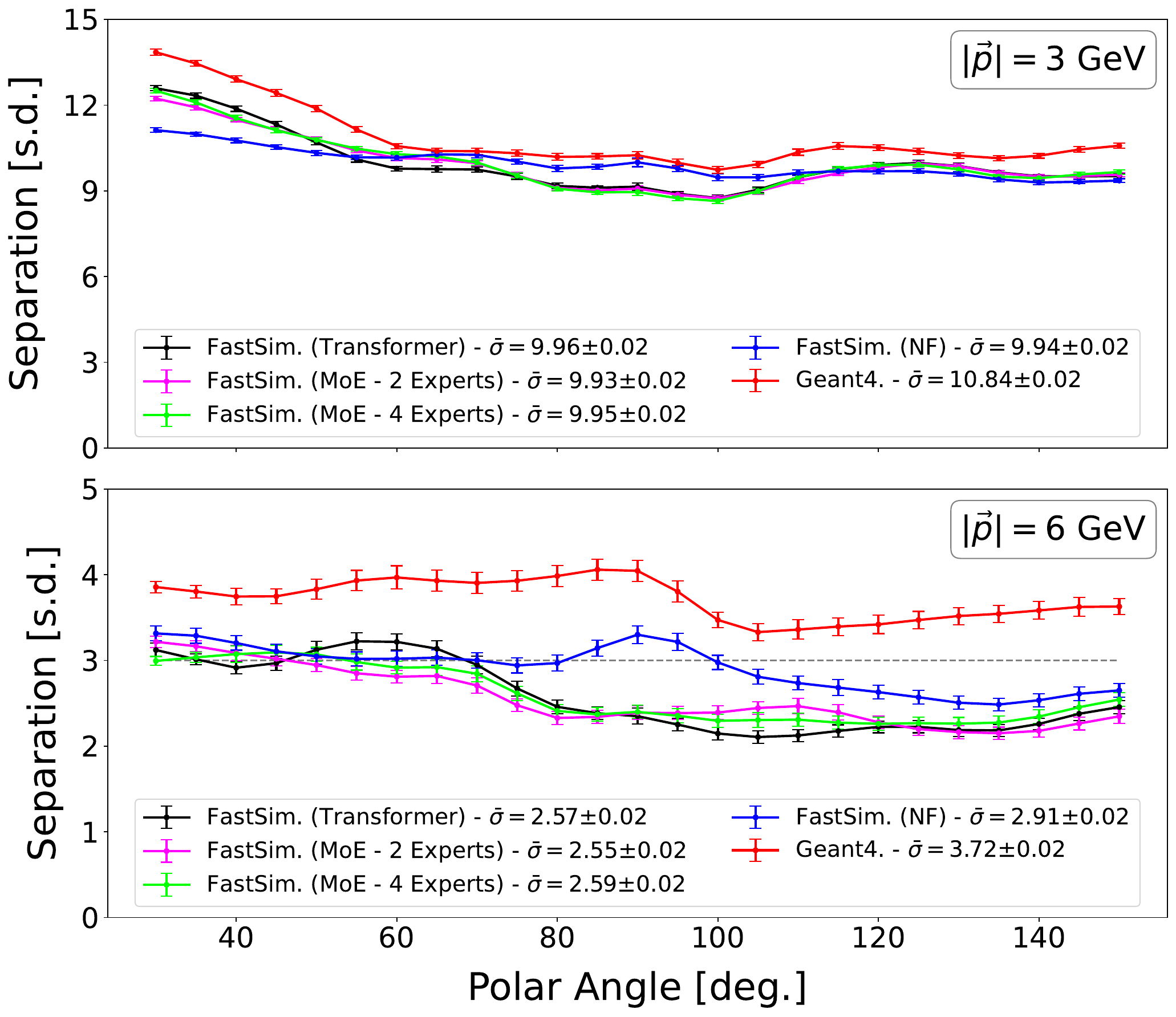}
    \caption{
    \textbf{Fidelity Studies:} The fidelity of the generated samples is assessed via pion–kaon separation power, using the FastDIRC KDE-based metric. Comparisons are shown between \textsc{Geant4} reference data and synthetic populations generated via fast simulation methods, including Normalizing Flows and our Transformer-based autoregressive approaches. Results are shown at 3~GeV/\textit{c} (top) and 6~GeV/\textit{c} (bottom) across a range of polar angles. The 3$\sigma$ PID requirement~\cite{khalek2022science} is indicated by the horizontal dashed line.
    }
    \label{fig:fastDIRC_performance}
\end{figure}

We again follow the methods devised in \cite{giroux2025generativemodelsfastsimulation}, translating fidelity measurements to a more interpretable metric of separation power using the KDE method of FastDIRC.\footnote{The exact metric formulation and associated methodology are described in~\cite{giroux2025generativemodelsfastsimulation}.
}
This evaluation is performed at fixed momenta—specifically 3 GeV/c and 6 GeV/c—to emphasize fidelity testing at specific kinematic points, which serves as a stringent stress test for our generative models trained over a fully continuous kinematic space, as motivated in the previous section.
Fig. \ref{fig:fastDIRC_performance} depicts the performance comparison for $\SI[per-mode=symbol]{3}{\giga\eVperc}$ (top) and $\SI[per-mode=symbol]{6}{\giga\eVperc}$ (bottom), for various values of the polar angle (in bins of $\SI{5}{\degree}$), with fixed sized reference populations of 800k, for \geant, Normalizing Flow (NF) \cite{giroux2025generativemodelsfastsimulation}, and our proposed methods of independent models, and models with two or four experts (indicated in legend).

From inspection of the figure, we initially see that our proposed method does degrade in performance in comparison to both \geant, and previously proposed methods such as NF. As pointed out in \cite{giroux2025generativemodelsfastsimulation}, any smoothing of the underlying probability distribution produced by our model will incur significant performance decreases in comparison to \geant given the ring structures only differ by a few pixels spatially at higher momentum (eventually converging at large enough momentum). 
Given the inherent uncertainty in our data as to which is the most optimal next hit, different sampling techniques possess the ability to greatly alter our generations---arguably more than other AR models operating in the traditional domain of language. In light of this, we observe that post-hoc optimization can be employed to further curate generated samples. While not specifically fine tuning weights, optimizing generation parameters (\textit{e.g.}, temperature, dynamic temperature, $p$ values, etc) can improve fidelity and is left for future studies.\footnote{We have implemented and tested a wide variety of sampling methods, see \url{https://github.com/wmdataphys/FM4DIRC} for more details.} We also intend to investigate increases in performance through the scaling potential of transformers, \textit{i.e.}, scaling both model and training data size has shown to lead to ``emergent'' characteristics in the space of modern FMs, potentially translating to improved performance in our case.
We also observe strong agreement between independently trained generative models and those employing class-conditional generation via the MoE framework, validating the ability to simulate both pions and kaons within a single architecture.

\subsubsection*{Geometric Effects}

\begin{figure}[h]
    \centering
      \begin{subfigure}[b]{0.43\textwidth}
        \centering
        \includegraphics[width=\textwidth]{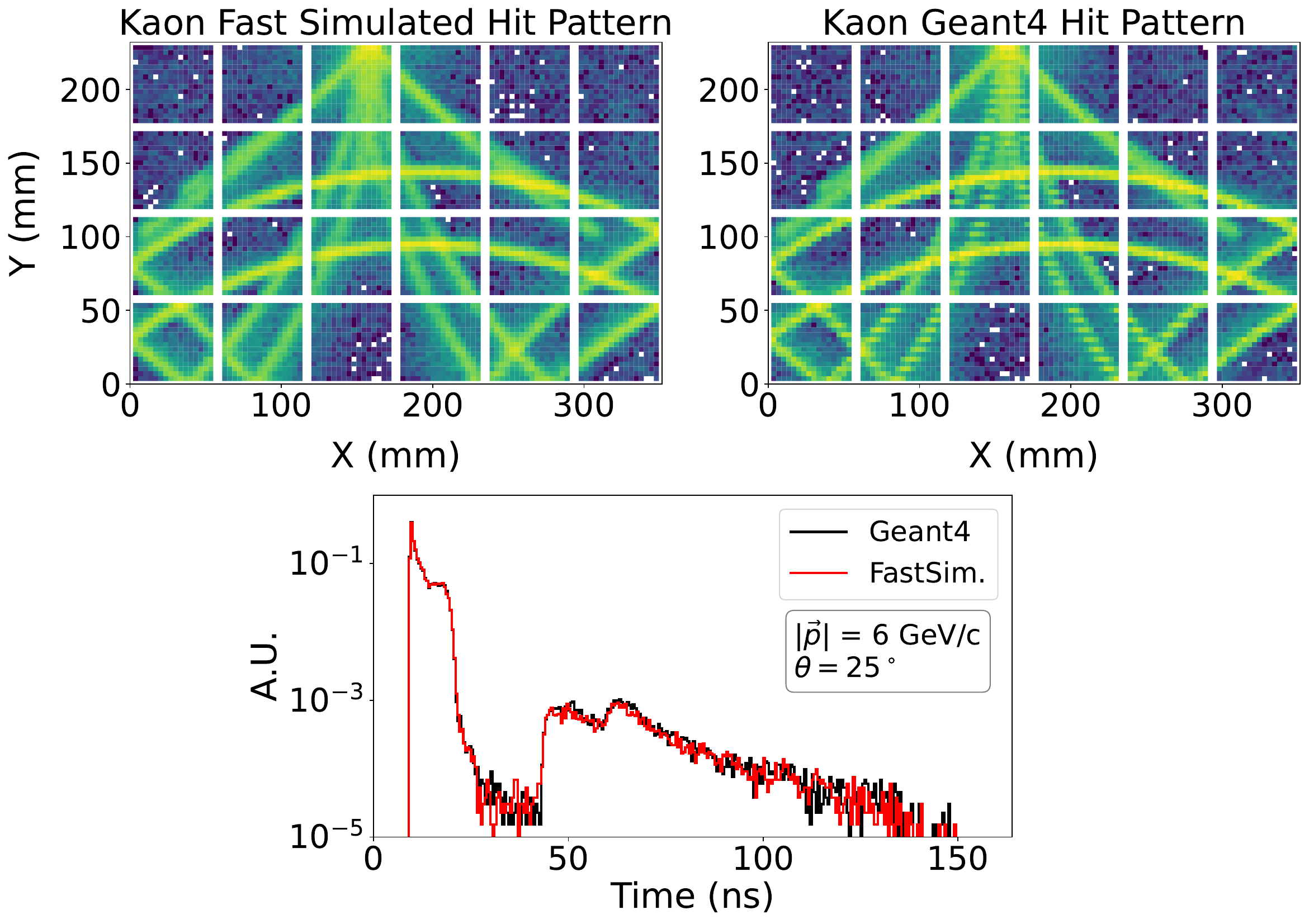}
        \caption{Normalizing Flows}
    \end{subfigure}
    \begin{subfigure}[b]{0.43\textwidth}
        \centering
        \includegraphics[width=\textwidth]{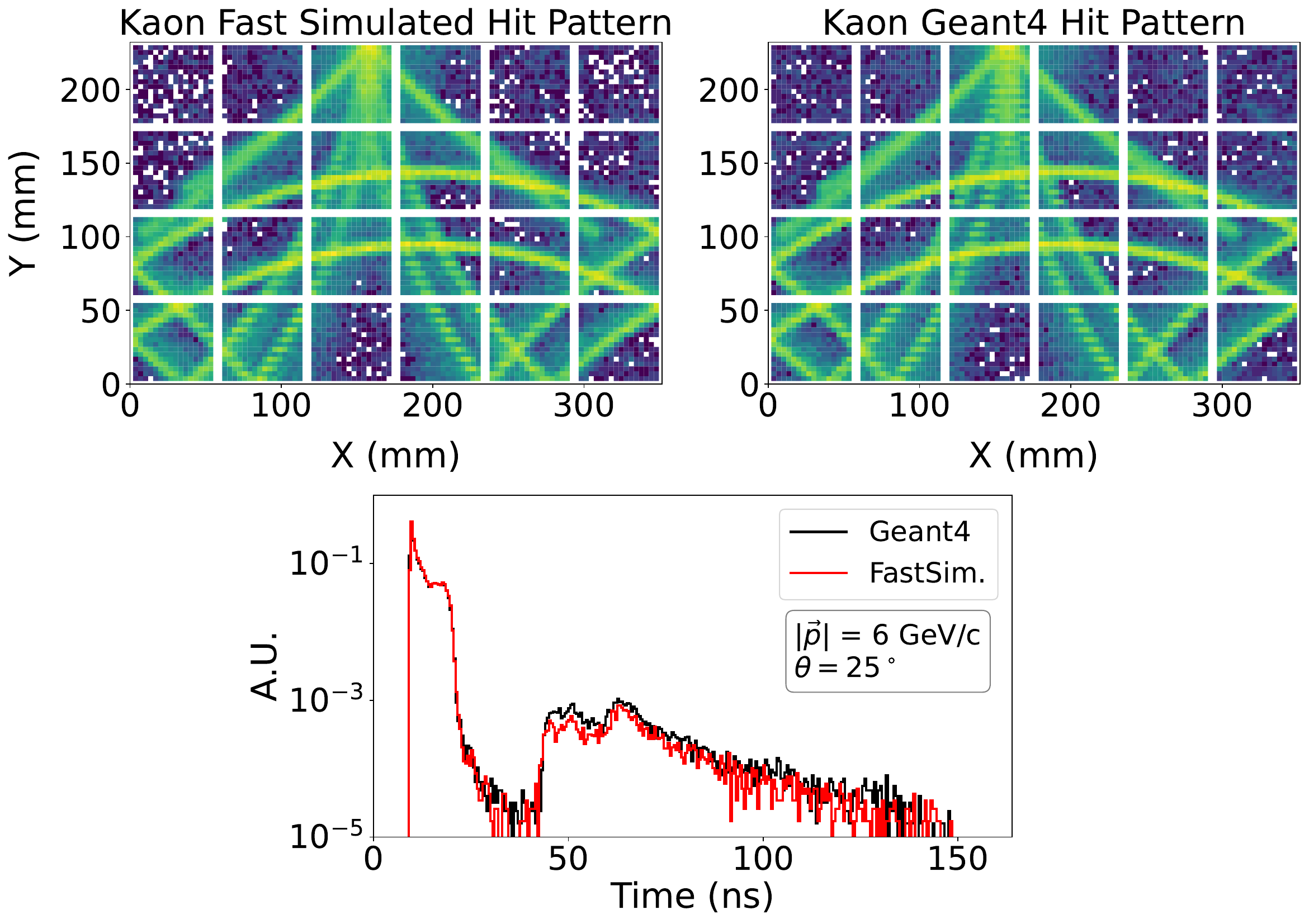}
        \caption{Autoregressive}
    \end{subfigure} 
    \caption{\textbf{Geometrical Effect Comparison at 6 GeV/c:} 
    Comparison of (a) Normalizing Flows and (b) our autoregressive model, with (b) more effectively capturing the kaleidoscopic effect.
    }
    \label{fig:geom_effects}
\end{figure}

As discussed earlier, the discrete nature of our model allows it to capture geometric effects—such as the kaleidoscopic effect—whose magnitude varies with polar angle and sampling method.
The kaleidoscopic effect manifests as pixel-level structure in the readout, with the effect being more pronounced the closer a track hits to the expansion volume. 
Fig. \ref{fig:geom_effects} depicts generations at the closest possible polar angle to the expansion volume ($\theta$ = \SI{25}{\degree}), for both (a) NF, and (b) our AR approach in the figure referenced as `Autoregressive'.

While promising, further research into the capacity of our model is required to fully capture this effect at all regions of the phase-space.

\subsection*{Downstream Tasks and Fine-tuning}

As shown in previous works \cite{birk2024omnijet,mikuni2024omnilearn}, the flexible structure of transformer backbones supports various downstream tasks (primarily fast simulation and reconstruction tasks such as PID). 
In the following, we evaluate our model's performance after fine-tuning the backbone architecture—consisting primarily of transformer blocks—from a generative task to sequence-level classification.
We also introduce another downstream task in the form of noise filtering---a potentially crucial application in high rate environments at the future EIC (\textit{e.g.}, these studies can be of particular relevance for a sub-detector such as the dual-RICH in ePIC, where SiPM dark current rates can be as high as $\mathcal{O}(100)$kHz/channel after years of radiation damage \cite{ePICDAQ2024}). 
In this case, we show some inherent limitations of fine-tuning and discuss the results in the context of prior objectives. In all cases, the models are trained under identical conditions (\textit{e.g.}, same learning rates, optimizers, etc.), with the only difference being in their initialization strategies.

\subsubsection*{Particle Identification}

\begin{figure}[!]
    \centering
      \begin{subfigure}[b]{0.49\textwidth}
        \centering
        \raisebox{1.25cm}{
        \includegraphics[width=\textwidth]{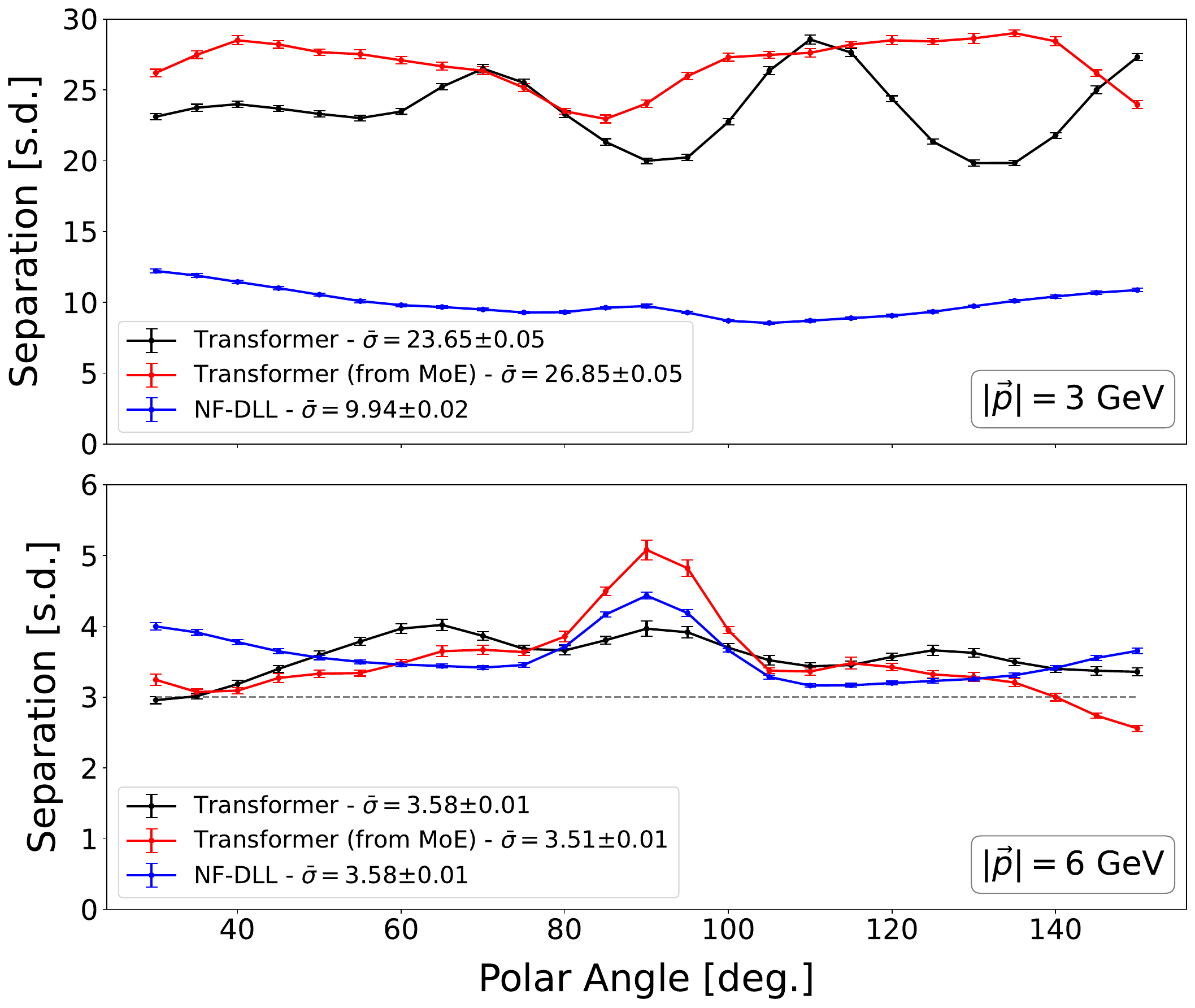}}
        \caption{Separation Power}
    \end{subfigure}
    \begin{subfigure}[b]{0.49\textwidth}
        \centering
        \includegraphics[width=\textwidth]{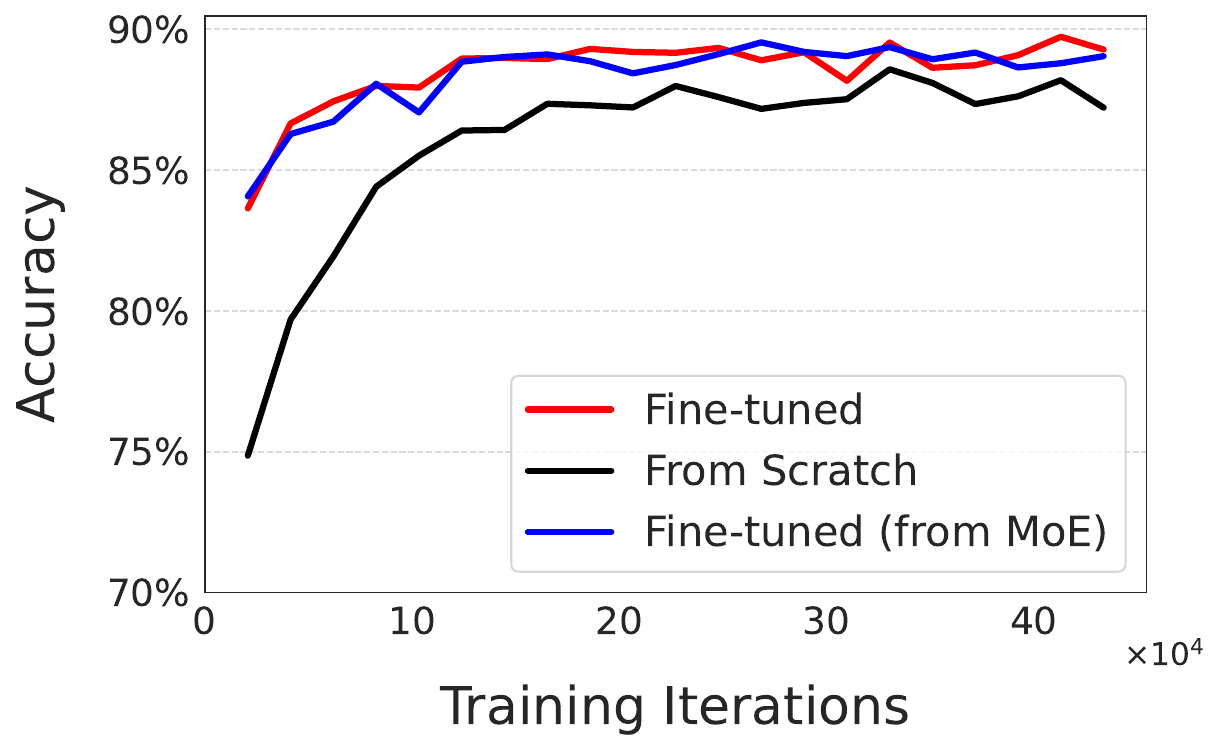} \\
        \includegraphics[width=\textwidth]{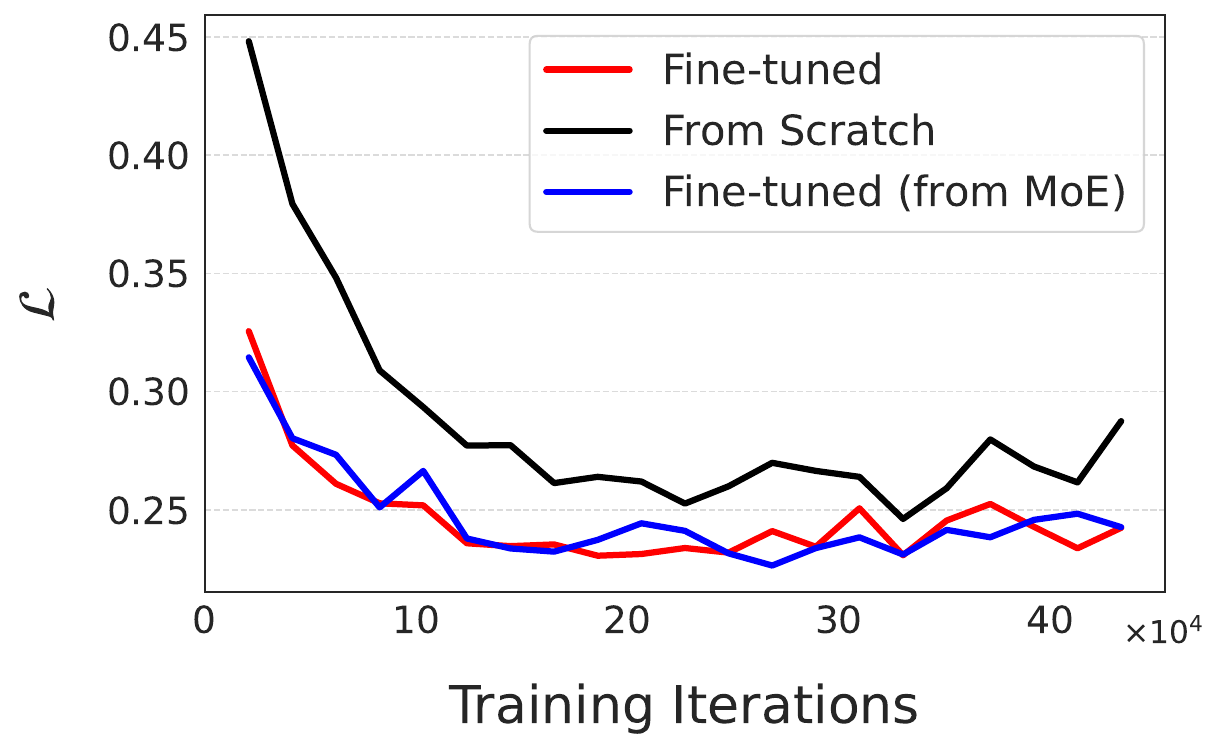}
        \caption{Fine Tuning}
    \end{subfigure} 
    \caption{\textbf{Particle Identification Performance Comparison:} PID performance of pions and kaons using (a) Normalizing Flow Delta-Loglikelihood (NF-DLL) method, and our method at 3 GeV/c (top) and 6 GeV/c (bottom) for various values of the polar angle, and (b) comparison of accuracy versus training iterations for models trained from scratch, and fine-tuned integrated over the entire phase-space. \textit{Fine-tuned} models are initialized with weights from a separately trained generative model (e.g., trained only on pions or kaons), while \textit{from MoE} models are initialized using weights from a multi-expert architecture with four experts.}
    \label{fig:classification_performance}
\end{figure}

We classify if charged tracks (sequence level) originate from pions or kaons within our detector, and show that fine-tuning our models from generation to classification provides inherent speed ups in convergence, along with slight performance increases. 
In the case of class conditional models using a MoE, we show that fine-tuning still remains valid in which we initialize the weights of the standard FFNNs as the average of the experts. While it is possible to directly use a MoE within models performing reconstruction tasks, our initial performance estimates show that the additional computation is not needed and this is left for future studies. Fig. \ref{fig:classification_performance} (a) shows the performance of our model in terms of separation power (the number of standard deviations separating the two latent distributions) at 3 Gev/c (top), and 6 Gev/c (bottom) at various values of the polar angle. We also provide performance comparisons using the NF models from \cite{giroux2025generativemodelsfastsimulation}, and the DLL method devised in \cite{fanelli2024deep}. The average separation power over the polar angle is reported in the legends. Fig. \ref{fig:classification_performance} (b) shows the accuracy of our model integrated over the phase-space as a function of training iteration (number of batches), for both fine-tuned and models trained from scratch. \textit{Fine-tuned} refers to models initialized with the weights of an independently trained generative model (\textit{e.g.}, one trained solely on pions or kaons), while \textit{from MoE} refers to models initialized from a multi-expert (four experts) architecture.

From inspection of Fig. \ref{fig:classification_performance} (a) we conclude that both NF, and our FM are able to surpass  required separation power of $3\sigma$ at $\sim \SI[per-mode=symbol]{6}{\giga\eVperc}$, as outlined in the EIC Yellow Report \cite{khalek2022science}. We also note the unique differences in their performance, in which our FM is more heavily impacted by changes in polar angle as depicted by higher non-uniformity (\textit{e.g.}, see separation power at 3 GeV/c). A potential artifact of model configuration, or perhaps method of kinematic conditioning that will be further investigated in future studies. 
With inspection of Fig. \ref{fig:classification_performance} (b) we show our model is capable of fine-tuning, it benefits in terms of both computational efficiency (rate of convergence), and slightly increases in accuracy. We also note the algorithm's clear ability to separate charged pion and kaons at 3 Gev/c, in which our separation power essentially doubles in contrast to NF. While impressive, the usefulness of such separation power at low momentum is debatable and preferred elsewhere at higher momentum where the particle identification becomes more challenging. In future studies we aim to shift some of the focus the model puts on this momentum region through momentum weighted training schemes, or trainings limited to higher impact kinematic ranges to allow increased performance.

\subsubsection*{Noise Filtering}

\begin{figure}[h]
    \centering
    \begin{subfigure}[b]{0.49\textwidth}
    \includegraphics[width=0.49\textwidth]{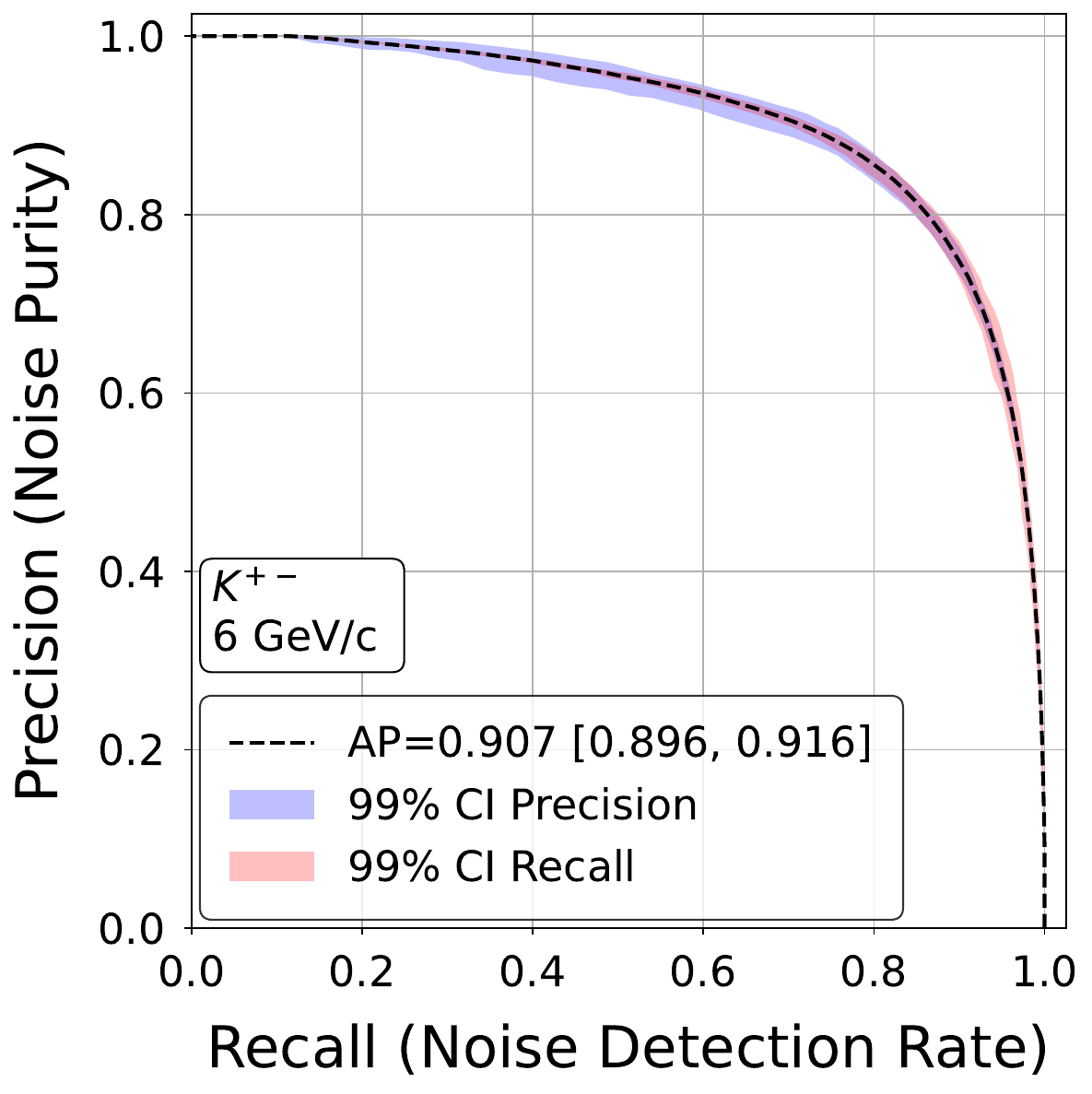}% 
    \includegraphics[width=0.49\textwidth]{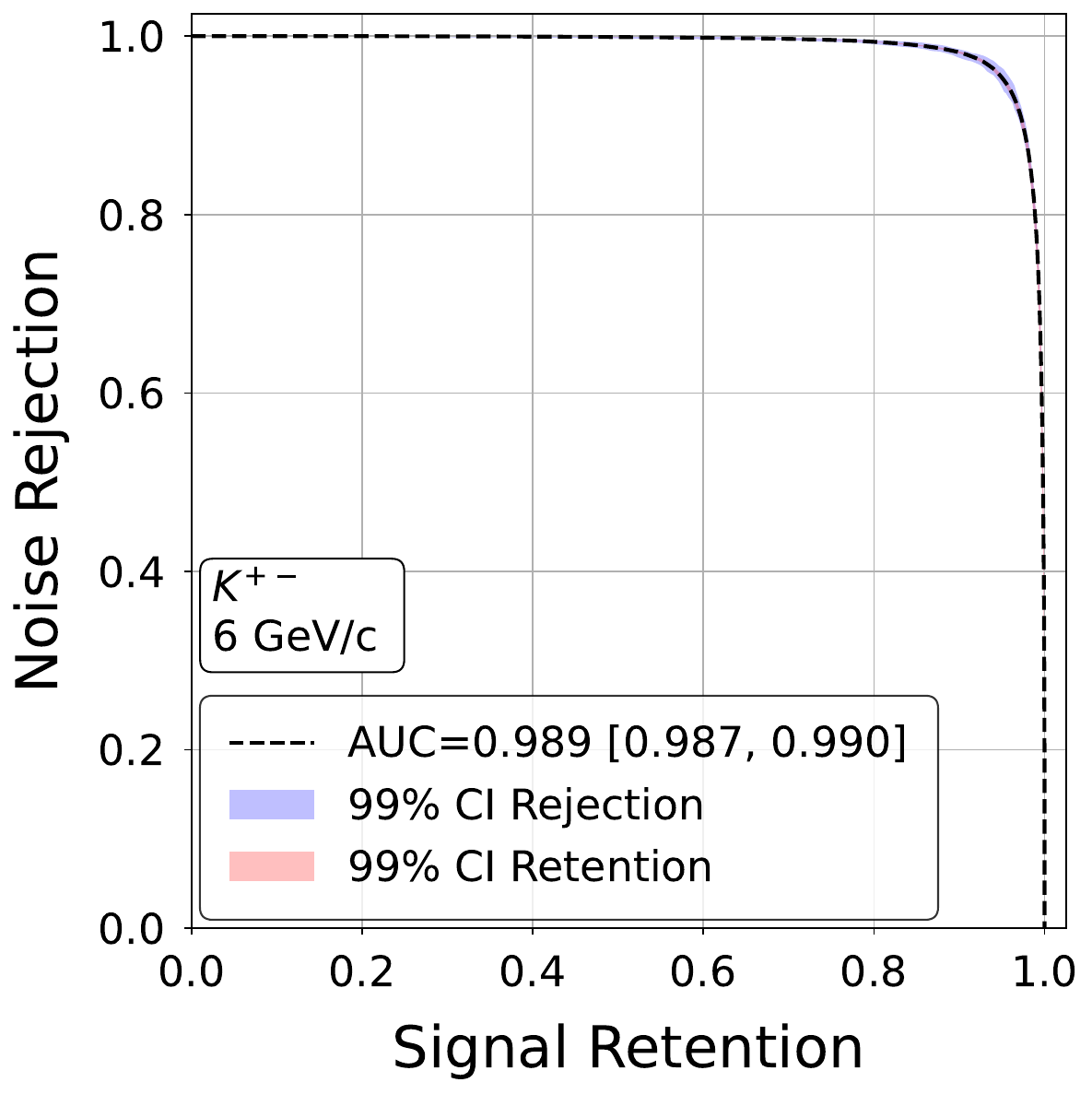} %
    \caption{Kaons}
    \end{subfigure} 
    \begin{subfigure}[b]{0.49\textwidth}
    \includegraphics[width=0.49\textwidth]{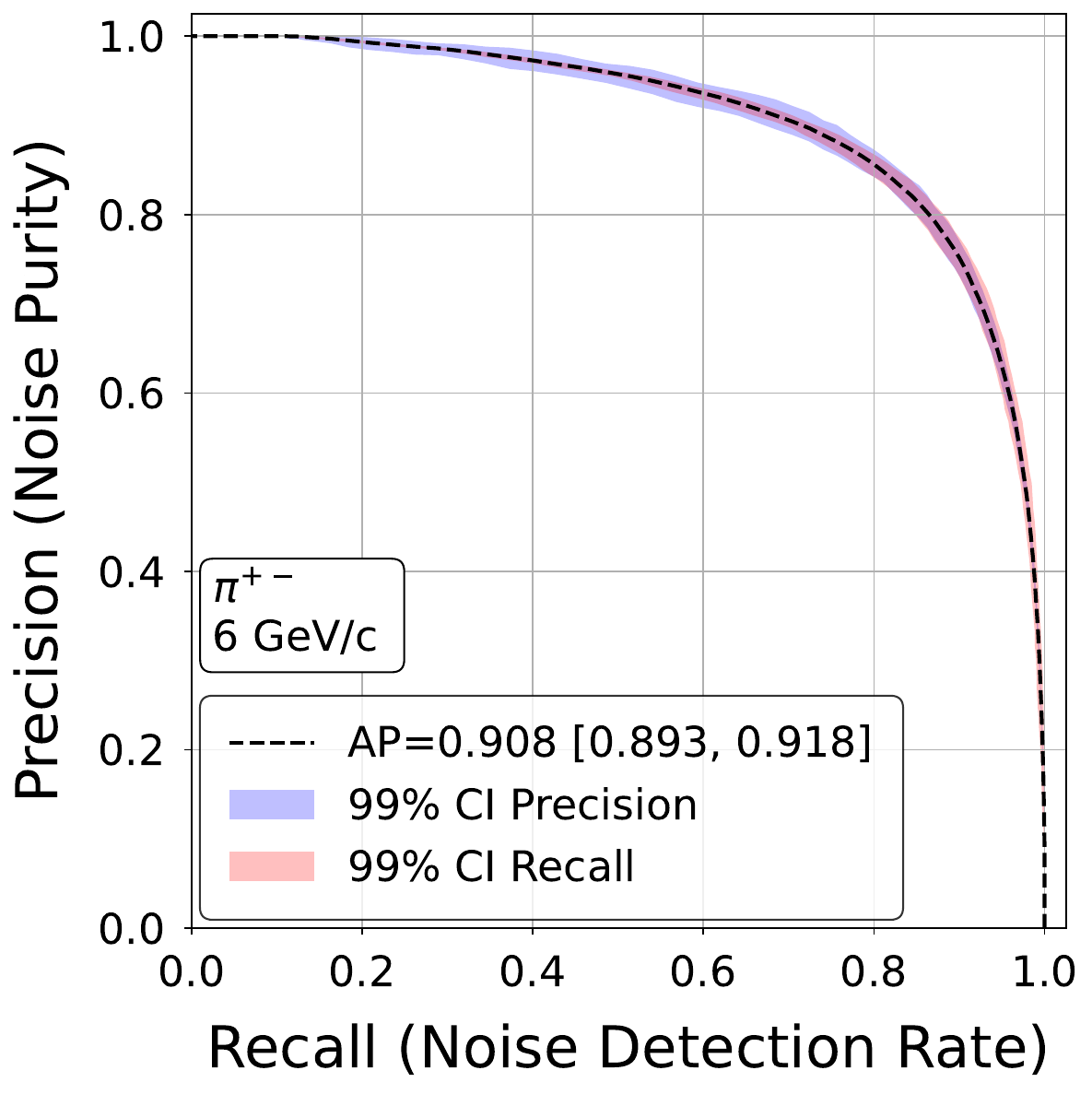}% \\
    \includegraphics[width=0.49\textwidth]{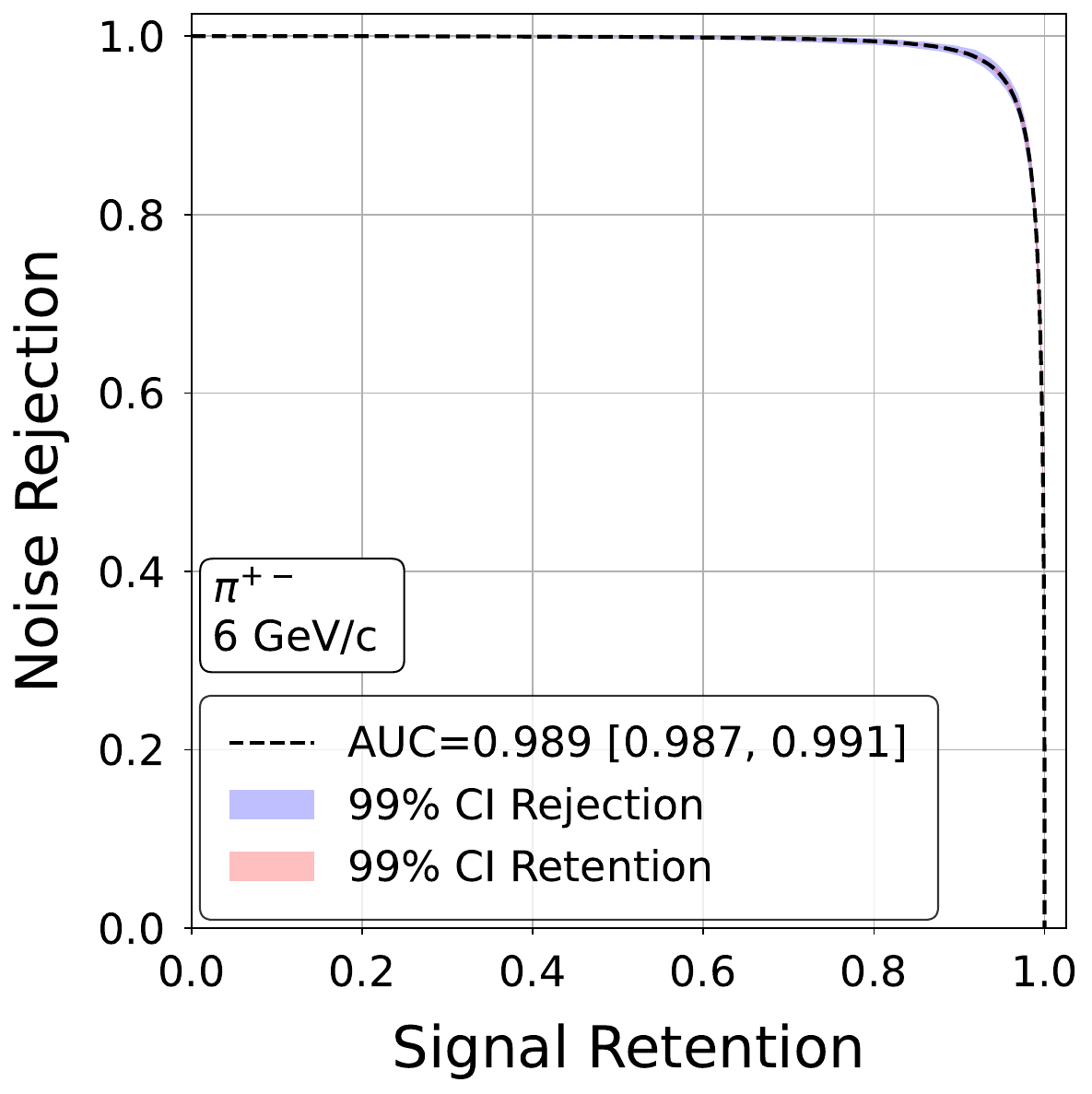} 
    \caption{Pions}
    \end{subfigure}
    \caption{
    \textbf{Noise Filtering at 6 GeV/c:} Noise filtering performance of our method at 6~GeV for (a) kaons and (b) pions, shown as precision-recall curves and noise rejection versus signal efficiency. The precision-recall curves are insensitive to class imbalance. The dark rate contributes approximately $8\text{--}10\%$ of total hits as noise.}
    \label{fig:Filtering_6GeV}
\end{figure}

We classify if individual hits (token level) created by an incoming charged track are signal or noise. The model is trained irrespective of class label, \textit{i.e.}, both pions and kaons are included under the same model. We use a simulated dark rate in the PMTs of $\SI{100}{\kilo\hertz}/\text{cm}^2$ in which we sample randomly during training. For a given track, both the configuration (ordering in time), and the number of dark photons is stochastic, forcing our architecture to learn contextual representations of noise and signal as opposed to relying on positional information. During evaluation, we perform the same sampling procedure in which we create fifty unique representation of each track in the testing set to obtain uncertainty estimates. Fig. \ref{fig:Filtering_6GeV} shows the performance of our method for (a) kaons and (b) pions in terms of precision-recall curves, and noise rejection as a function of signal efficiency. The former being agnostic to the inherent class imbalance.\footnote{Our configuration of dark rate provides approximately $8-10\%$ of total hits as noise.}  

From inspection of the precision-recall curves in Fig. \ref{fig:Filtering_6GeV}, we see our model is able to efficiently able to characterize detector noise from that of signal at the token level, noting an average precision (AP) of $\sim 90\%$ for both pions and kaons. We also note exceptional performance in terms of noise rejection as a function of signal retention, in which the area under the curve (AUC) is $\sim 99\%$.
Plots for additional kinematics (\textit{e.g.}, 3 GeV/c and 9 GeV/c) can be found in \ref{app:filtering} in which we see uniform performance at both lower and high momenta.

\begin{figure}[h]
    \centering
    \includegraphics[width=0.41\textwidth]{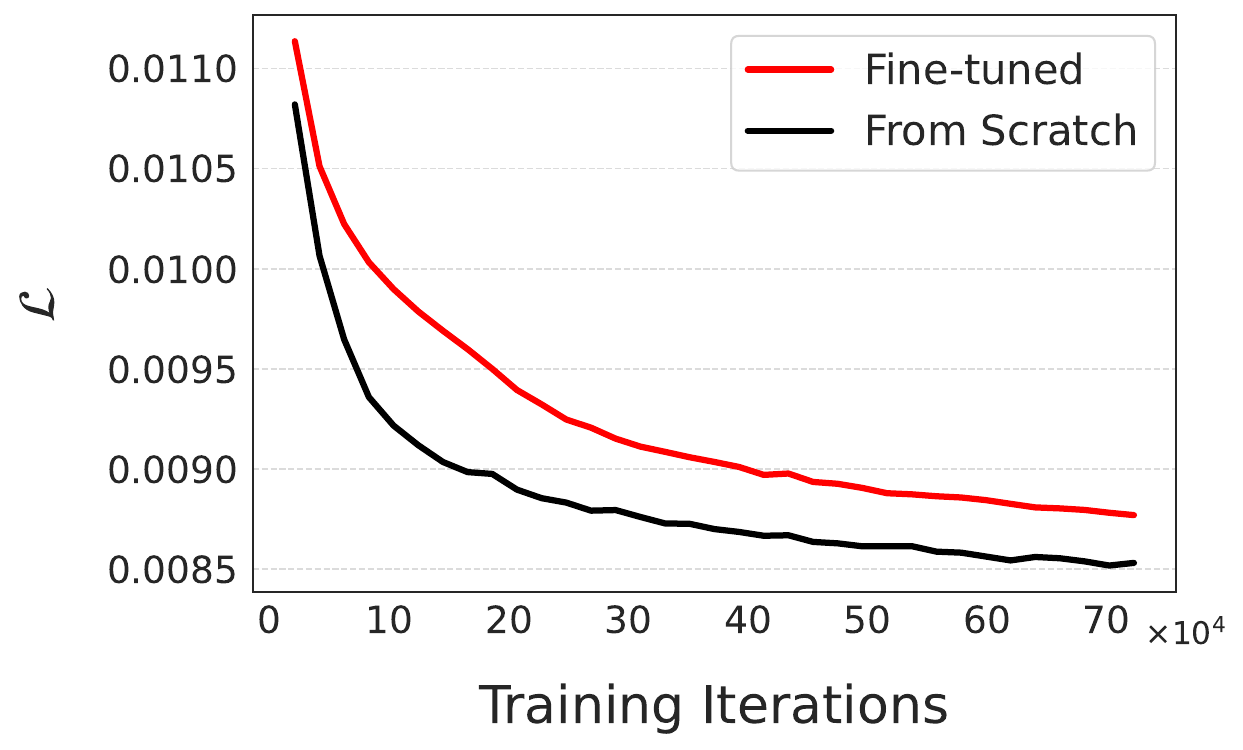}%
    \hfill
    \includegraphics[width=0.59\textwidth,height=3.76cm]{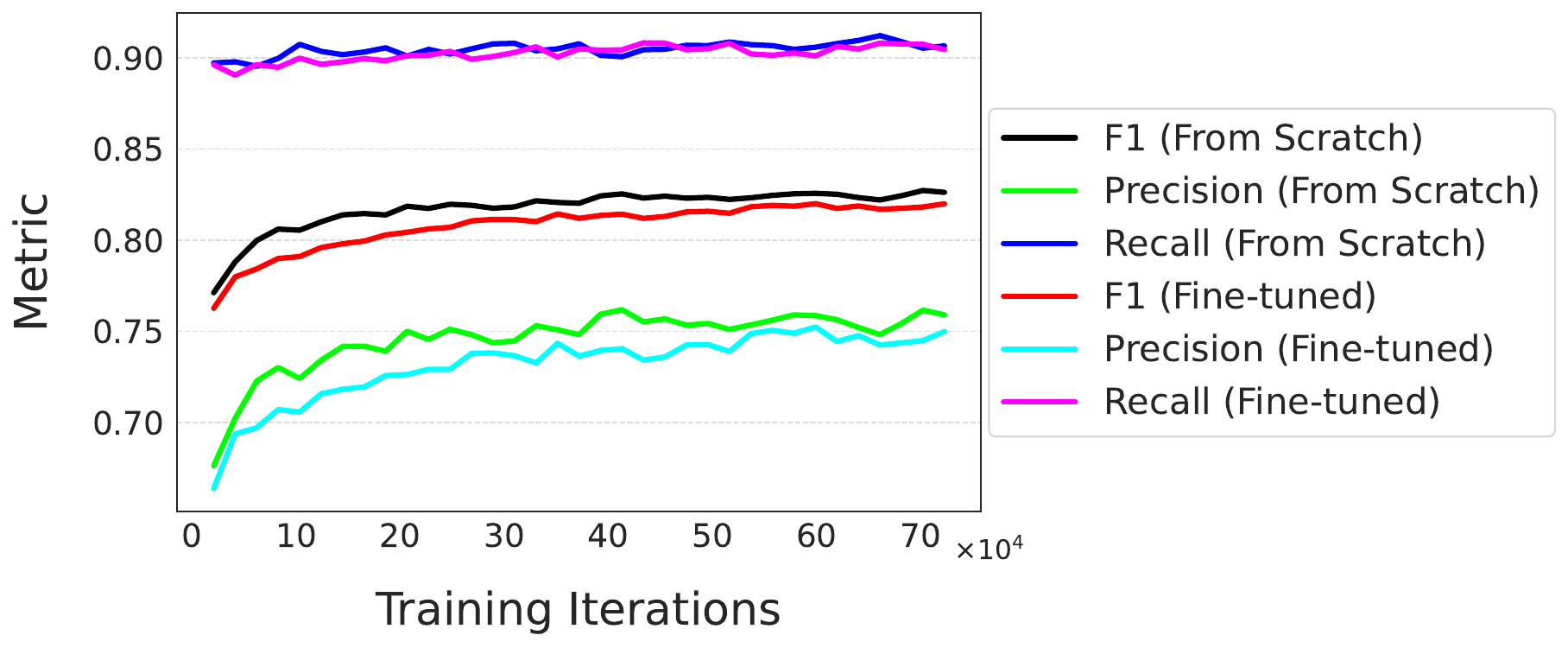}
    \caption{\textbf{Noise Filtering Fine-Tuning Comparison:} Validation loss as a function of training iterations for a model trained from scratch and fine-tuned from an independent generative model (left), and the corresponding validation metrics, precision, recall, and F1 score across training iterations for both approaches (right).}
    \label{fig:filter_finetune}
\end{figure}

As mentioned prior, there exists limitations of fine-tuning the pre-existing models to downstream tasks. The main influence in the effectiveness of fine-tuning is the degree of contextual similarity between the original task and the target task. For example, in generative modeling, transformer blocks typically learn a more global context to generate coherent outputs across an entire sequence. This global understanding is highly beneficial for both generation and sequence level classification tasks but can pose challenges when adapting the same model for token level classification.
Token-level classification requires a shift in focus from global to local context, where the model needs to make decisions based on finer, more immediate surroundings of each token rather than broad sequence-wide patterns. 
During fine-tuning for classification, the model is required to partially ``unlearn'' previously encoded information and instead realign it's attention mechanisms. This contextual realignment is non-trivial and can limit the efficiency or performance of the model on the downstream task. Fig. \ref{fig:filter_finetune} (left) shows the validation loss as a function of training iteration for both models trained from scratch and a fine-tuned generative model.\footnote{We fine-tune from an independent generative model only, and not from a model using a MoE given the inherent limitations seen.} Fig. \ref{fig:filter_finetune} (right) shows the validation metrics, namely the precision, recall and F1 score for both fine-tuned and models trained from scratch as a function of training iteration.

From inspection of Fig. \ref{fig:filter_finetune} it is apparent that fine-tuning poses no benefits under the task of filtering. While not detrimental in terms of performance (metrics remain similar in value), the model trained from scratch is more optimal---in which we have identified a clear limitation. In light of this, identifying task dependent behaviors is crucial as FMs become increasingly popular within the physics community. In certain scenarios, fine-tuning may fail to fully exploit the representational capacity of the pre-trained model, thereby limiting potential performance improvements.

%\input{5_impacts}

%\clearpage

\section{Summary and Conclusions}\label{sec:summary}

We present a (proto) foundation model designed to process low-level detector inputs (\textit{e.g.}, pixelated readout systems), with demonstrated use in nuclear physics contexts and potential relevance to other particle detection environments.
Through a series of closure tests, we have demonstrated that our model can perform both generative and reconstruction objectives, leveraging fine-tuning to improve computational efficiency for targeted applications.

In the case of generative modeling, our discretized approach in space allows potential for capturing low-level geometric effects seen in DIRC detectors, an advancement over prior methods \cite{giroux2025generativemodelsfastsimulation}. Moreover, we have introduced a method in which we potentially alleviate the need for excess quantization over continuous domains, allowing the physical sensor resolutions of detectors to be respected through adjacent vocabularies. Our method fuses spatial and temporal information through Causal Multi-Head Cross Attention, yielding temporally aware spatial embeddings that are subsequently processed using standard transformer blocks. 
We further show that multiple classes (\textit{e.g.}, pions and kaons) can be contained within a singular generative model using a Mixture of Experts without loss of granularity in comparison to multiple independent models. This is a key advancement over previous methods in which independent models were used to prevent over mixing of modes as momentum increases---crucial for particle identification applications where the probability density function differs marginally at the hit level.
We believe this architecture is both flexible and broadly applicable to a variety of detector systems, including calorimeters. Given the inherent stochasticity in physical readout systems, the concept of a many-to-many mapping approximated through split vocabularies may offer promising results. 
We have also demonstrated that our model achieves the desired separation power ($3\sigma$ at 6 GeV/c) for charged pion and kaon identification, with performance comparable to or exceeding that of other Deep Learning methods \cite{fanelli2024deep}. Interestingly, our foundation model exhibits enhanced sensitivity to variations in polar angle, potentially reflecting its internal representation of detector geometry or the influence of kinematic conditioning—an effect that will be investigated in future work.
Additionally, we find that fine-tuning not only accelerates convergence but also yields slight gains in classification accuracy. Notably, the model demonstrates exceptionally high distinguishing power at lower momenta---though the practical importance of such performance may be limited. To better align model focus with experimental priorities, future studies will explore momentum weighted, or limited phase-space training strategies that emphasize high-impact kinematic regions.

Finally, we have demonstrated that our architecture can also be applied to noise filtering—a critical task for high-noise sub-detector systems at the future EIC-via token-level classification. In this context, fine-tuning offers limited benefit, as the model must effectively ``unlearn'' global context in favor of localized information.
While the difference in performance between a model trained from scratch and one that is fine-tuned is marginal, we have identified potential limitations of fine-tunability within reconstruction tasks. 

Despite the encouraging performance observed, further refinement of our method is necessary to enable higher-fidelity simulation and improved classification performance, particularly at higher momentum. As seen in Large Language Models, emergent capabilities often arise as a function of scale \cite{wei2022emergent}. Accordingly, we intend to investigate the scalability of our approach through detailed ablation studies focused on both model capacity and the volume of training data.

Lastly, we aim to assess the model's ability to fine-tune across experiments, with particular interest in the \gluex DIRC \cite{stevens2016gluex} and the hpDIRC at the future EIC \cite{kalicy2024high}. While differences in production mechanisms and detector geometries exist between \gluex and the EIC, we hypothesize that the underlying physics governing DIRC detectors remains sufficiently invariant to enable effective cross-experiment fine-tuning.

\section*{Code Availability}
The code is publicly available at \href{https://github.com/wmdataphys/FM4DIRC}{https://github.com/wmdataphys/FM4DIRC}.

%\clearpage

\section*{Acknowledgments}
%
%We thank William \& Mary for supporting the work of JG and CF through CF's start-up funding. 
%
The authors acknowledge William \& Mary  Research Computing for providing computational resources and technical support that have contributed to the results reported within this article.
%

%\clearpage

\section*{References}
\bibliographystyle{iopart-num}
\bibliography{biblio}

\clearpage
%\newpage
\appendix

%%%%%%%%%%% 3 GeV %%%%%%%%%%%%%%%%%%%%%
\section{Evaluation at $\SI[per-mode=symbol]{3}{\giga\eVperc}$} \label{app:3GeV}

\begin{figure}[h]
    \centering
    \includegraphics[width=0.49\textwidth]{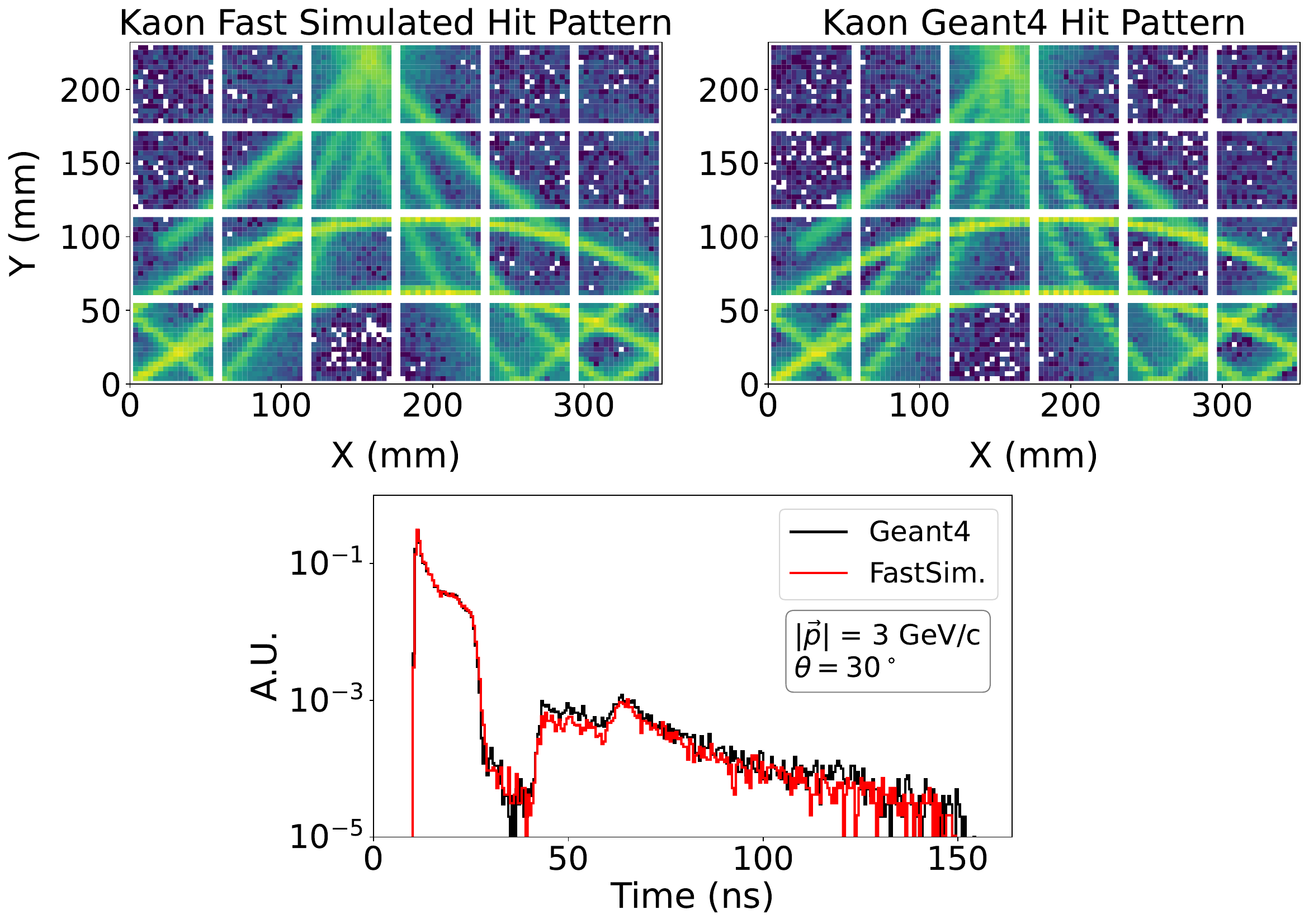}% 
   \includegraphics[width=0.49\textwidth]{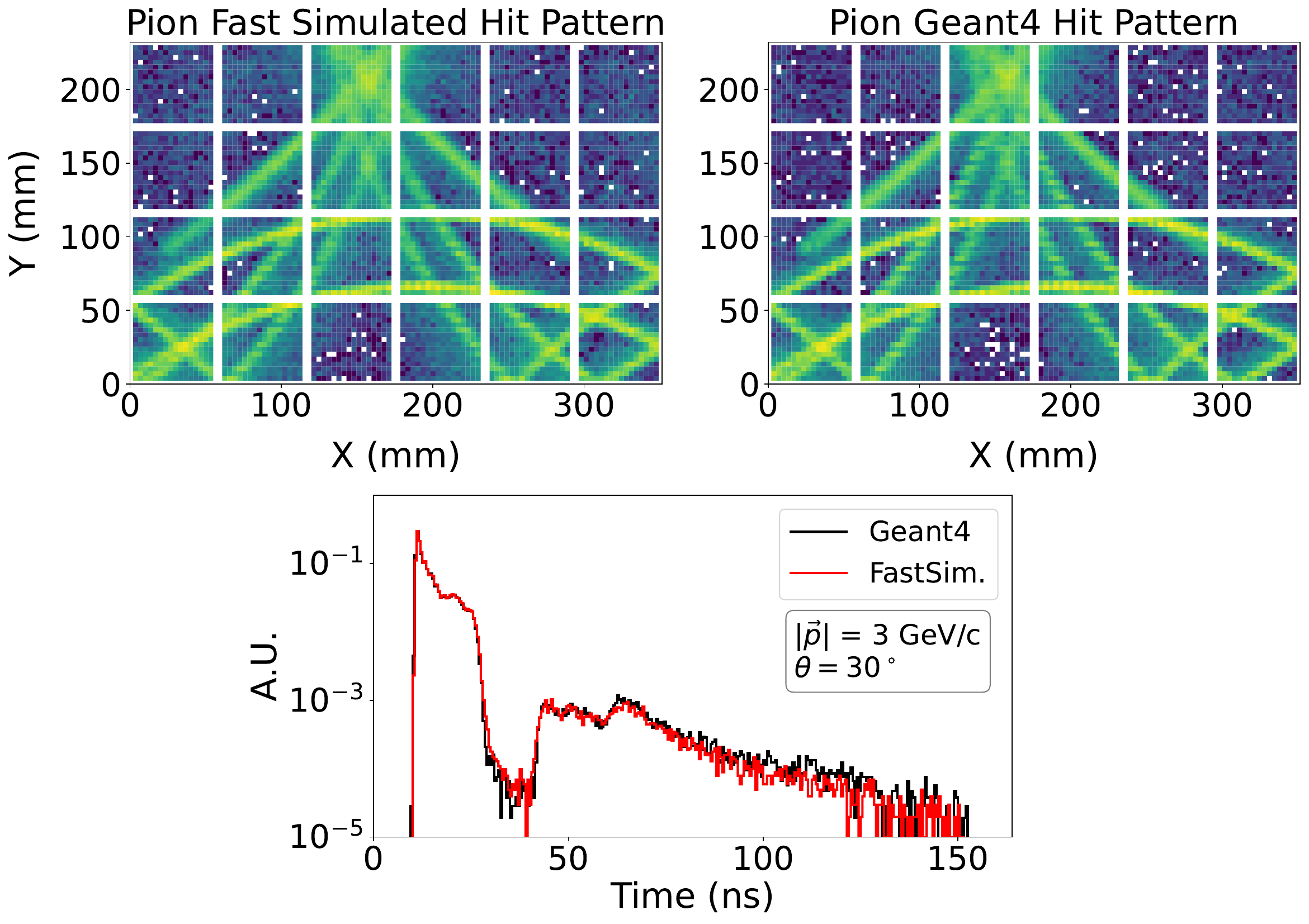} \\
    \includegraphics[width=0.49\textwidth]{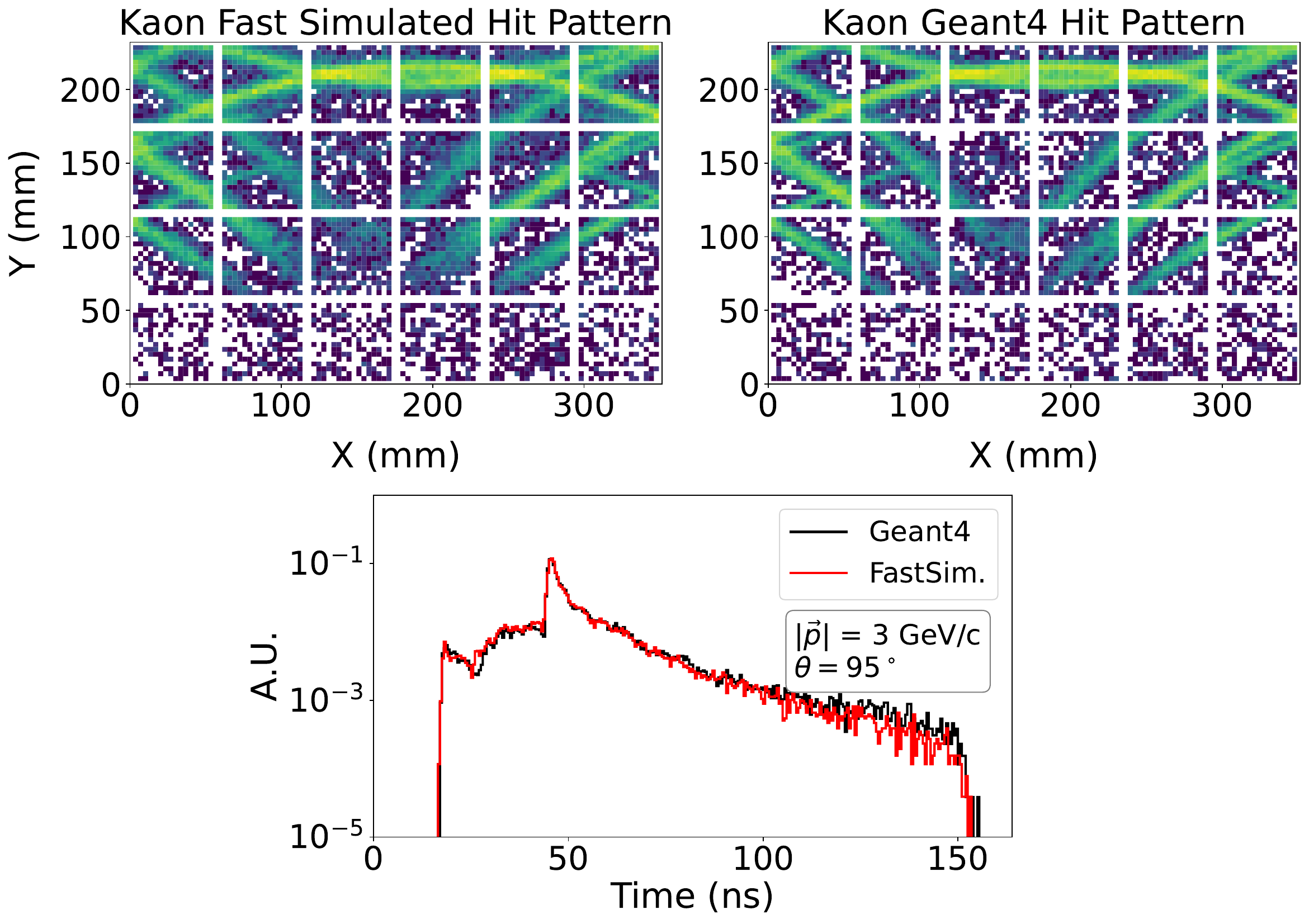} %
    \includegraphics[width=0.49\textwidth]{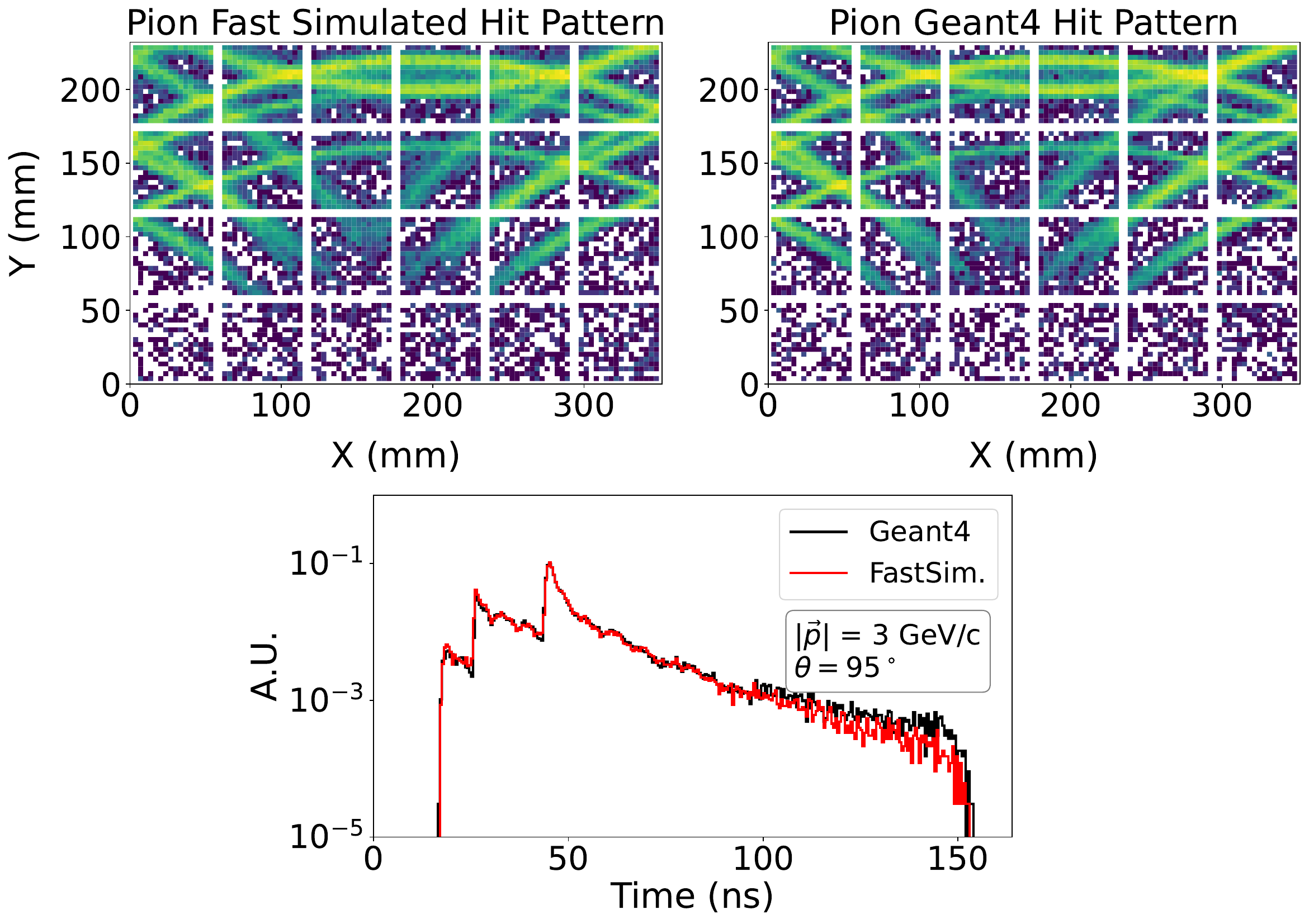} \\
    \includegraphics[width=0.49\textwidth]{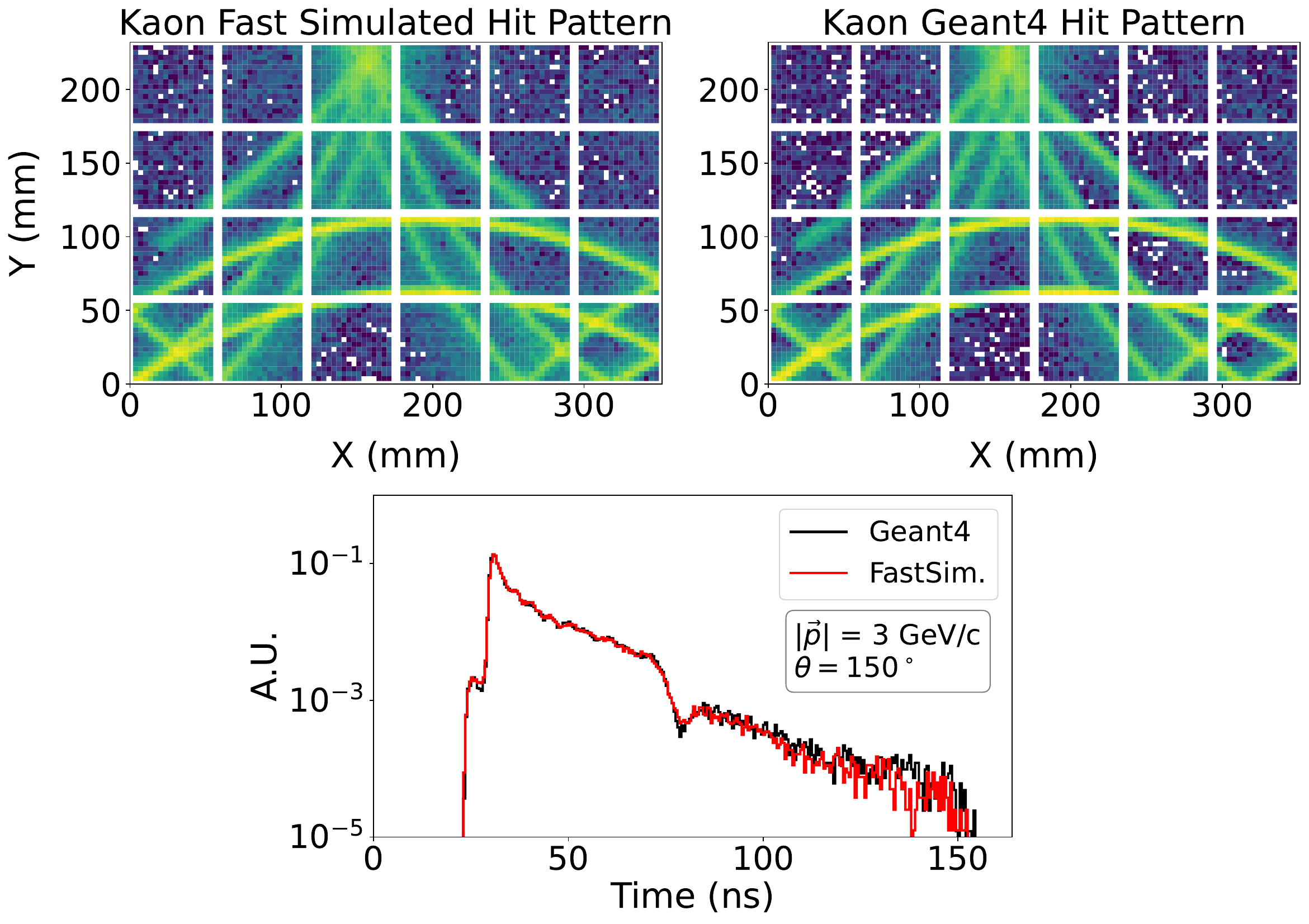} %
    \includegraphics[width=0.49\textwidth]{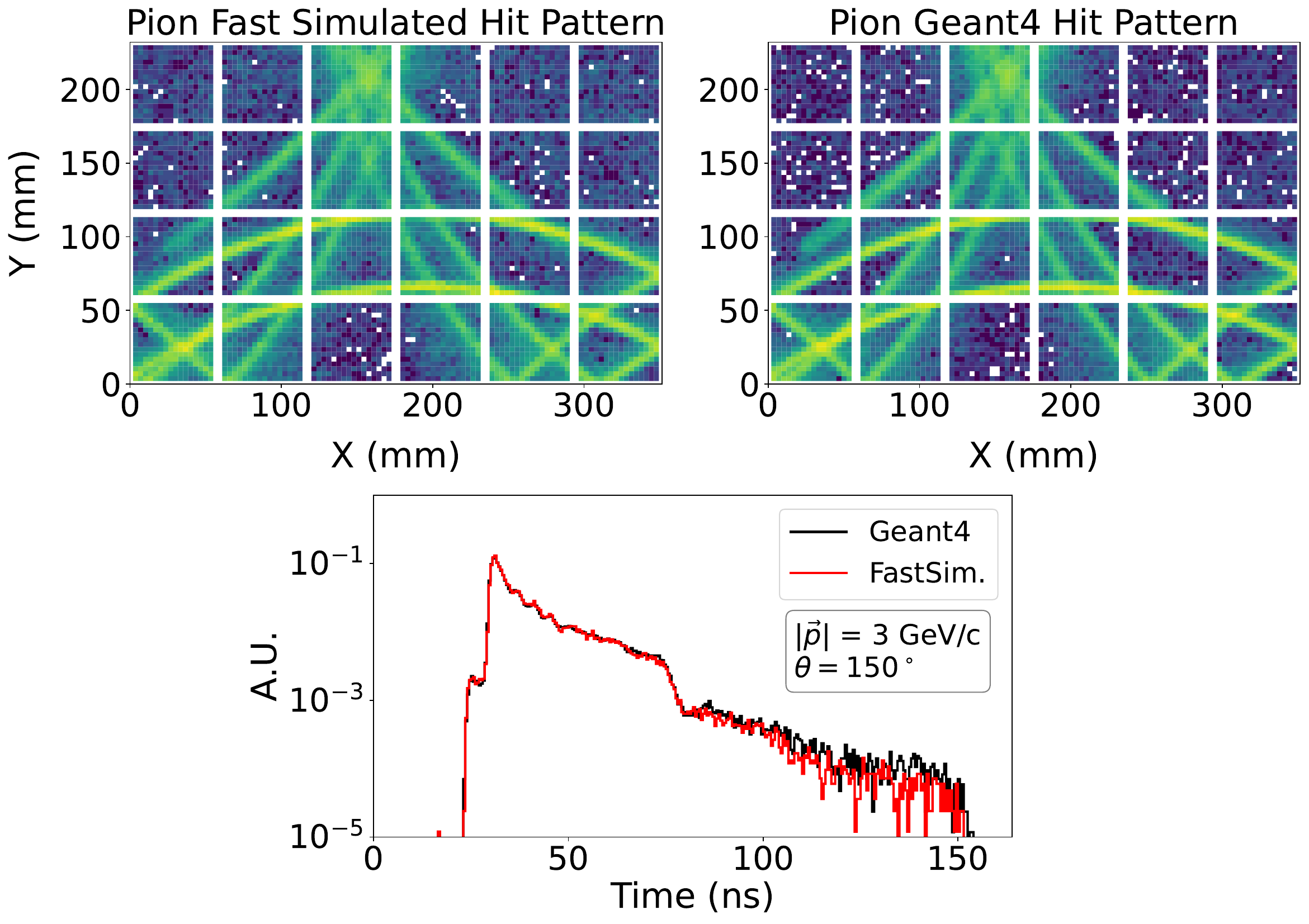} %
    \caption{
    \textbf{Fast Simulation at 3 GeV/c:} Fast Simulation of kaons (left column of plots), and pions (right column of plots) at 3 GeV/c and various polar angles, using Nucleus sampling, fixed temperature and a mixture of four experts (class conditional routing).}
    \label{fig:Generations_3GeV}
\end{figure}

% \begin{figure}[h]
%     \centering
%       \begin{subfigure}[b]{0.49\textwidth}
%         \centering
%         \includegraphics[width=\textwidth]{Figures/Generations/4Experts/3GeV/Ratios_Kaon.pdf}
%         \caption{Kaons}
%     \end{subfigure}
%     \begin{subfigure}[b]{0.49\textwidth}
%         \centering
%         \includegraphics[width=\textwidth]{Figures/Generations/4Experts/3GeV/Ratios_Pion.pdf}
%         \caption{Pions}
%     \end{subfigure} 
%     \caption{\textbf{Ratio Plots at 3 GeV/c:} Ratio plots for kaons (top) and pions (bottom), using Nucleus sampling, fixed temperature and a mixture of four experts (class conditional routing).}
%     \label{fig:ratios_3GeV}
% \end{figure}

\begin{figure}[h]
    \centering
      \begin{subfigure}[b]{0.49\textwidth}
        \centering
        \includegraphics[width=\textwidth]{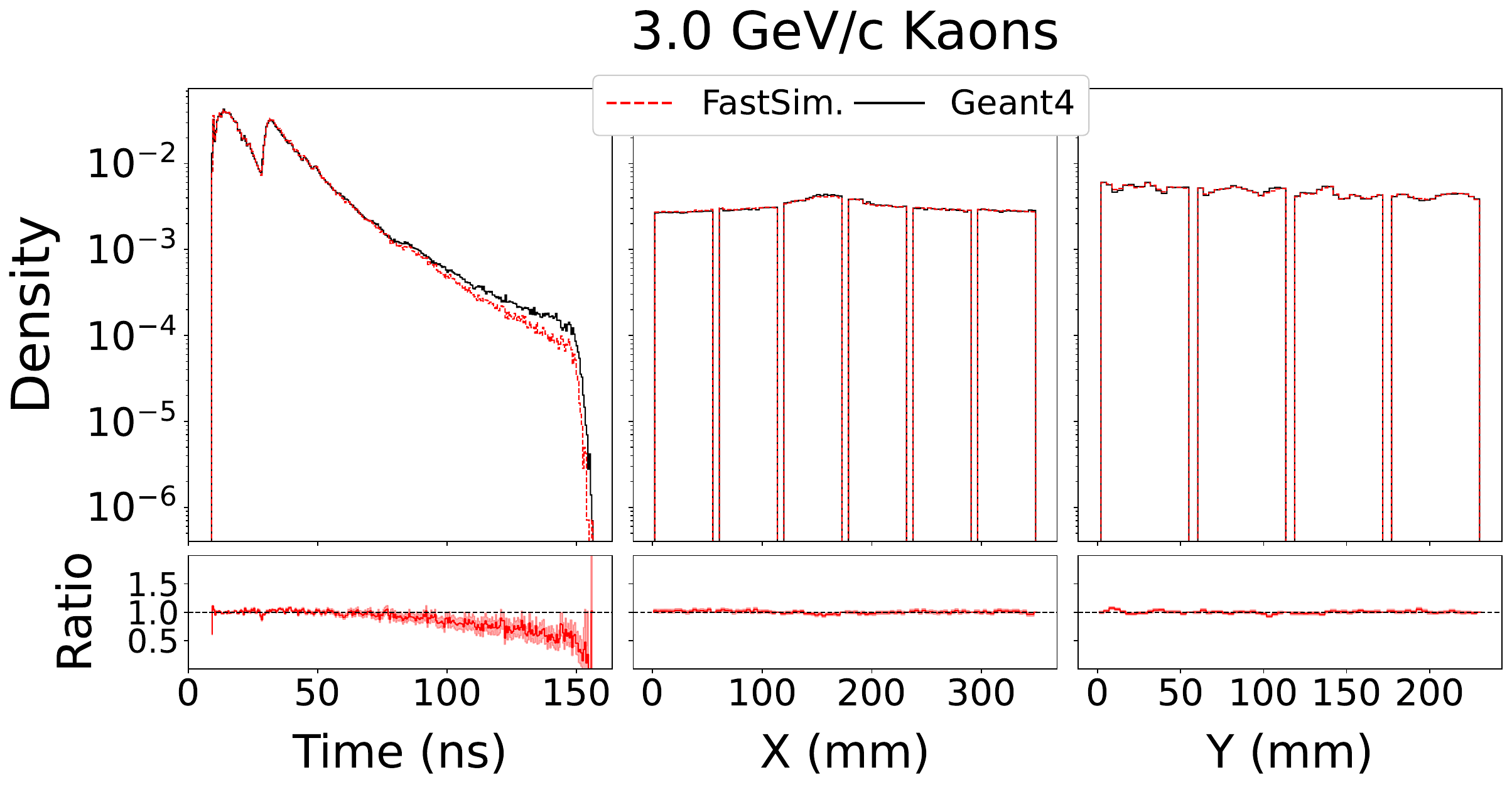}
        \caption{Kaons - Independent Model}
    \end{subfigure}
    \begin{subfigure}[b]{0.49\textwidth}
        \centering
        \includegraphics[width=\textwidth]{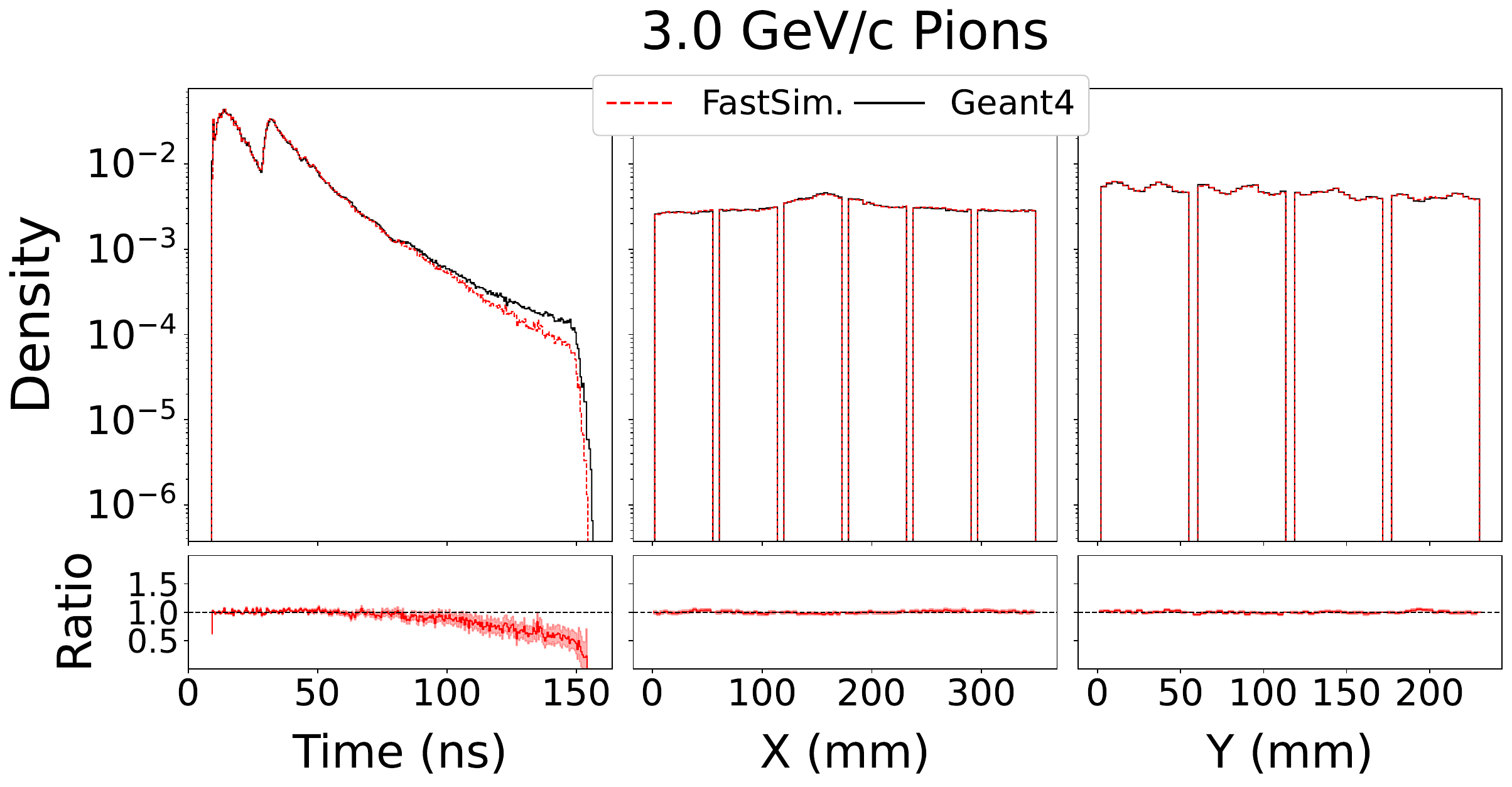}
        \caption{Pions - Independent Model}
    \end{subfigure} \\
    \begin{subfigure}[b]{0.49\textwidth}
        \centering
        \includegraphics[width=\textwidth]{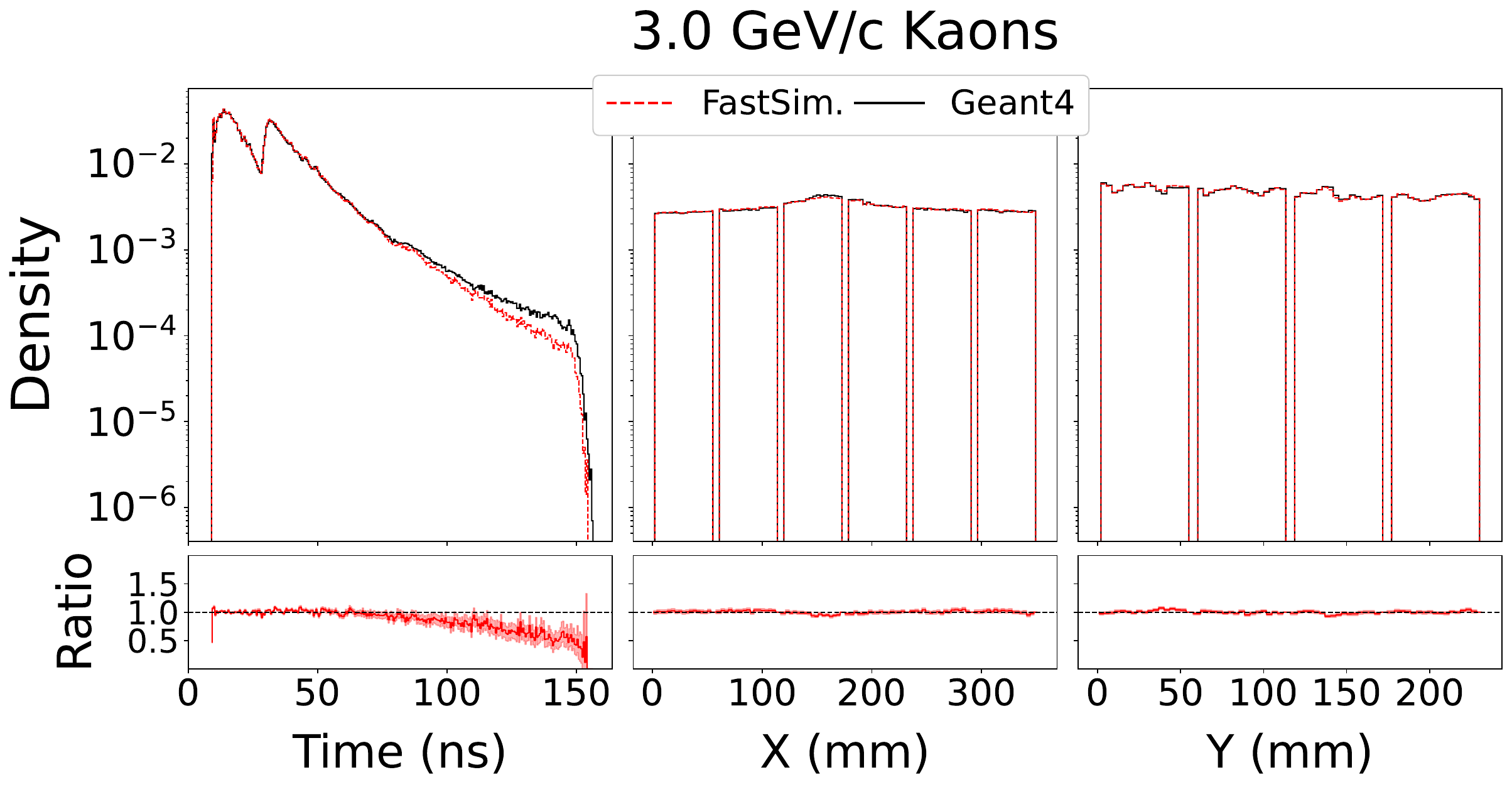}
        \caption{Kaons - Two Experts}
    \end{subfigure} 
    \begin{subfigure}[b]{0.49\textwidth}
        \centering
        \includegraphics[width=\textwidth]{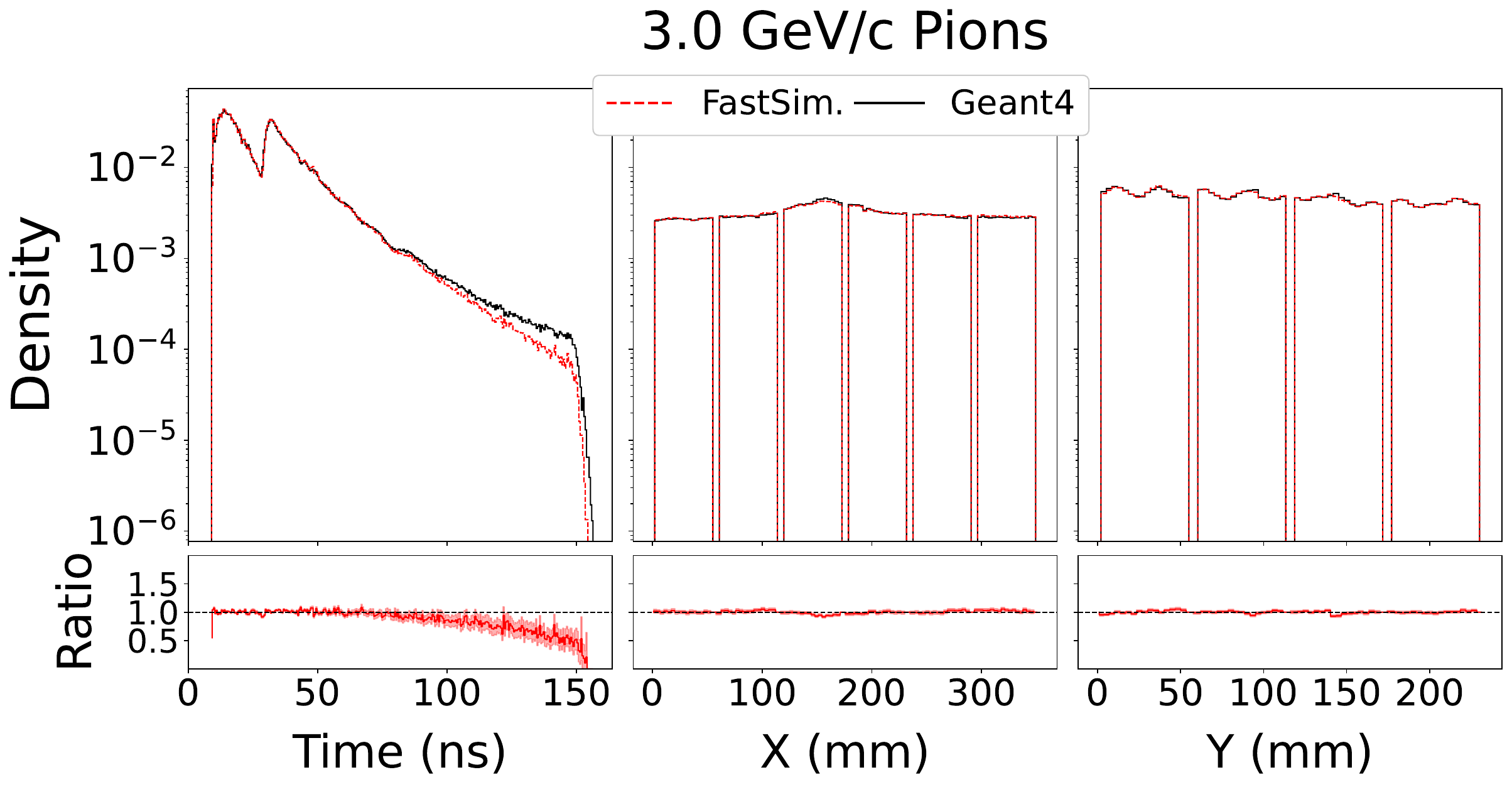}
        \caption{Pions - Two Experts}
    \end{subfigure} \\
    \begin{subfigure}[b]{0.49\textwidth}
        \centering
        \includegraphics[width=\textwidth]{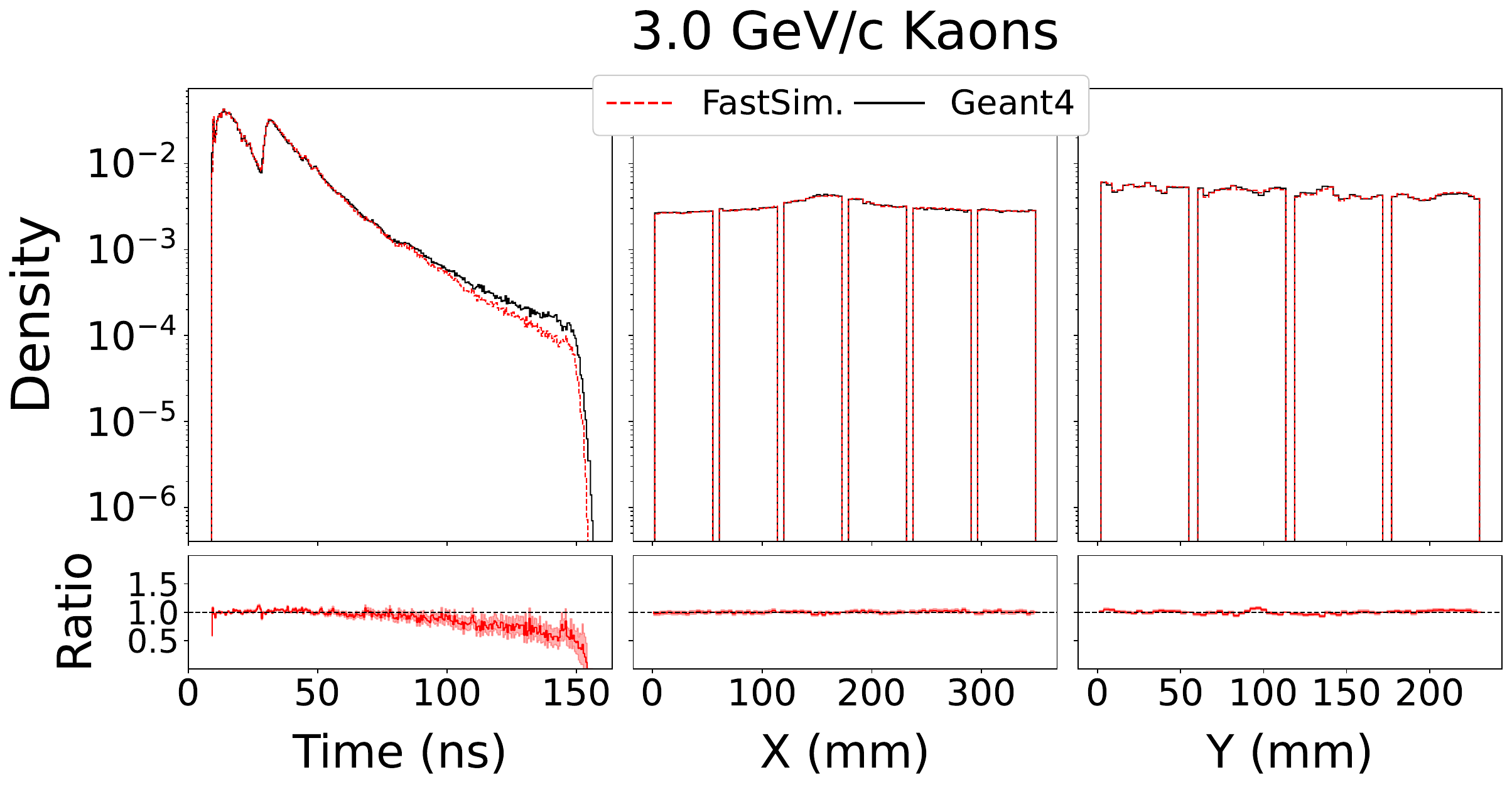}
        \caption{Kaons - Four Experts}
    \end{subfigure} 
    \begin{subfigure}[b]{0.49\textwidth}
        \centering
        \includegraphics[width=\textwidth]{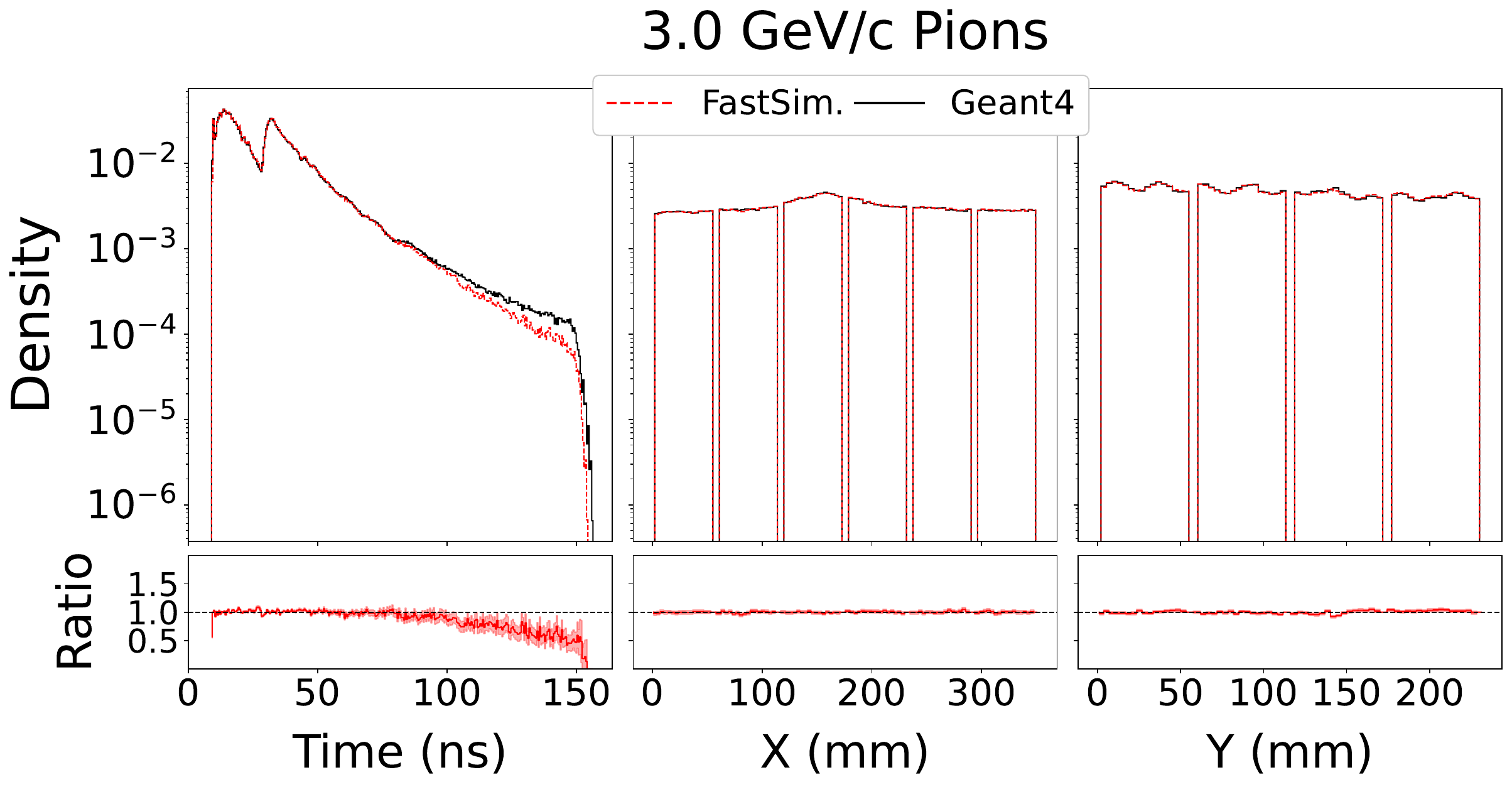}
        \caption{Pions - Four Experts}
    \end{subfigure} 
    \caption{\textbf{Ratio Plots at 3 GeV/c:} Ratio plots for kaons (left column) and pions (right column), using Nucleus sampling and fixed temperature for independent models (top row), a combined model with two experts (middle row) and a combined model with four experts (bottom row).}
    \label{fig:ratios_3GeV}
\end{figure}

% \begin{figure}[h]
%     \centering
%       \begin{subfigure}[b]{0.49\textwidth}
%         \centering
%         \includegraphics[width=\textwidth]{Figures/Generations/4Experts/3GeV/Photon_Yield_Kaon.pdf}
%         \caption{Kaons}
%     \end{subfigure}
%     \begin{subfigure}[b]{0.49\textwidth}
%         \centering
%         \includegraphics[width=\textwidth]{Figures/Generations/4Experts/3GeV/Photon_Yield_Pion.pdf}
%         \caption{Pions}
%     \end{subfigure} 
%     \caption{\textbf{Photon Yield Comparison at 3 GeV/c:} Comparison of generated photon yield for kaons (top) and pions (bottom), using Nucleus sampling, fixed temperature and a mixture of four experts (class conditional routing).}
%     \label{fig:yields_3GeV}
% \end{figure}

\begin{figure}[!]
    \centering
      \begin{subfigure}[b]{0.49\textwidth}
        \centering
        \includegraphics[width=\textwidth]{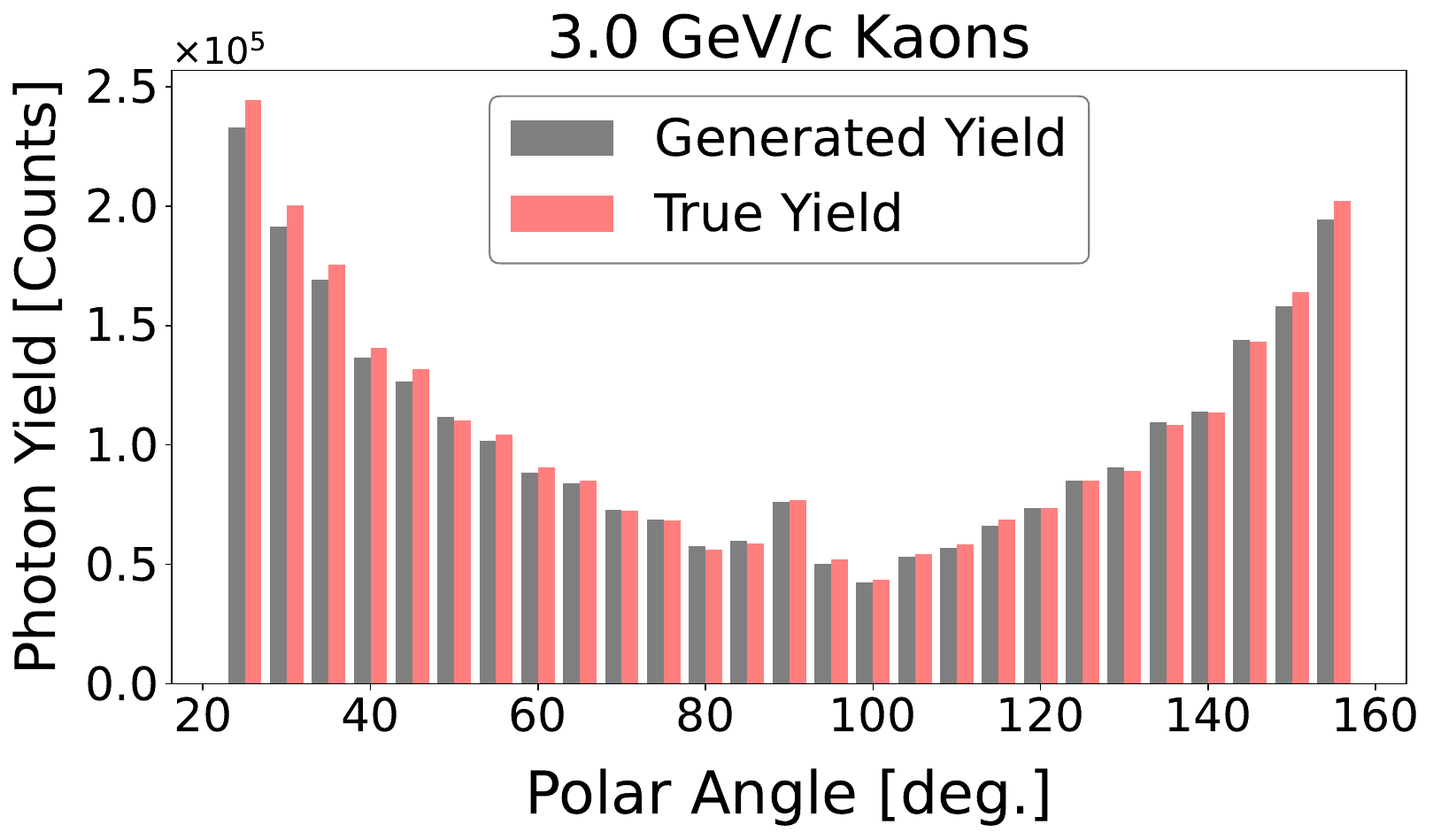}
        \caption{Kaons  - Independent Model}
    \end{subfigure}
    \begin{subfigure}[b]{0.49\textwidth}
        \centering
        \includegraphics[width=\textwidth]{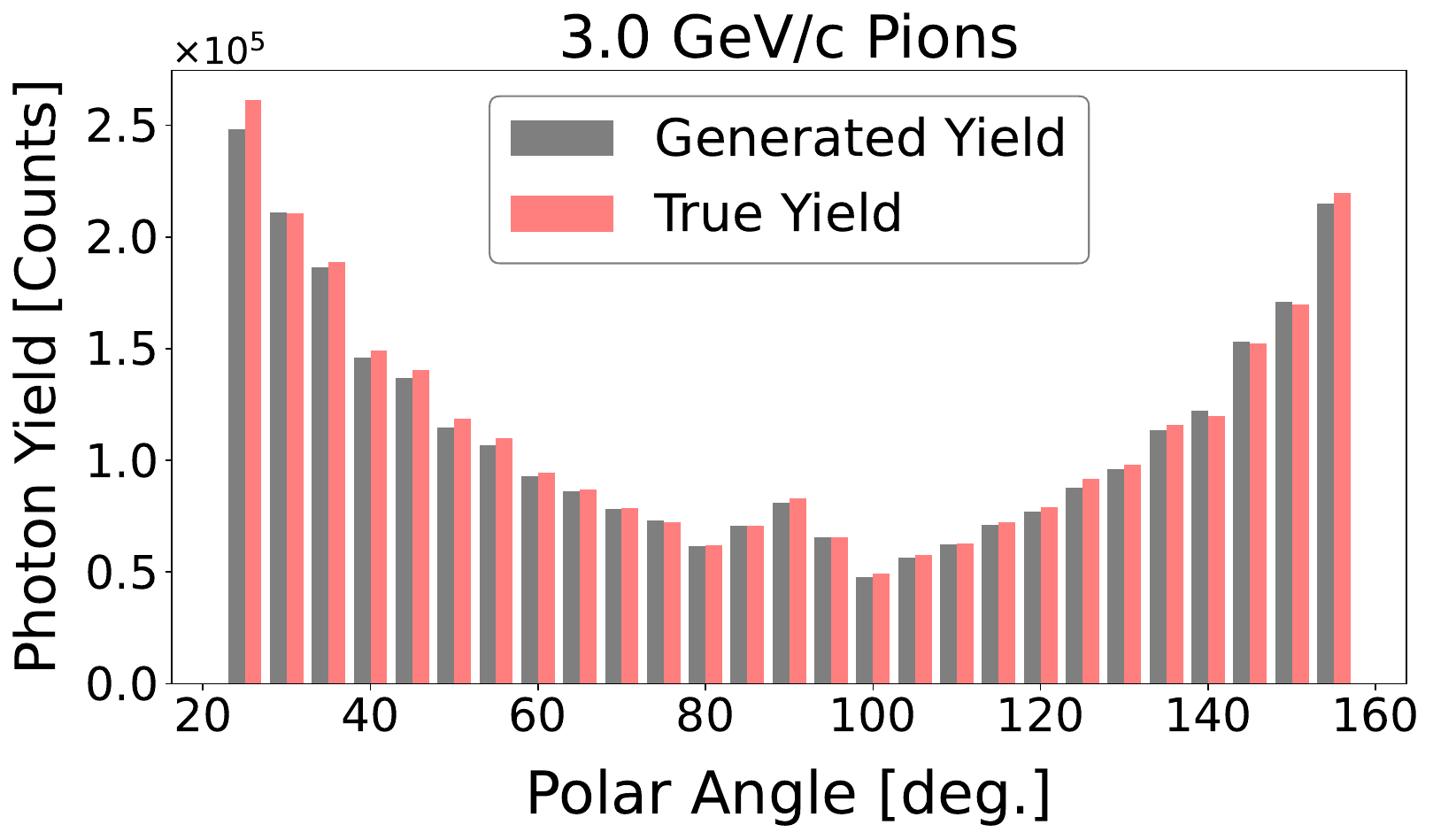}
        \caption{Pions  - Independent Model}
    \end{subfigure} \\
      \begin{subfigure}[b]{0.49\textwidth}
        \centering
        \includegraphics[width=\textwidth]{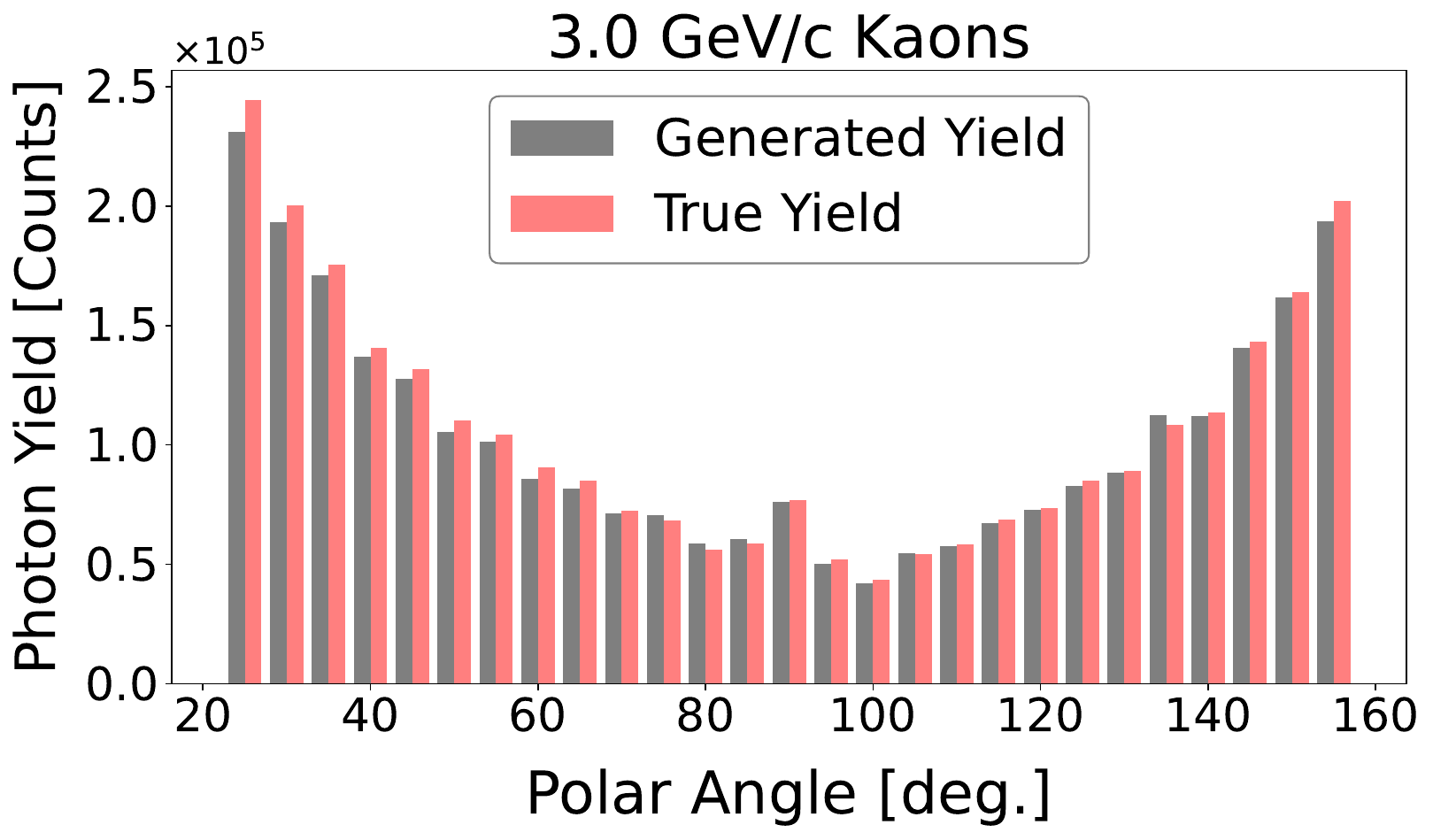}
        \caption{Kaons - Two Experts}
    \end{subfigure}
    \begin{subfigure}[b]{0.49\textwidth}
        \centering
        \includegraphics[width=\textwidth]{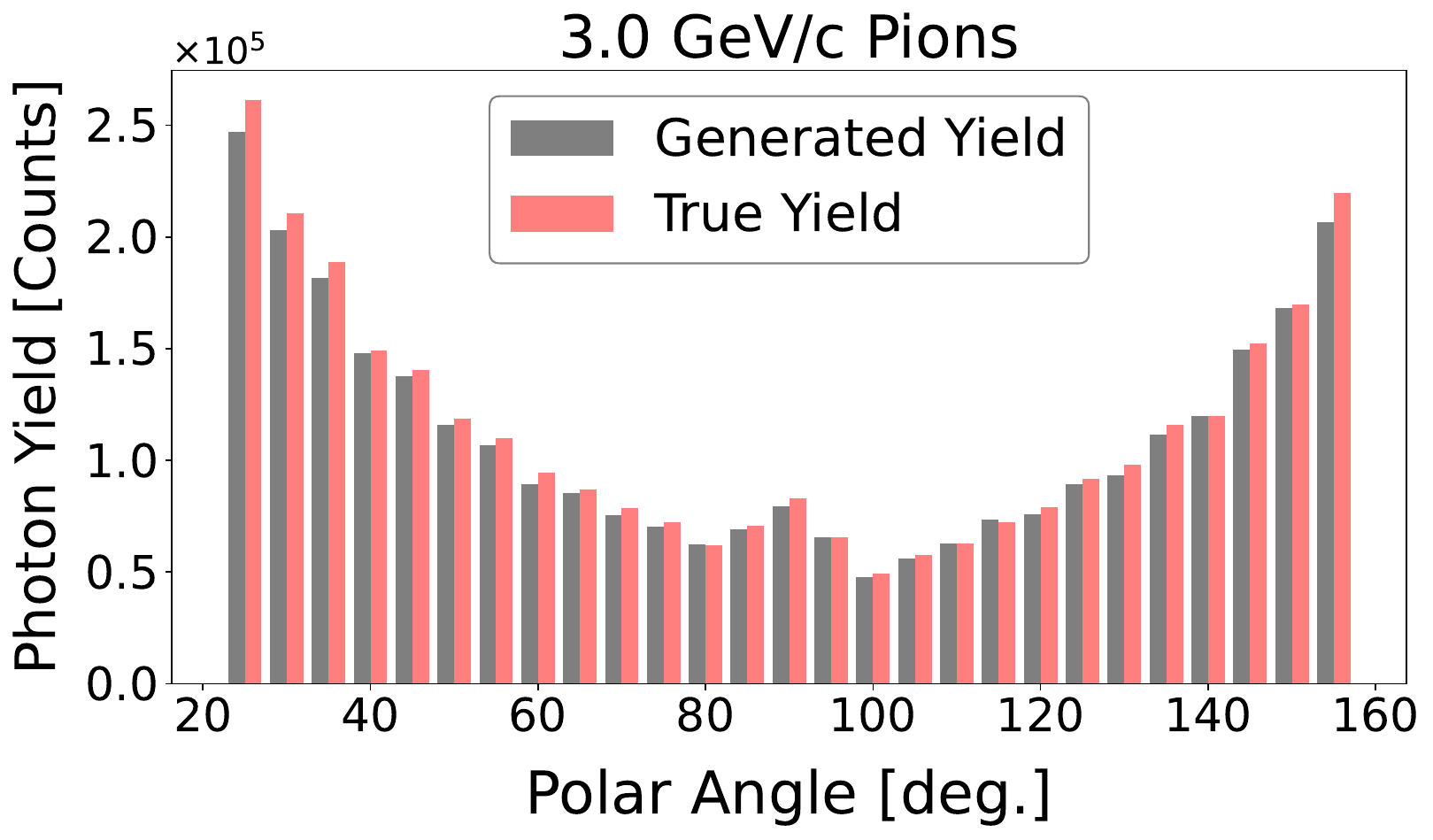}
        \caption{Pions - Two Experts}
    \end{subfigure} 
      \begin{subfigure}[b]{0.49\textwidth}
        \centering
        \includegraphics[width=\textwidth]{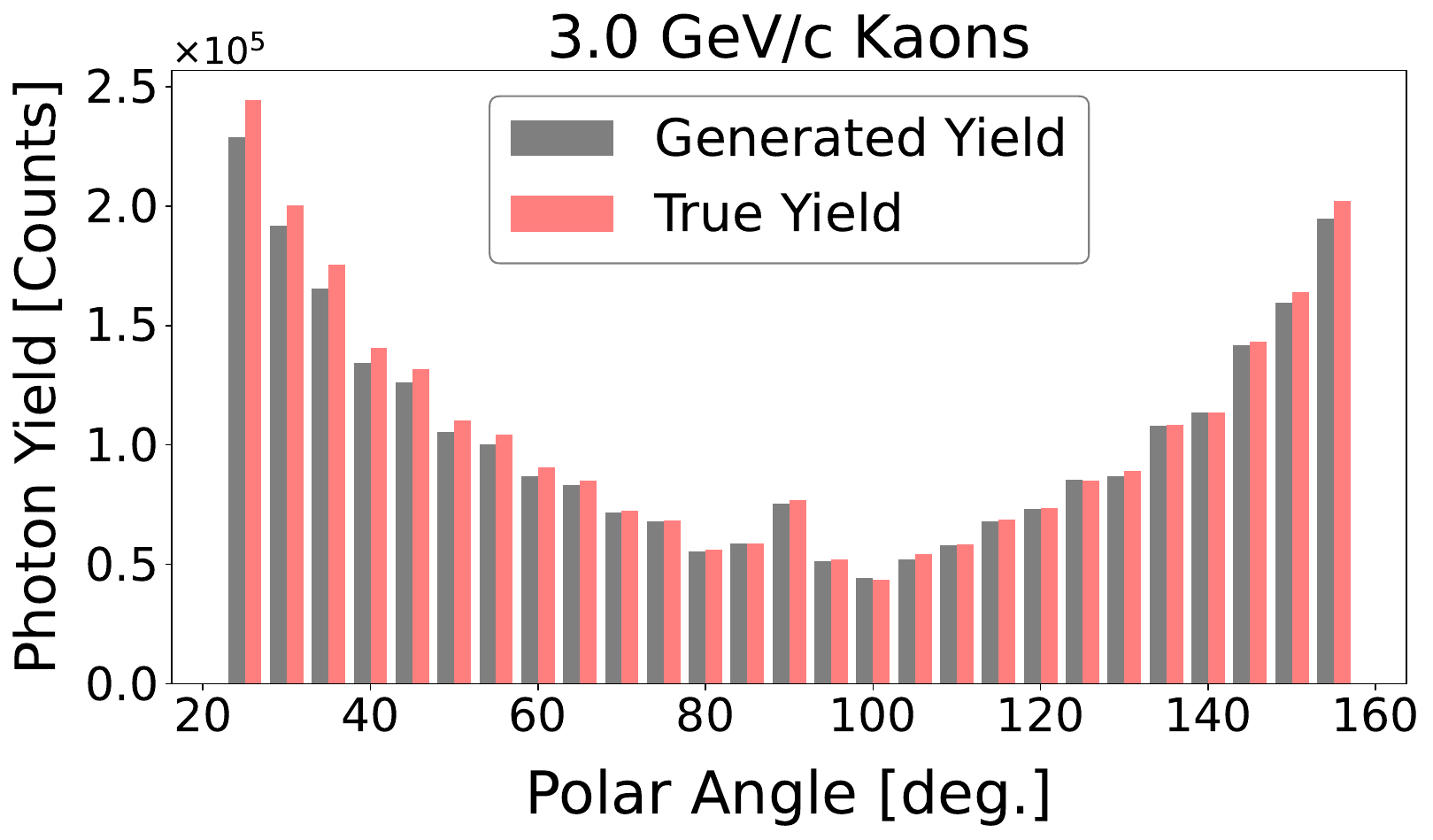}
        \caption{Kaons - Four Experts}
    \end{subfigure}
    \begin{subfigure}[b]{0.49\textwidth}
        \centering
        \includegraphics[width=\textwidth]{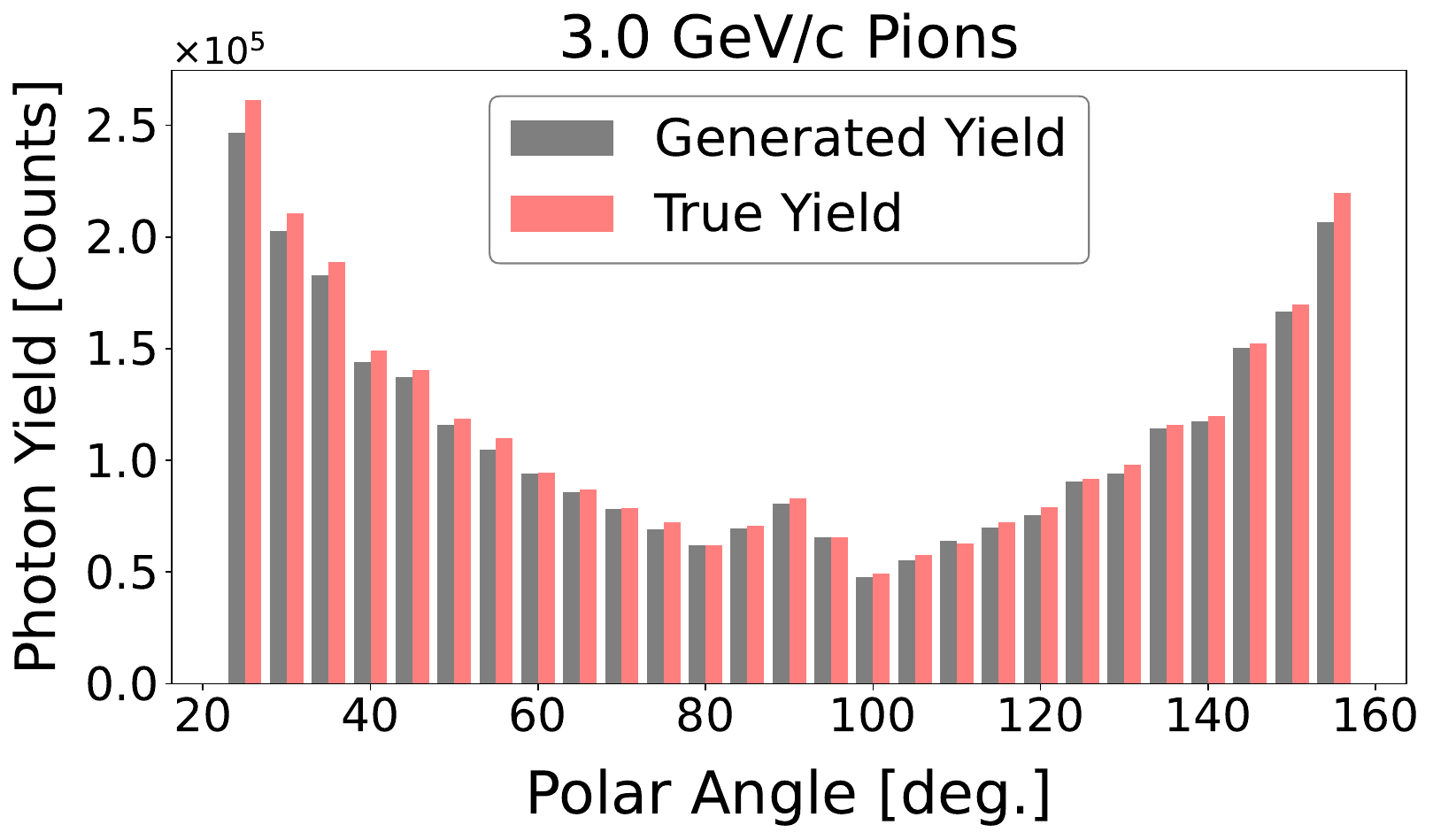}
        \caption{Pions - Four Experts}
    \end{subfigure} 
    \caption{\textbf{Photon Yield Comparison at 3 GeV/c:} Comparison of generated photon yield for kaons (left column) and pions (right column), using Nucleus sampling and fixed temperature for independent models (top row), a combined model with two experts (middle row) and a combined model with four experts (bottom row).}
    \label{fig:yields_3GeV}
\end{figure}

%%%%%%%%%%%%%% 9 GeV %%%%%%%%%%%%%%%%%%%%%
\clearpage
\section{Evaluation at $\SI[per-mode=symbol]{9}{\giga\eVperc}$} \label{app:9GeV}

\begin{figure}[h]
    \centering
    \includegraphics[width=0.49\textwidth]{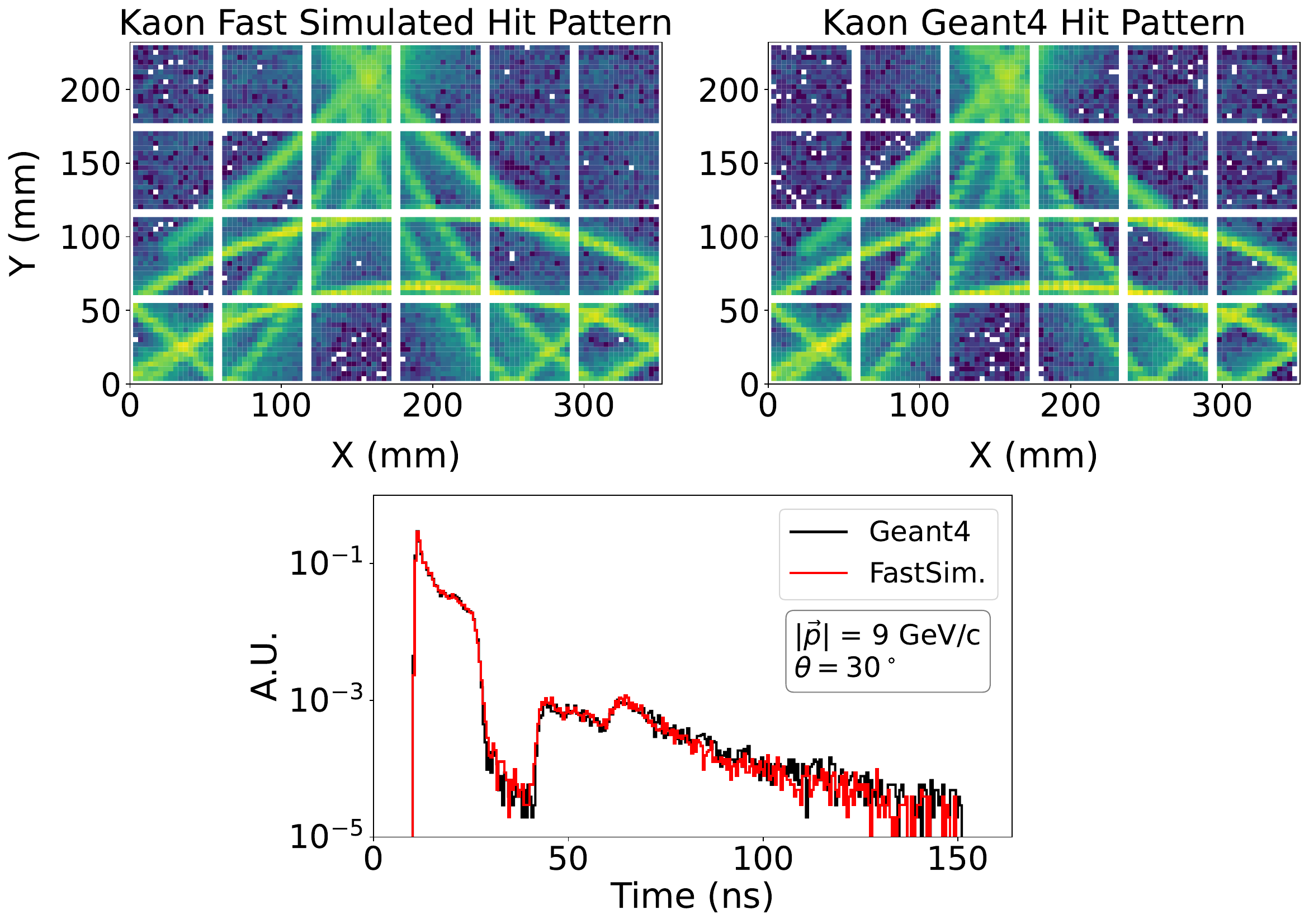}% 
   \includegraphics[width=0.49\textwidth]{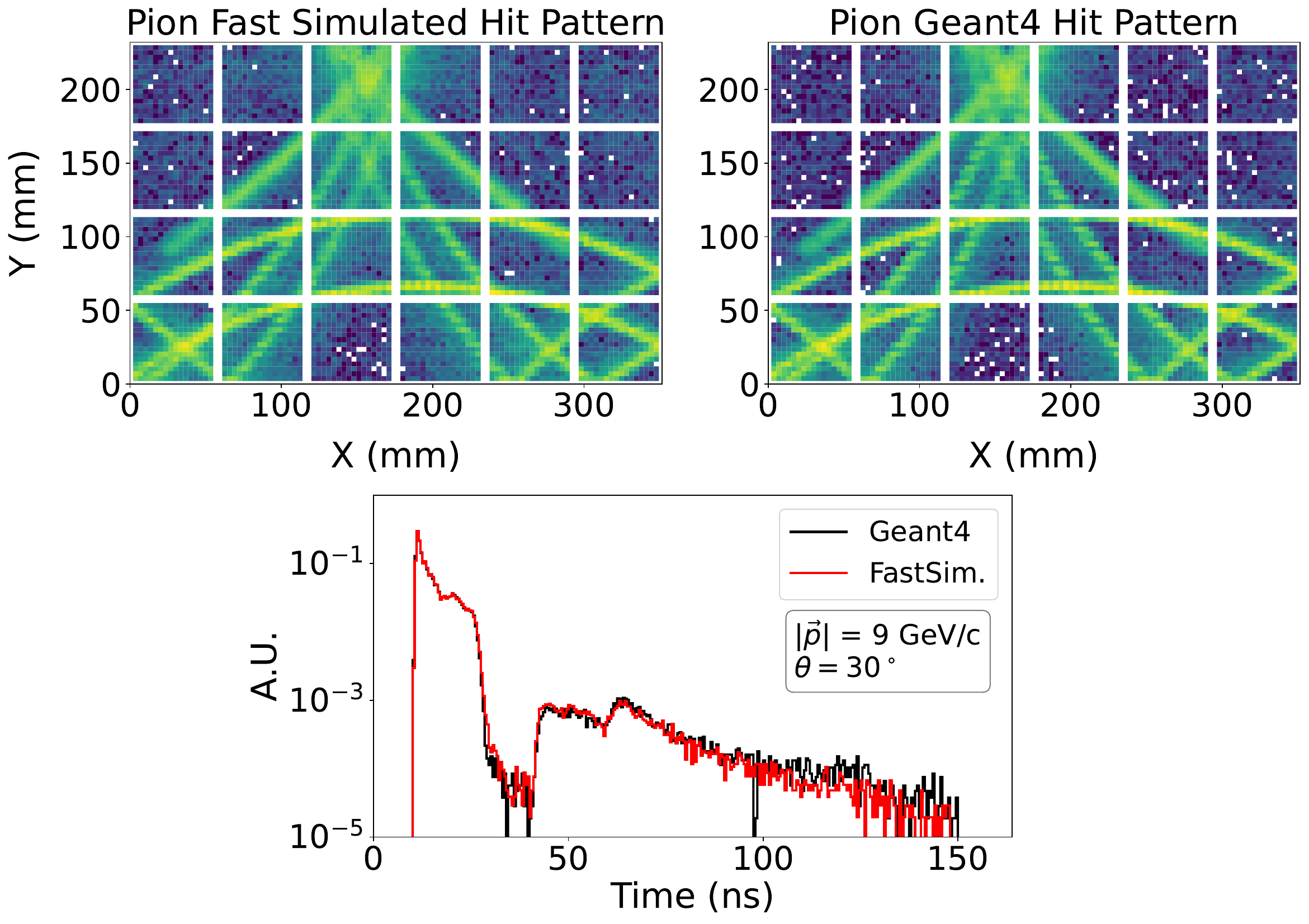} \\
    \includegraphics[width=0.49\textwidth]{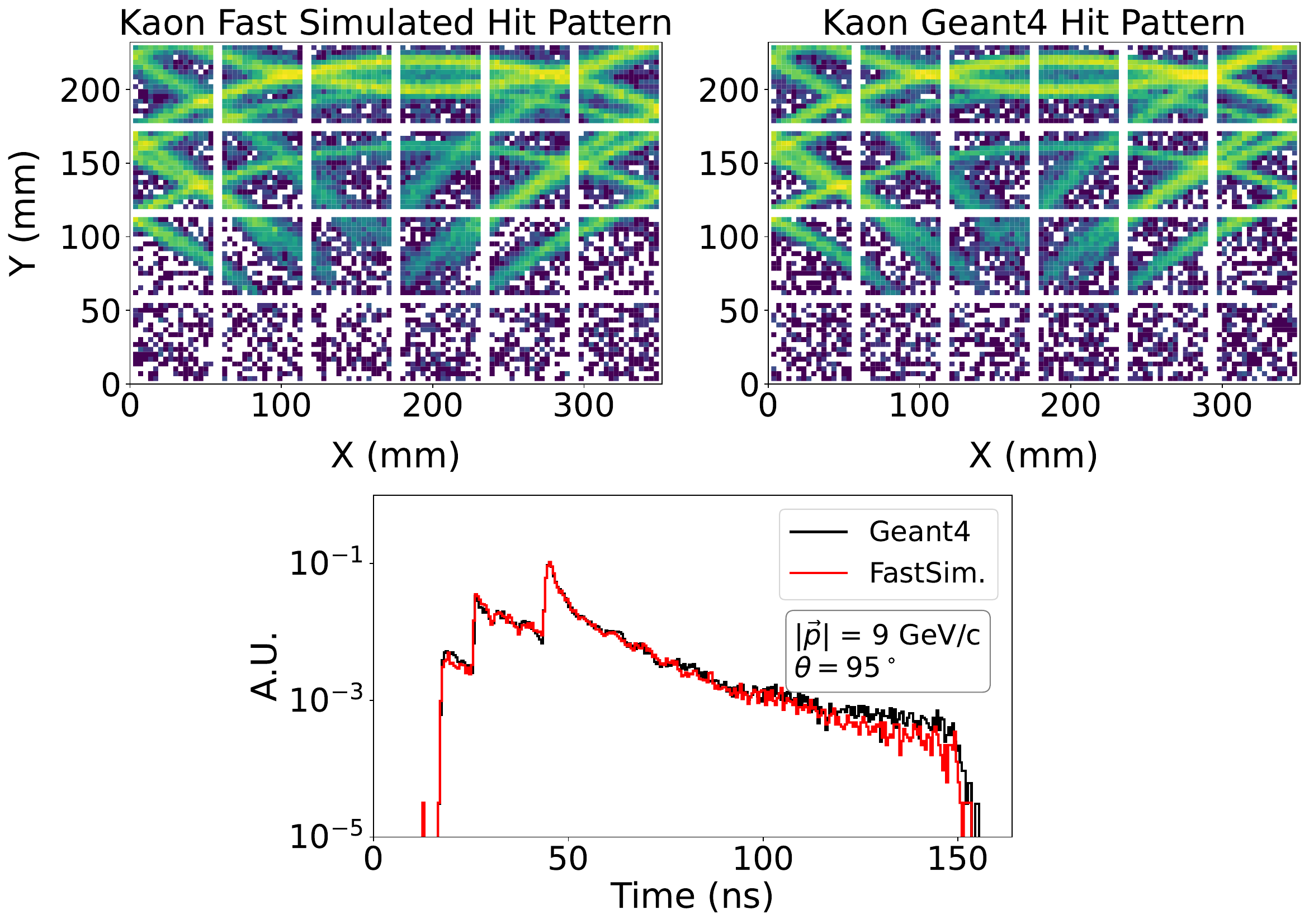} %
    \includegraphics[width=0.49\textwidth]{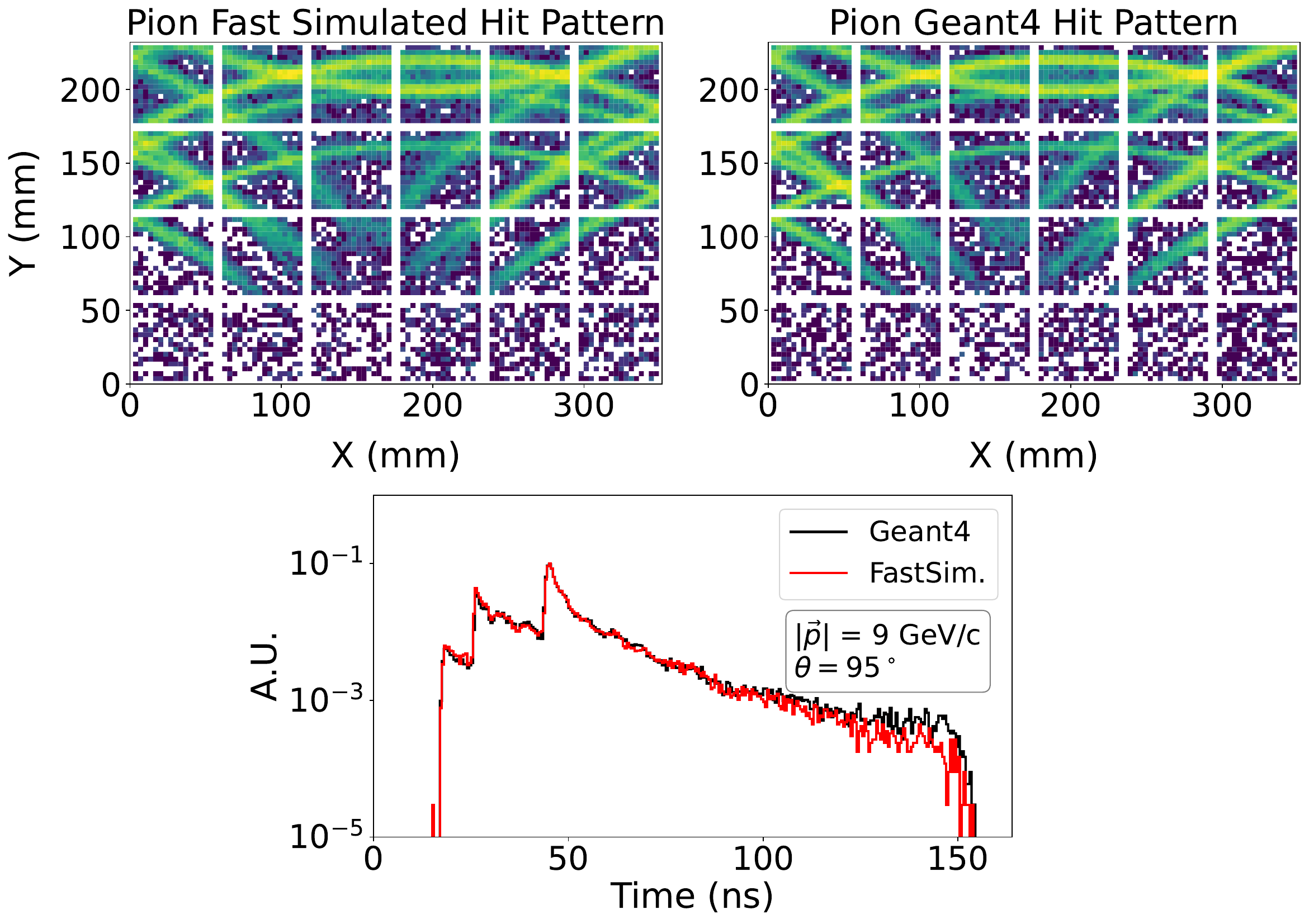} \\
    \includegraphics[width=0.49\textwidth]{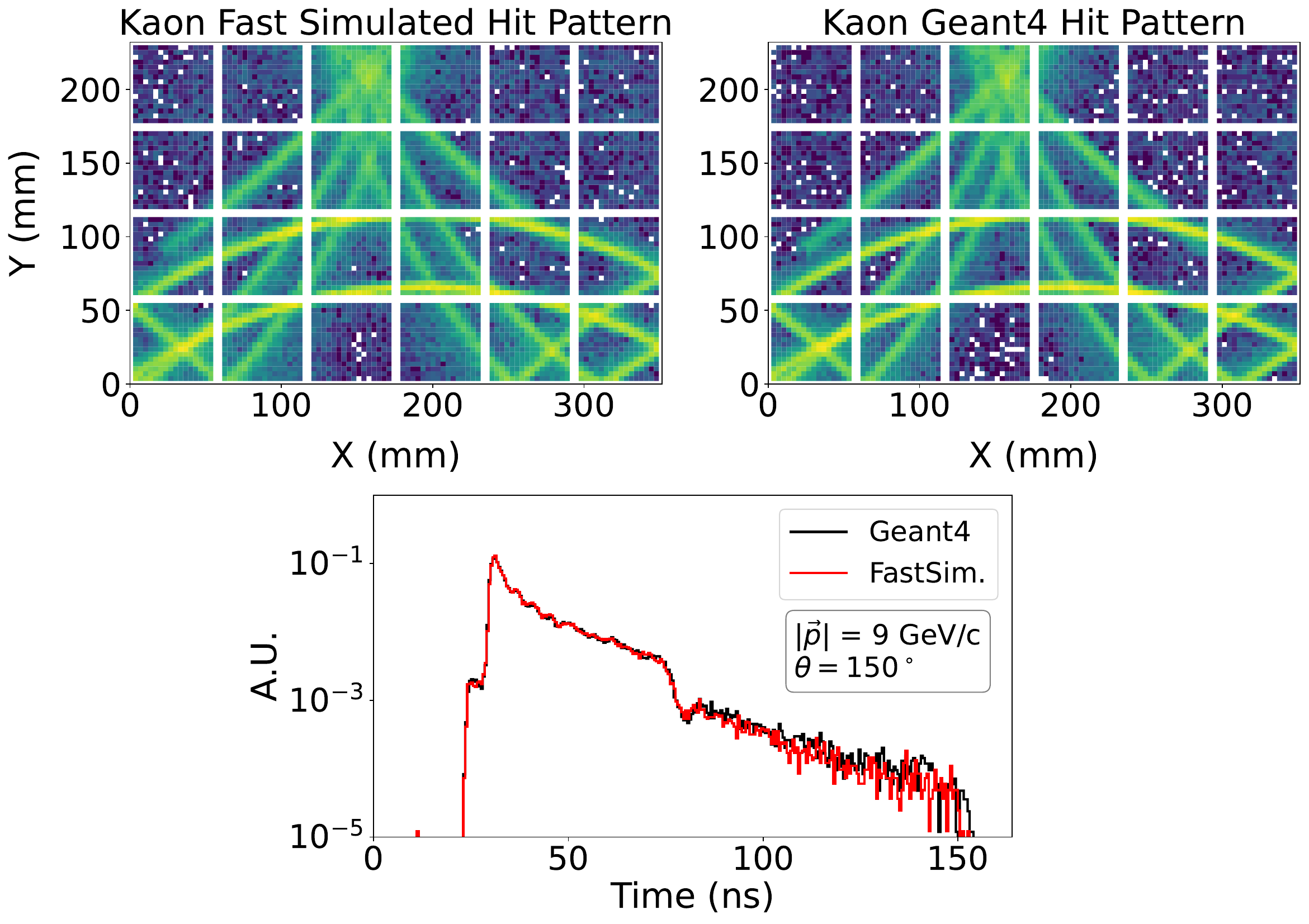} %
    \includegraphics[width=0.49\textwidth]{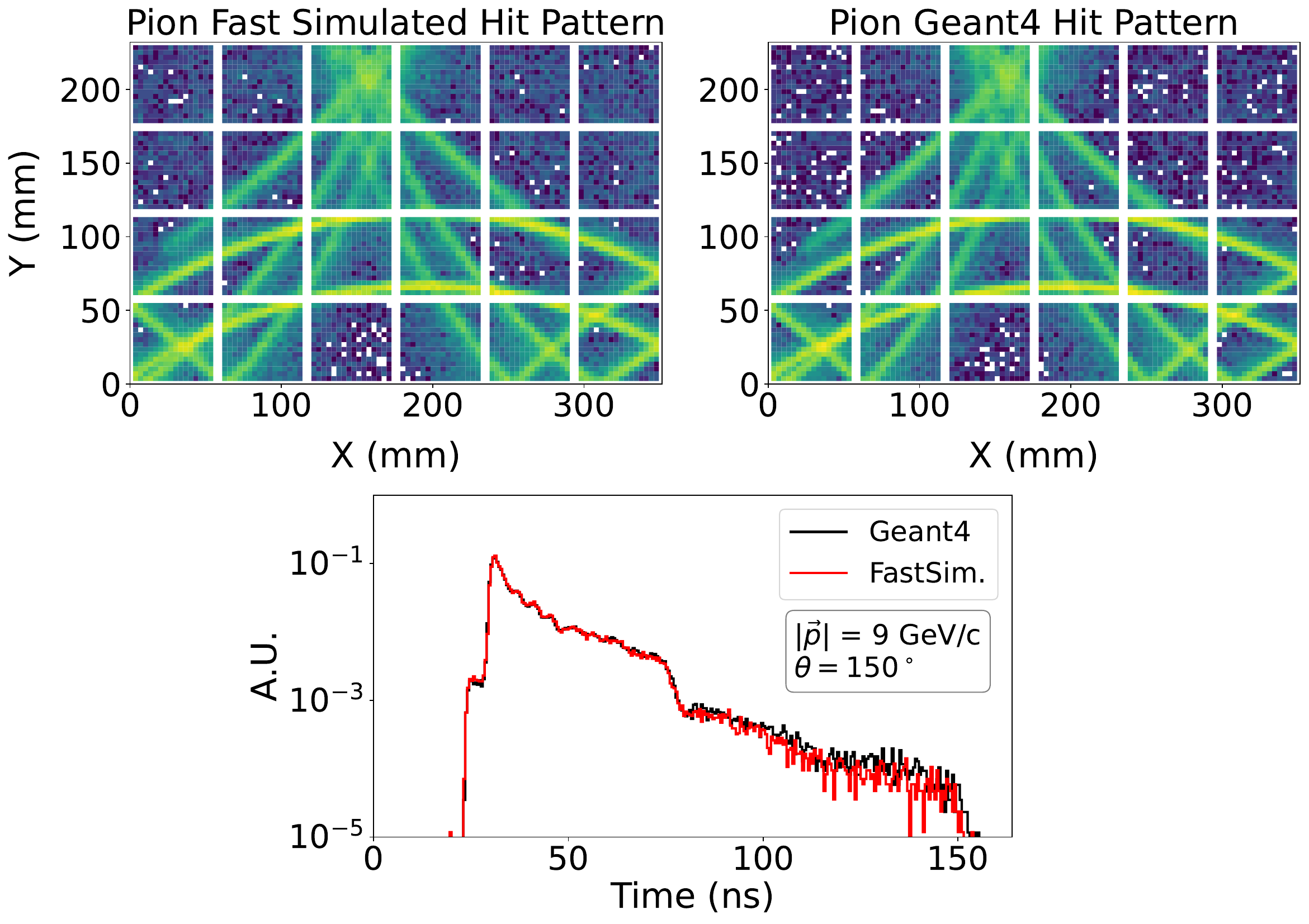} %
    \caption{
    \textbf{Fast Simulation at 9 GeV/c:} Fast Simulation of kaons (left column of plots), and pions (right column of plots) at 9 GeV/c and various polar angles, using Nucleus sampling and fixed temperature.}
    \label{fig:Generations_9GeV}
\end{figure}

\begin{figure}[h]
    \centering
      \begin{subfigure}[b]{0.49\textwidth}
        \centering
        \includegraphics[width=\textwidth]{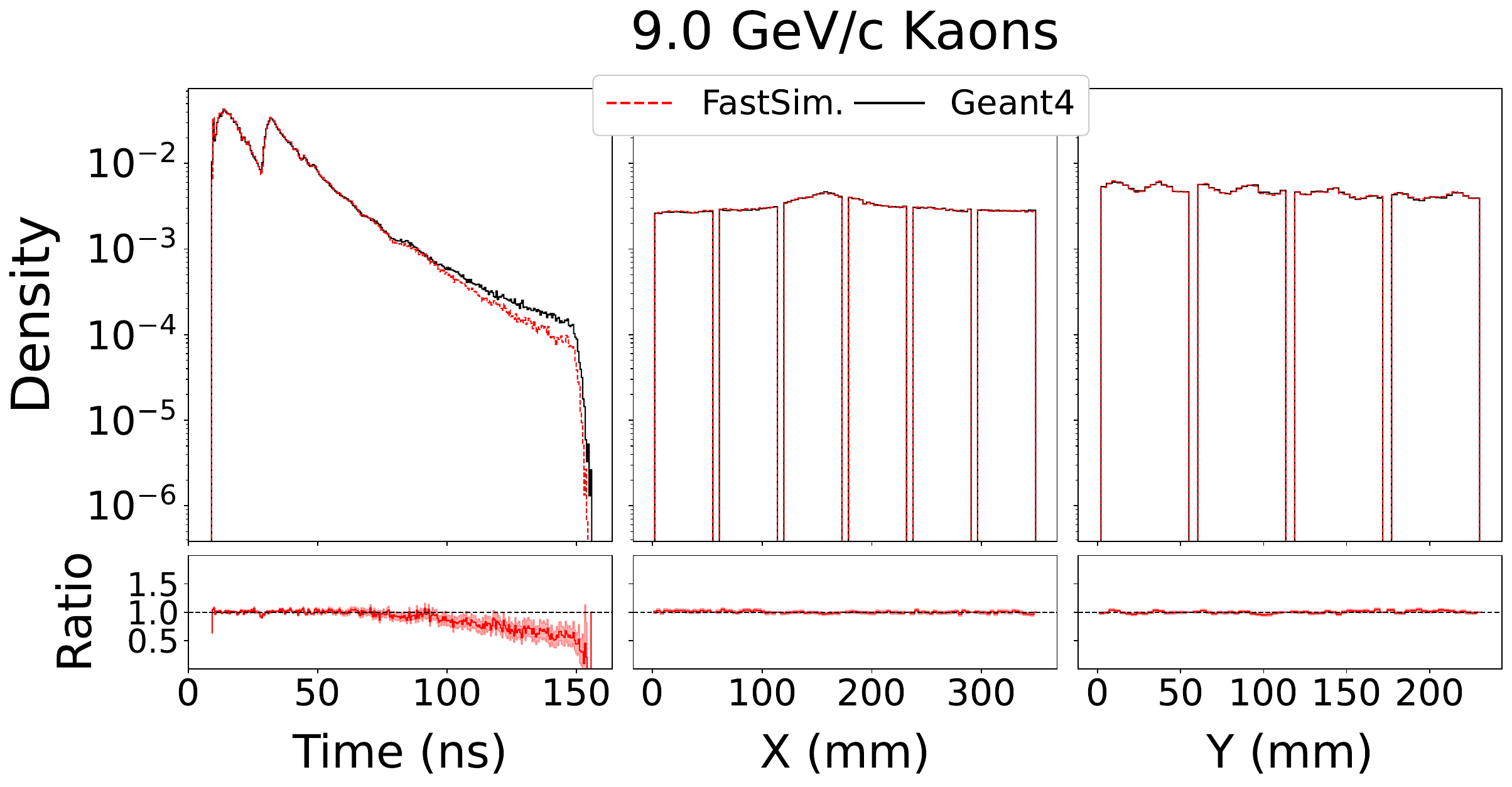}
        \caption{Kaons - Independent Model}
    \end{subfigure}
    \begin{subfigure}[b]{0.49\textwidth}
        \centering
        \includegraphics[width=\textwidth]{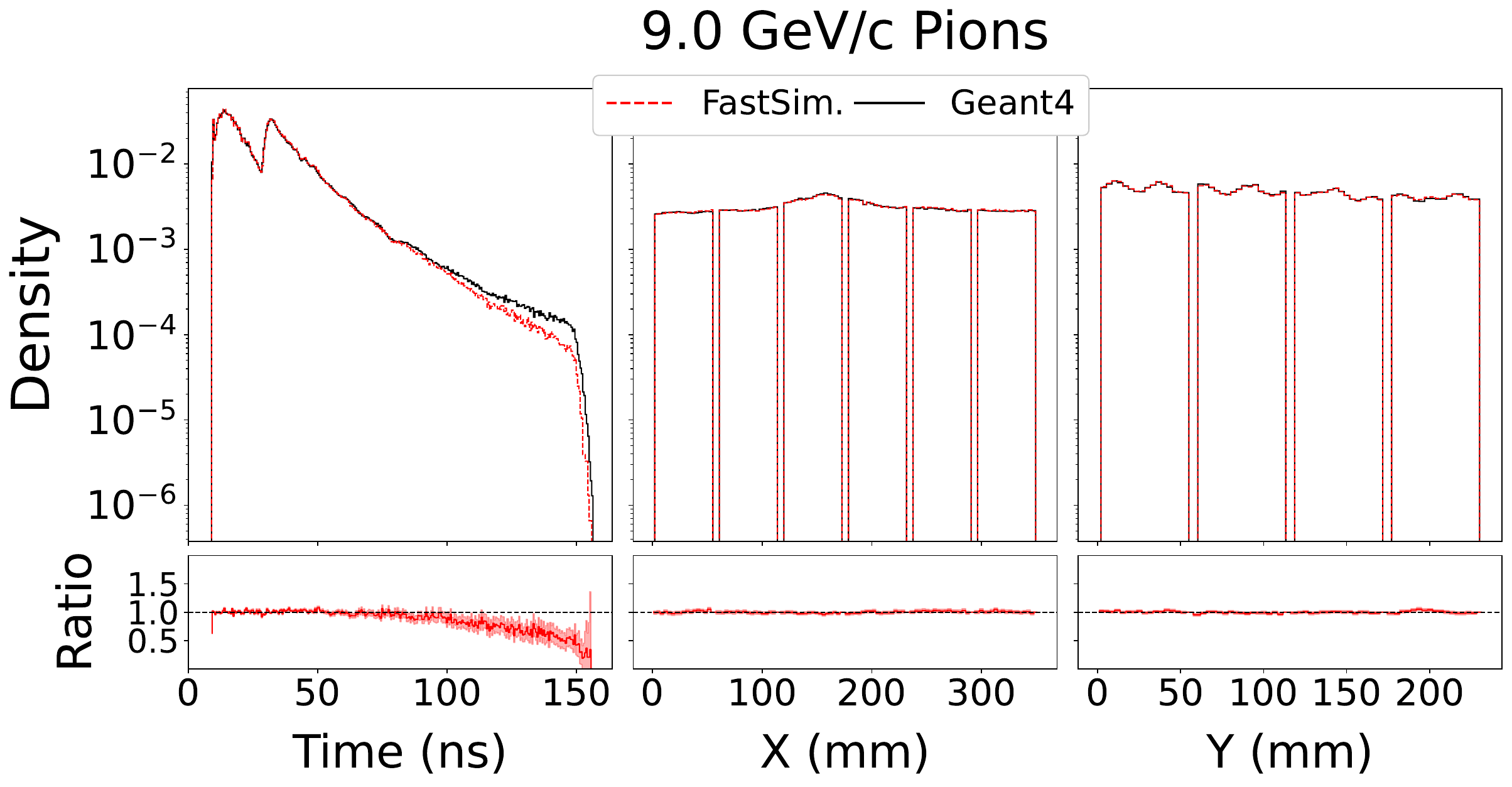}
        \caption{Pions - Independent Model}
    \end{subfigure} \\
    \begin{subfigure}[b]{0.49\textwidth}
        \centering
        \includegraphics[width=\textwidth]{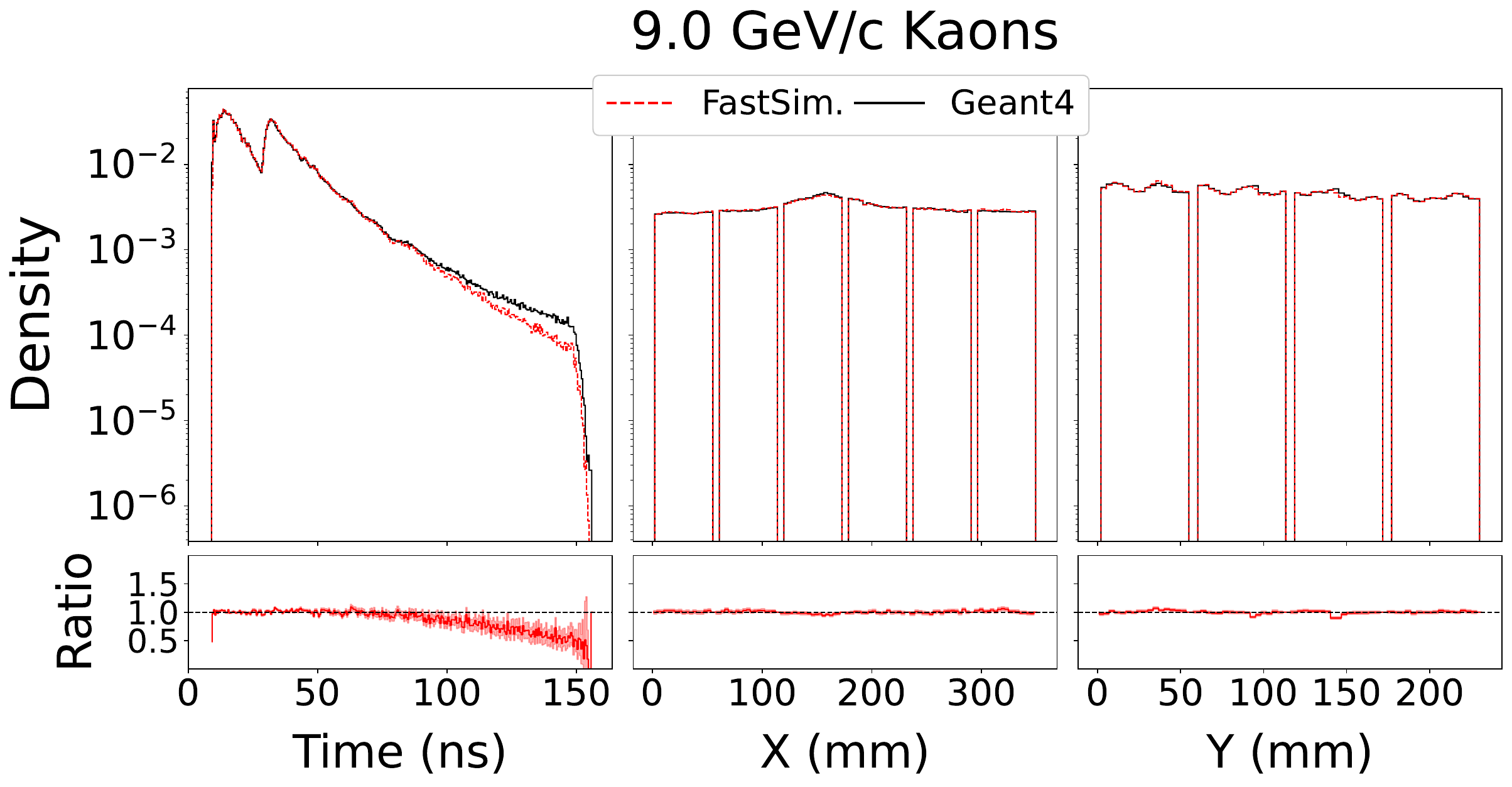}
        \caption{Kaons - Two Experts}
    \end{subfigure} 
    \begin{subfigure}[b]{0.49\textwidth}
        \centering
        \includegraphics[width=\textwidth]{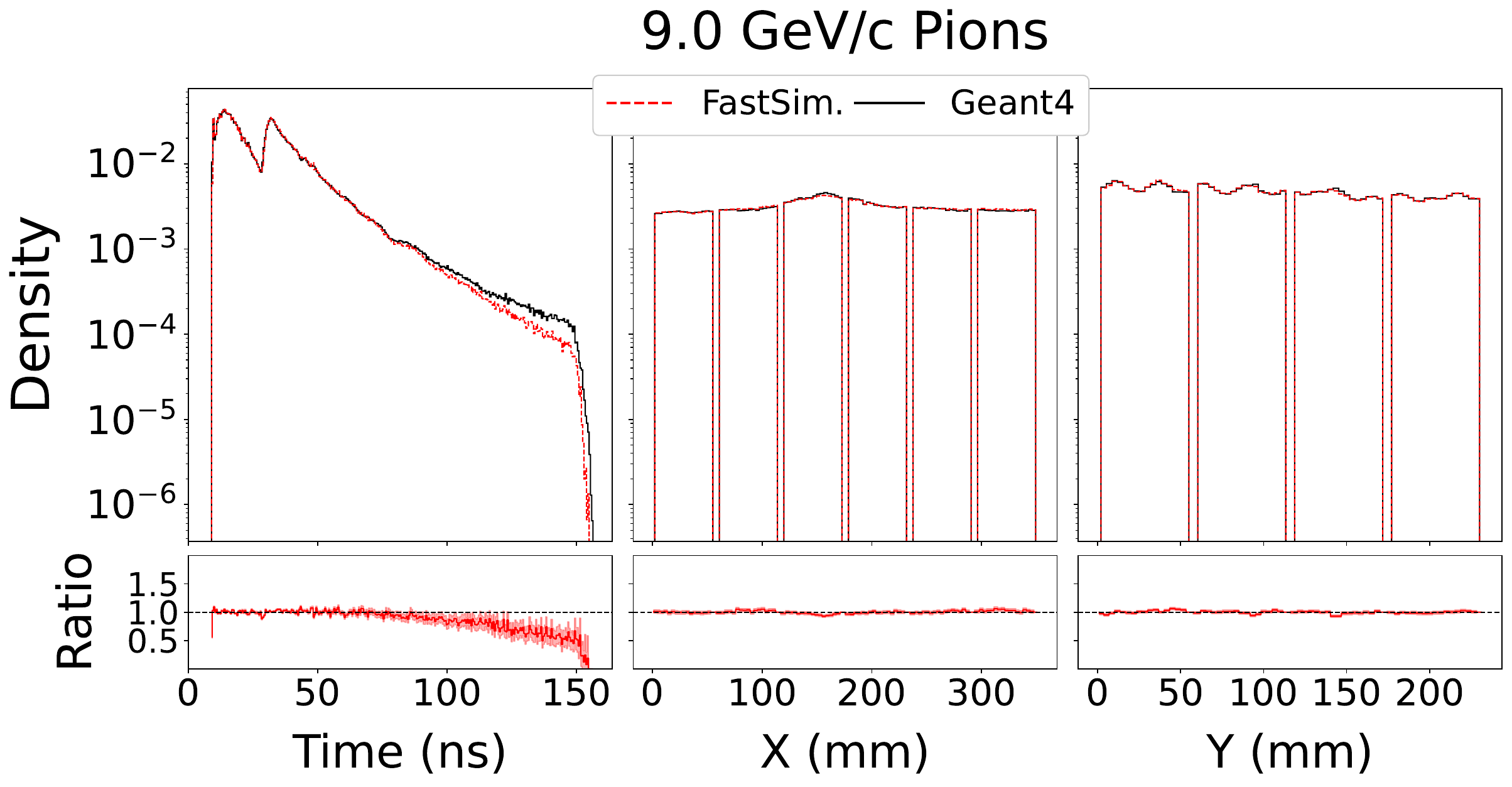}
        \caption{Pions - Two Experts}
    \end{subfigure} \\
    \begin{subfigure}[b]{0.49\textwidth}
        \centering
        \includegraphics[width=\textwidth]{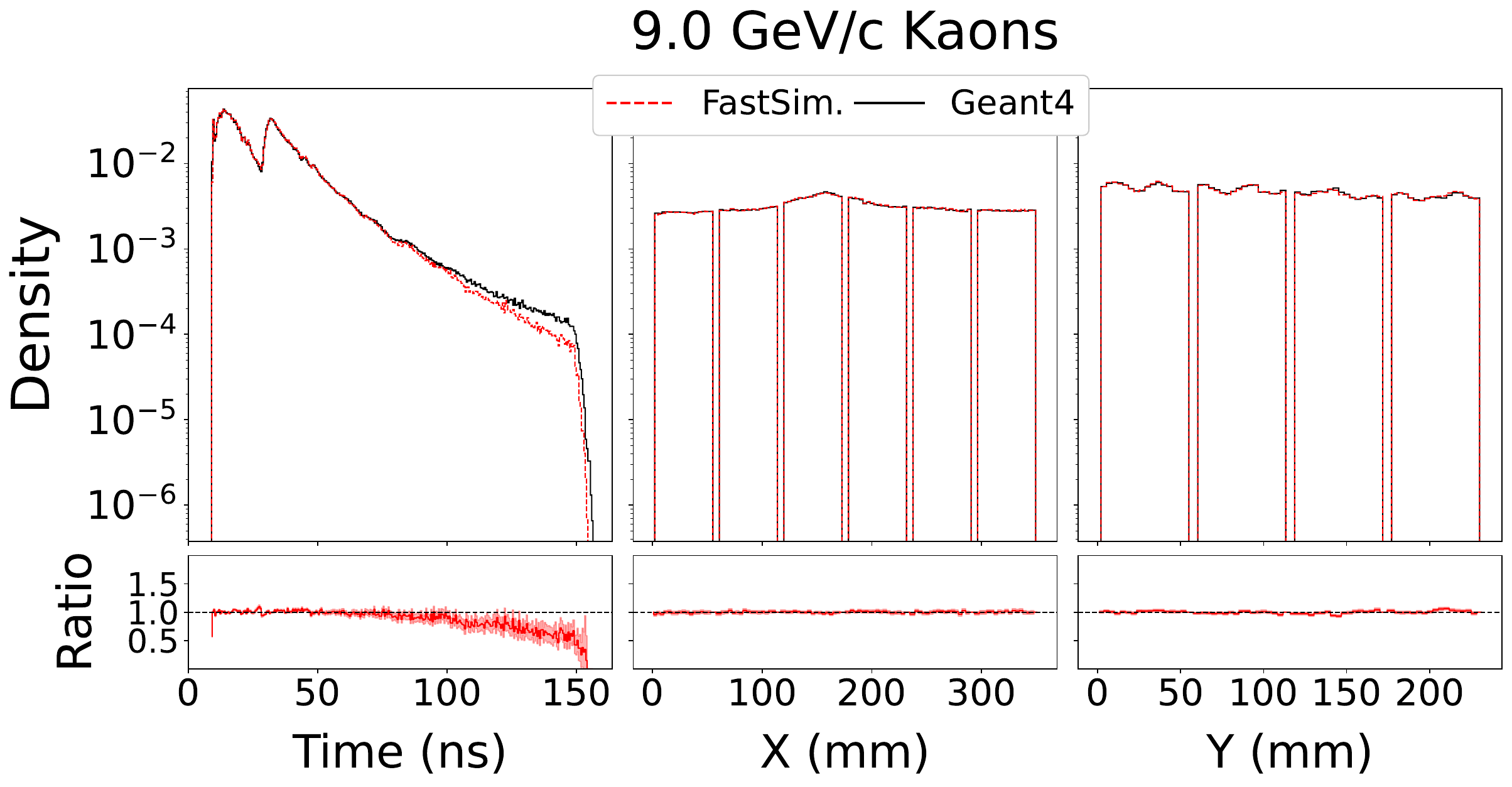}
        \caption{Kaons - Four Experts}
    \end{subfigure} 
    \begin{subfigure}[b]{0.49\textwidth}
        \centering
        \includegraphics[width=\textwidth]{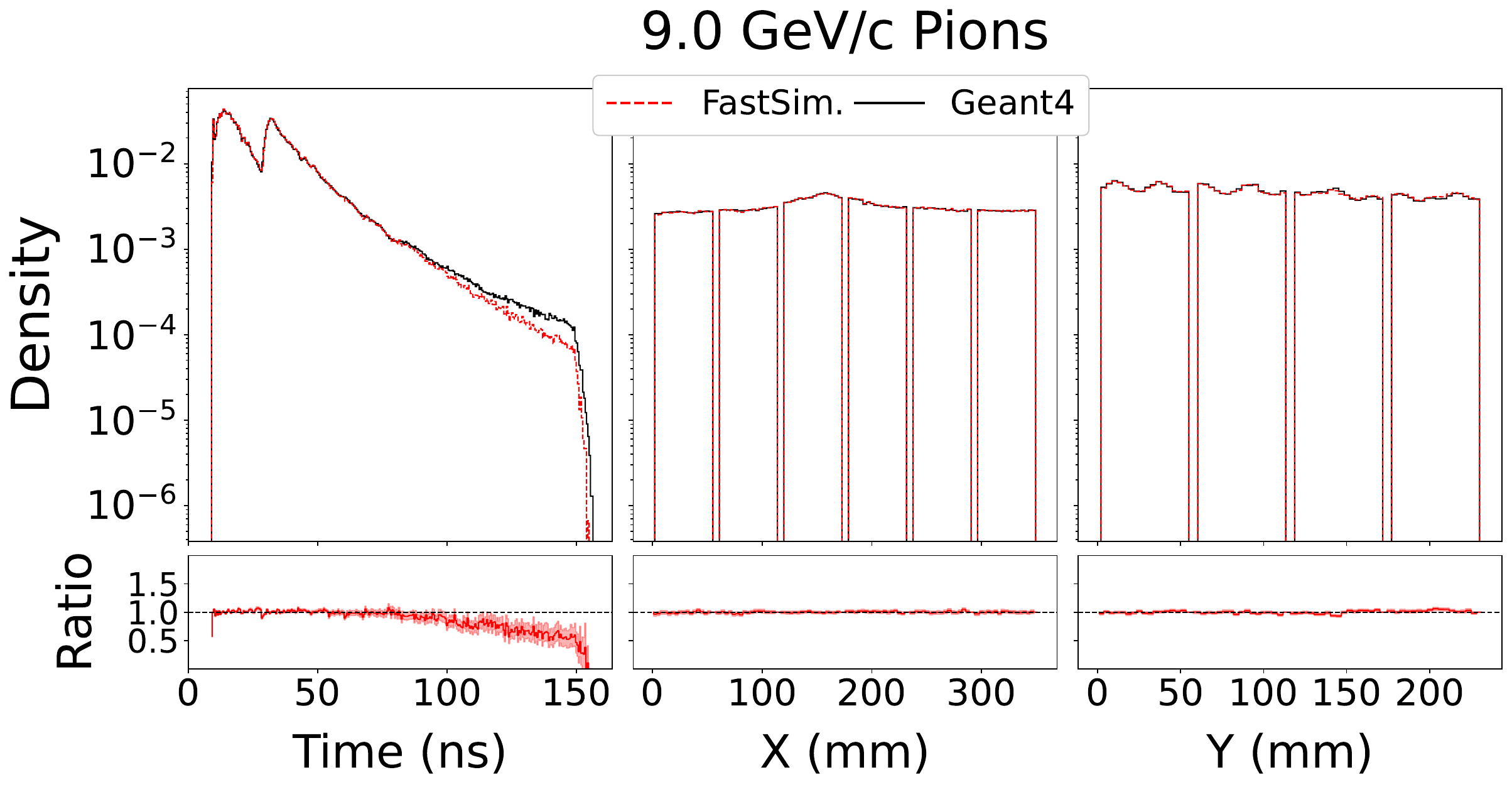}
        \caption{Pions - Four Experts}
    \end{subfigure} 
    \caption{\textbf{Ratio Plots at 9 GeV/c:} Ratio plots for kaons (left column) and pions (right column), using Nucleus sampling and fixed temperature for independent models (top row), a combined model with two experts (middle row) and a combined model with four experts (bottom row).}
    \label{fig:ratios_9GeV}
\end{figure}

\begin{figure}[!]
    \centering
      \begin{subfigure}[b]{0.49\textwidth}
        \centering
        \includegraphics[width=\textwidth]{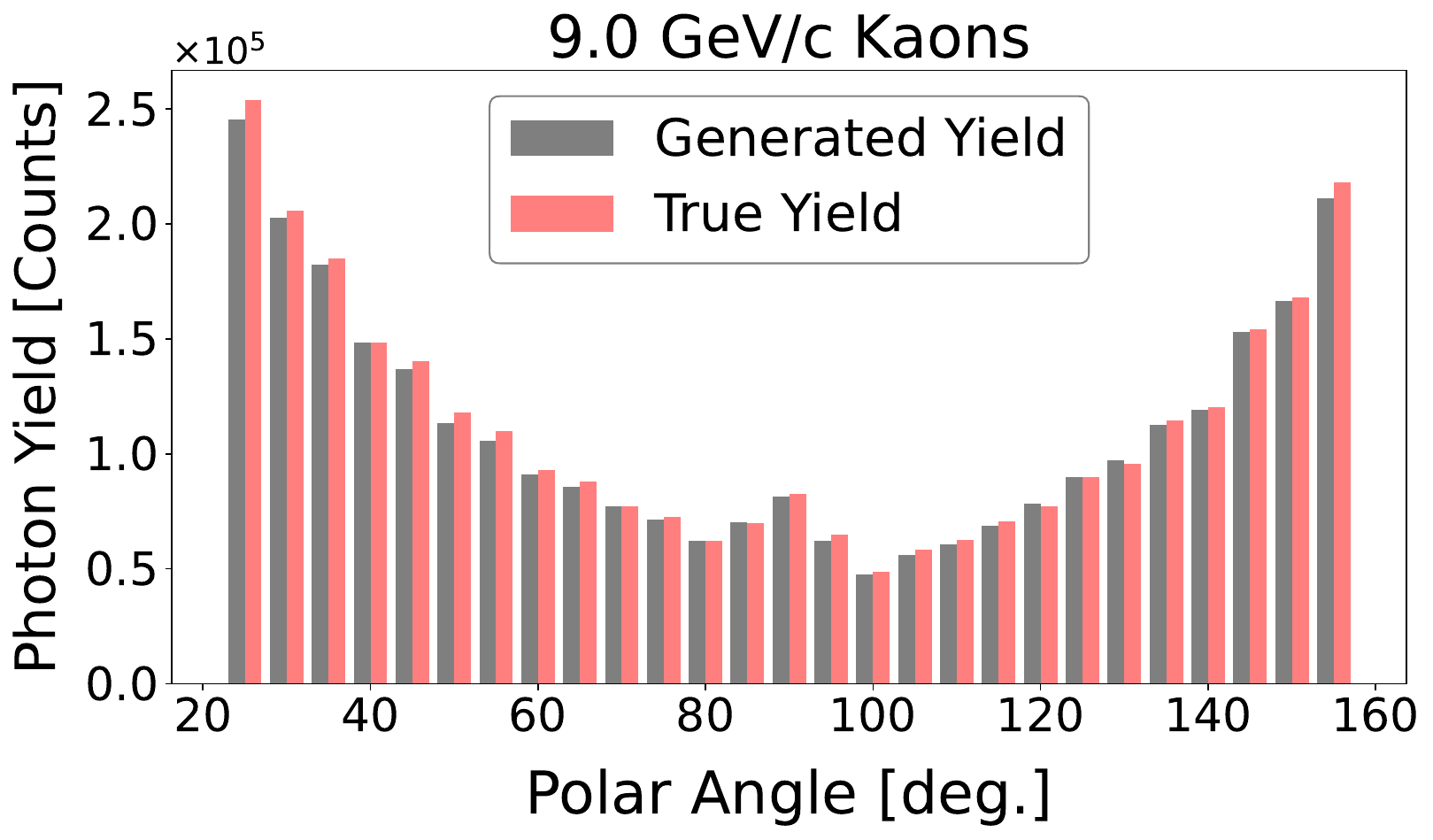}
        \caption{Kaons  - Independent Model}
    \end{subfigure}
    \begin{subfigure}[b]{0.49\textwidth}
        \centering
        \includegraphics[width=\textwidth]{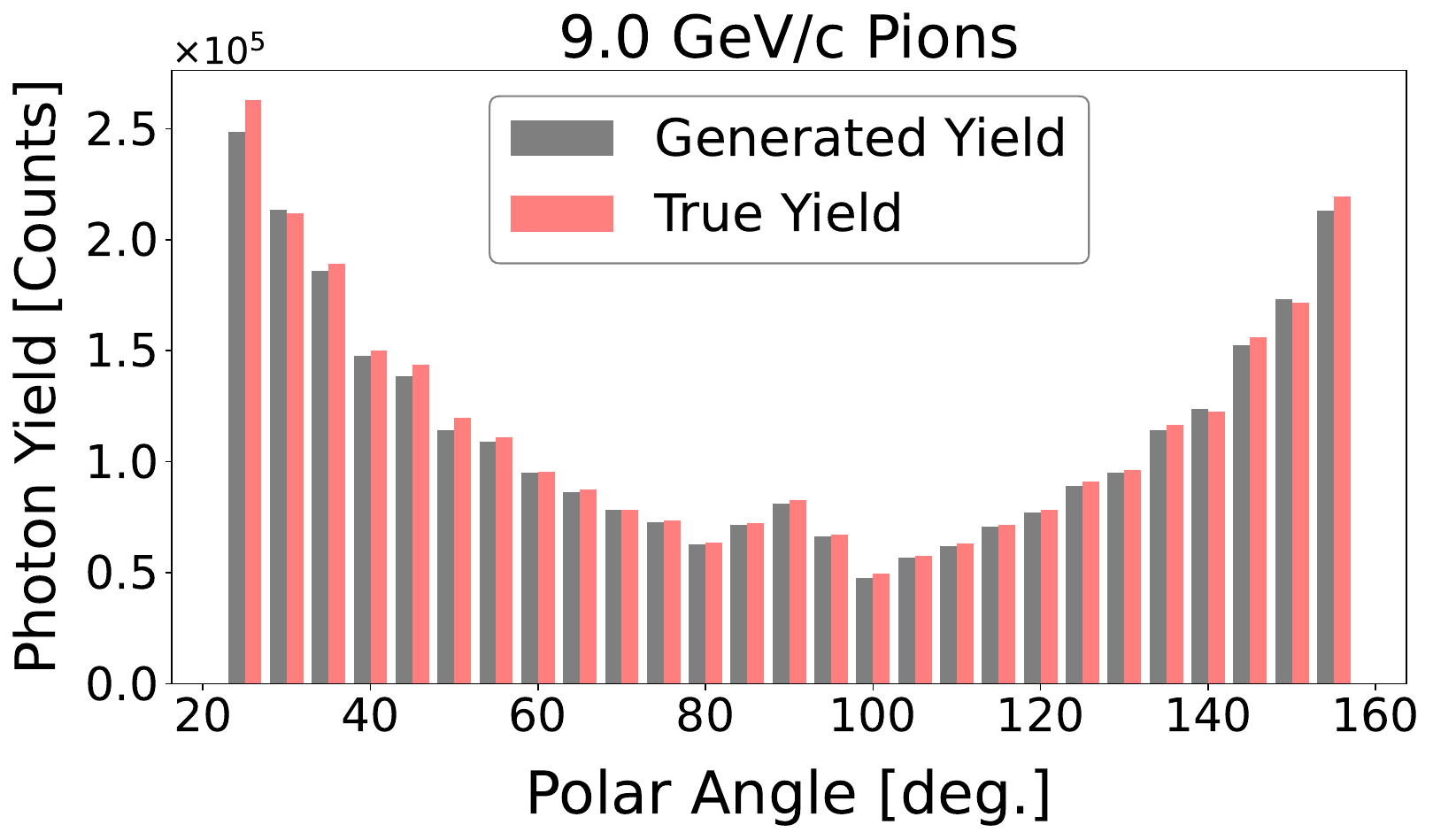}
        \caption{Pions  - Independent Model}
    \end{subfigure} \\
      \begin{subfigure}[b]{0.49\textwidth}
        \centering
        \includegraphics[width=\textwidth]{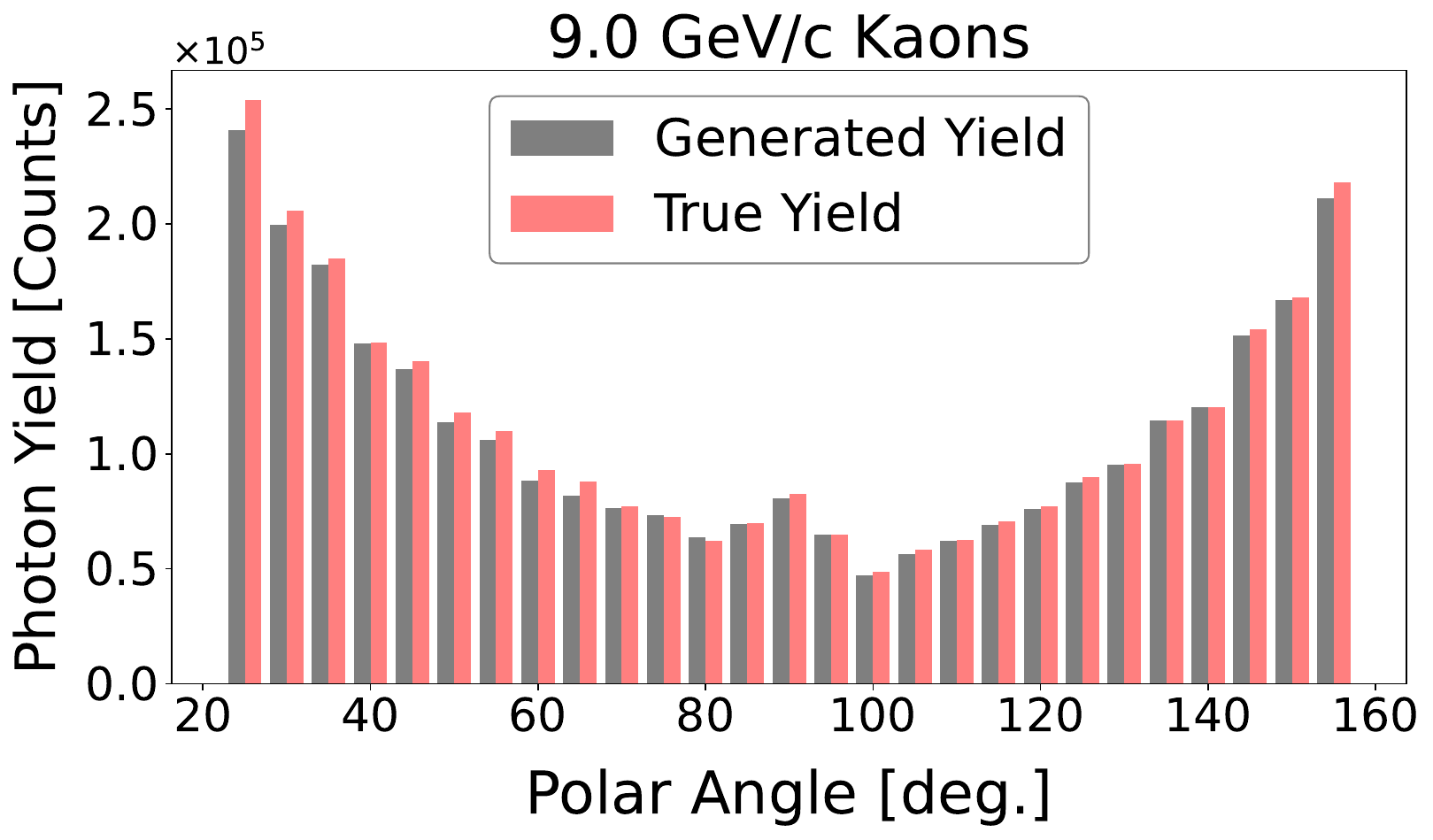}
        \caption{Kaons - Two Experts}
    \end{subfigure}
    \begin{subfigure}[b]{0.49\textwidth}
        \centering
        \includegraphics[width=\textwidth]{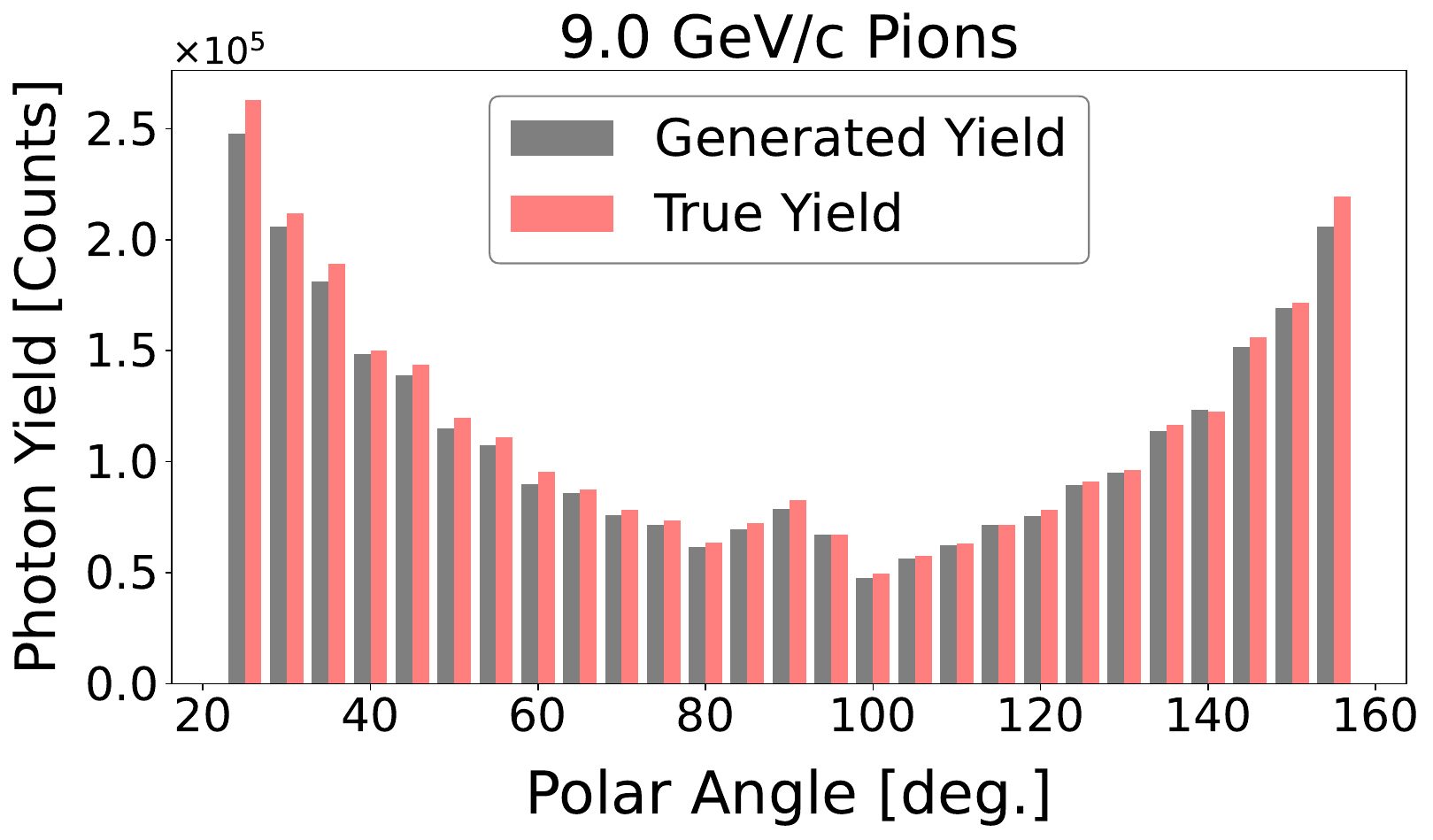}
        \caption{Pions - Two Experts}
    \end{subfigure} \\
      \begin{subfigure}[b]{0.49\textwidth}
        \centering
        \includegraphics[width=\textwidth]{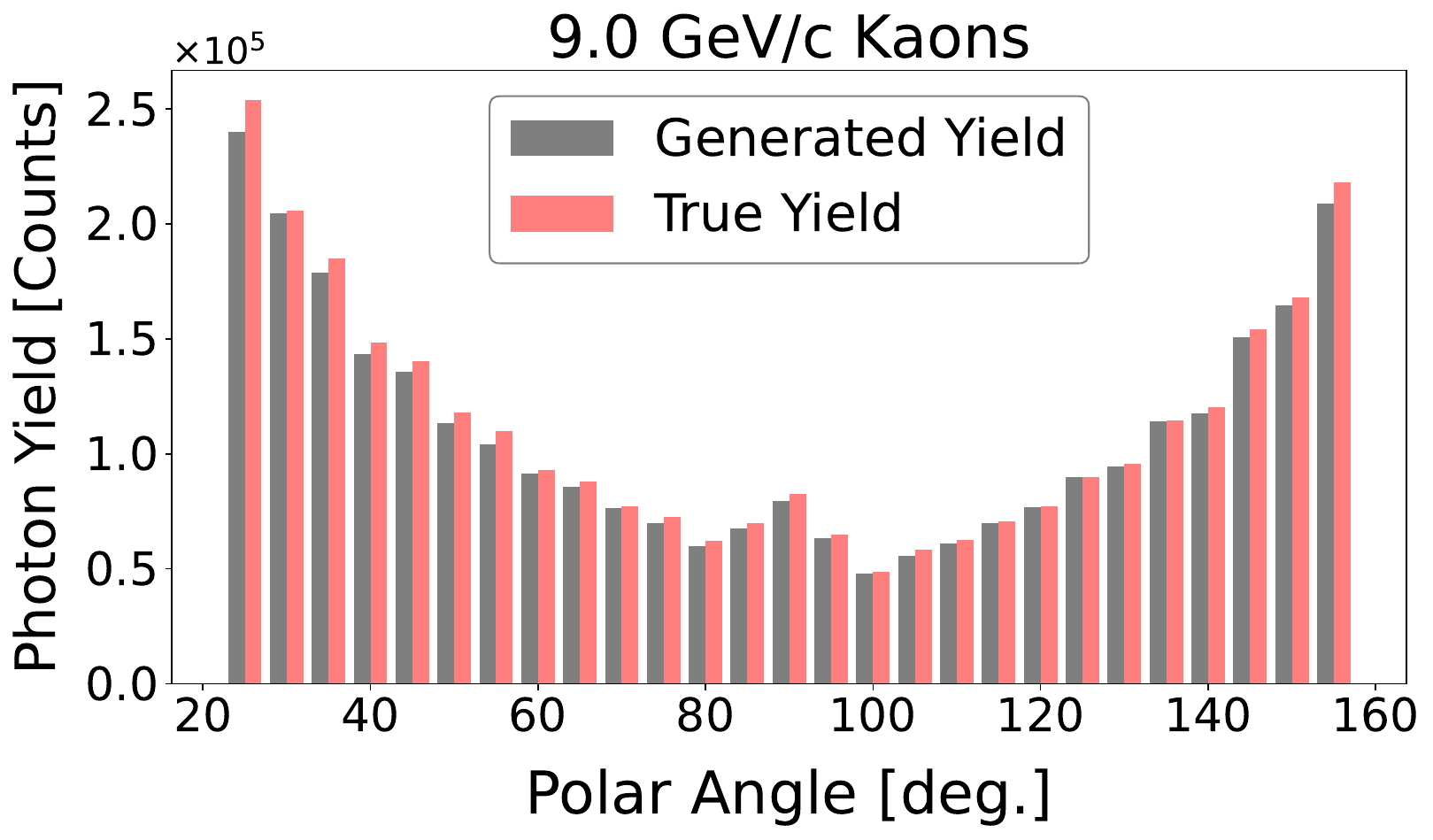}
        \caption{Kaons - Four Experts}
    \end{subfigure}
    \begin{subfigure}[b]{0.49\textwidth}
        \centering
        \includegraphics[width=\textwidth]{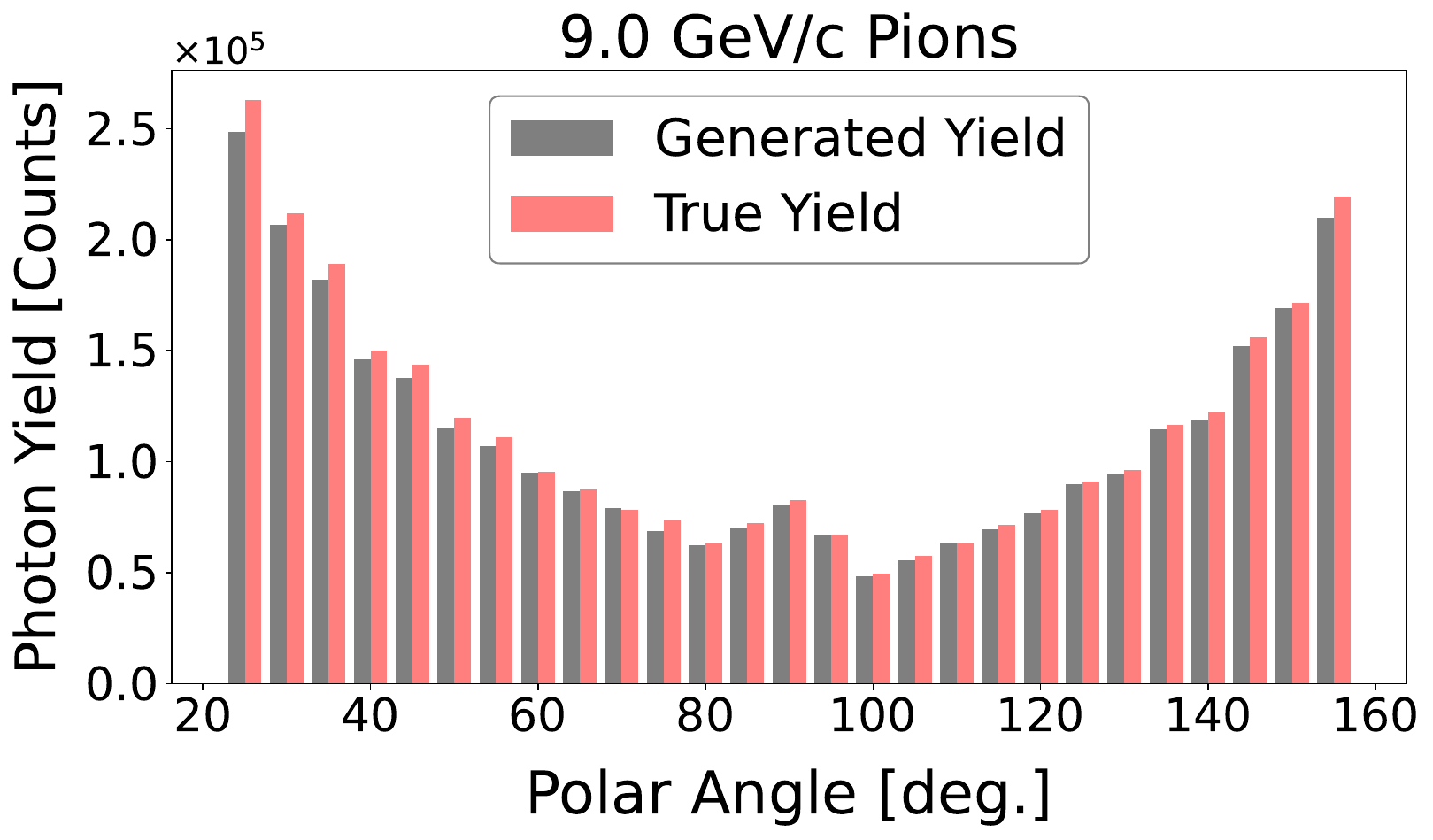}
        \caption{Pions - Four Experts}
    \end{subfigure} 
    \caption{\textbf{Photon Yield Comparison at 9 GeV/c:} Comparison of generated photon yield for kaons (left column) and pions (right column), using Nucleus sampling and fixed temperature for independent models (top row), a combined model with two experts (middle row) and a combined model with four experts (bottom row).}
    \label{fig:yields_9GeV}
\end{figure}

\clearpage
\section{Filtering Evaluation at Additional Kinematics} \label{app:filtering}

\begin{figure}[h]
    \centering
    \begin{subfigure}[b]{0.49\textwidth}
    \includegraphics[width=0.49\textwidth]{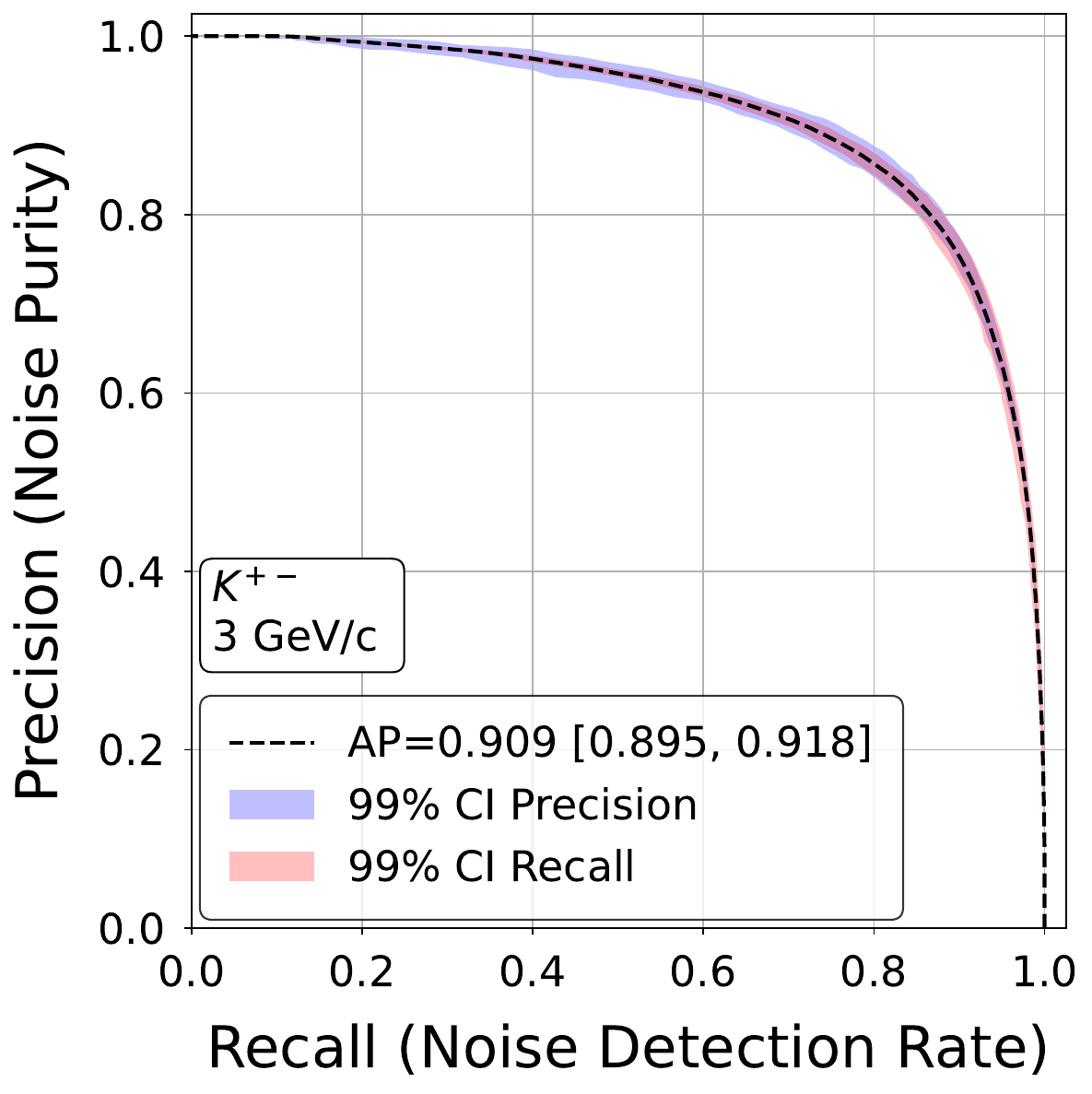}% 
    \includegraphics[width=0.49\textwidth]{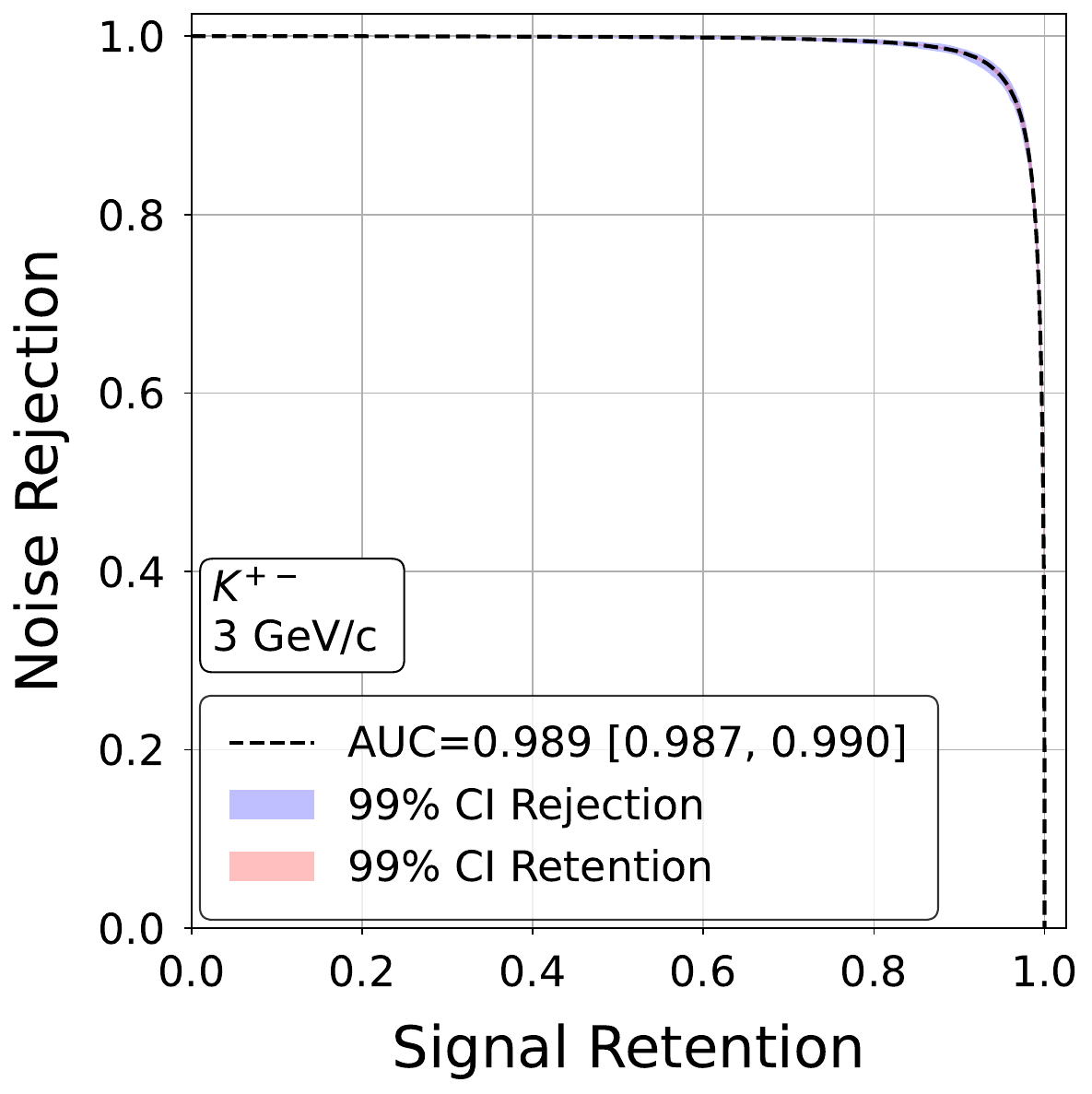} %
    \caption{Kaons}
    \end{subfigure} 
    \begin{subfigure}[b]{0.49\textwidth}
    \includegraphics[width=0.49\textwidth]{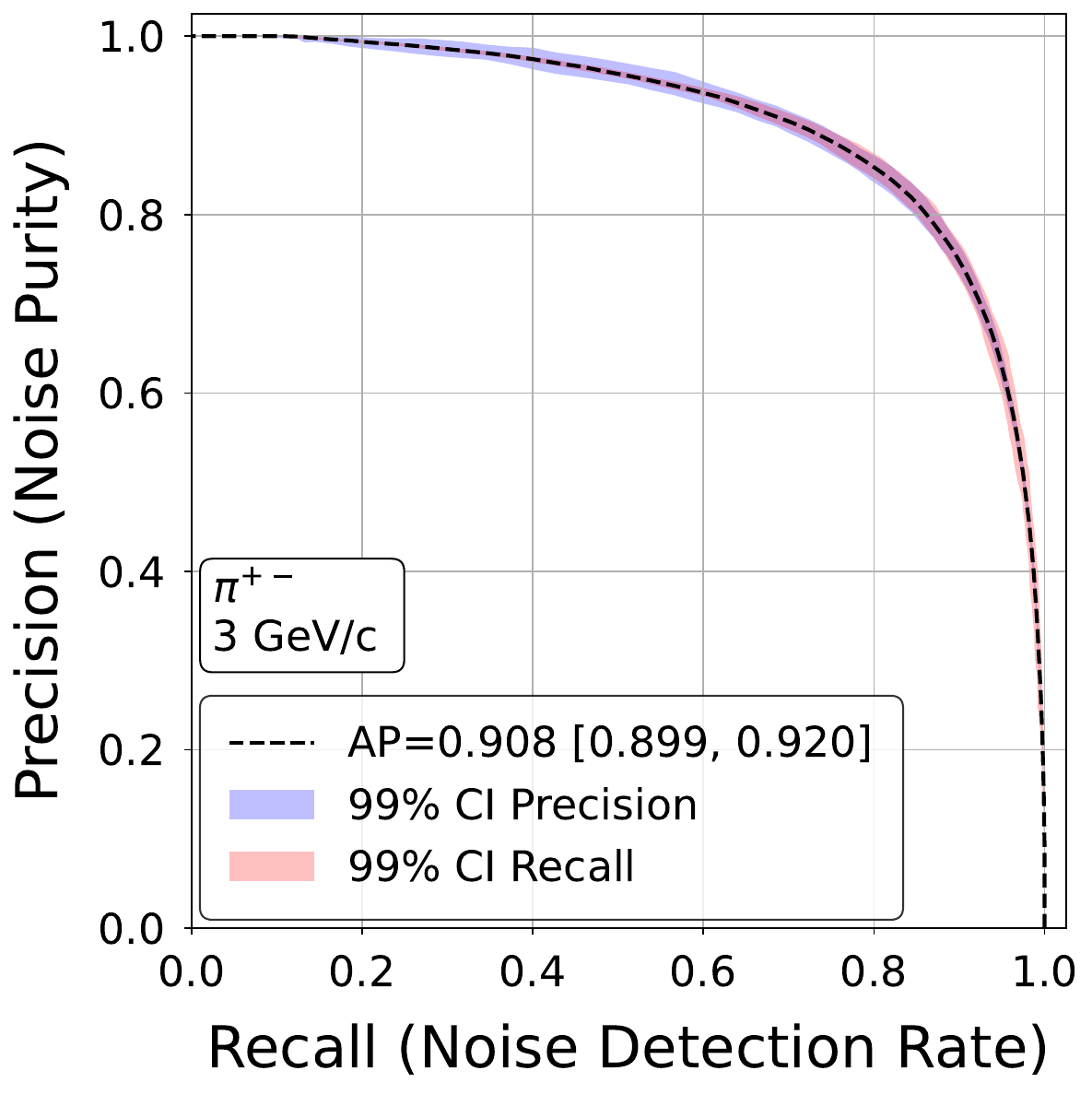}% \\
    \includegraphics[width=0.49\textwidth]{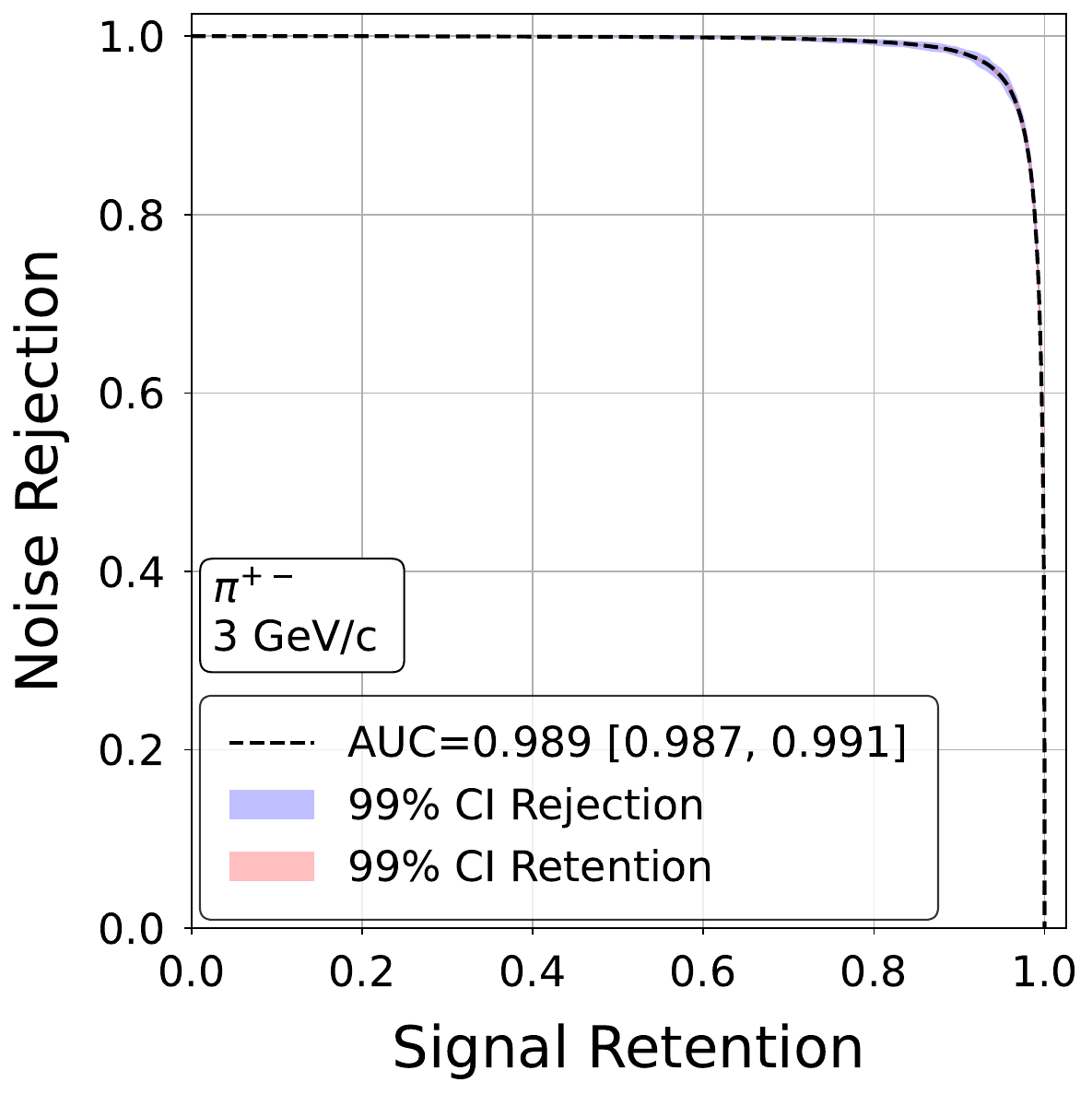} 
    \caption{Pions}
    \end{subfigure}
    \caption{
    \textbf{Noise Filtering at 3 GeV/c:} Noise filtering performance of our method at 3~GeV for (a) kaons and (b) pions, shown as precision-recall curves and noise rejection versus signal efficiency. The precision-recall curves are insensitive to class imbalance. The dark rate contributes approximately $8\text{--}10\%$ of total hits as noise.}
    \label{fig:Filtering_3GeV}
\end{figure}

\begin{figure}[h]
    \centering
    \begin{subfigure}[b]{0.49\textwidth}
    \includegraphics[width=0.49\textwidth]{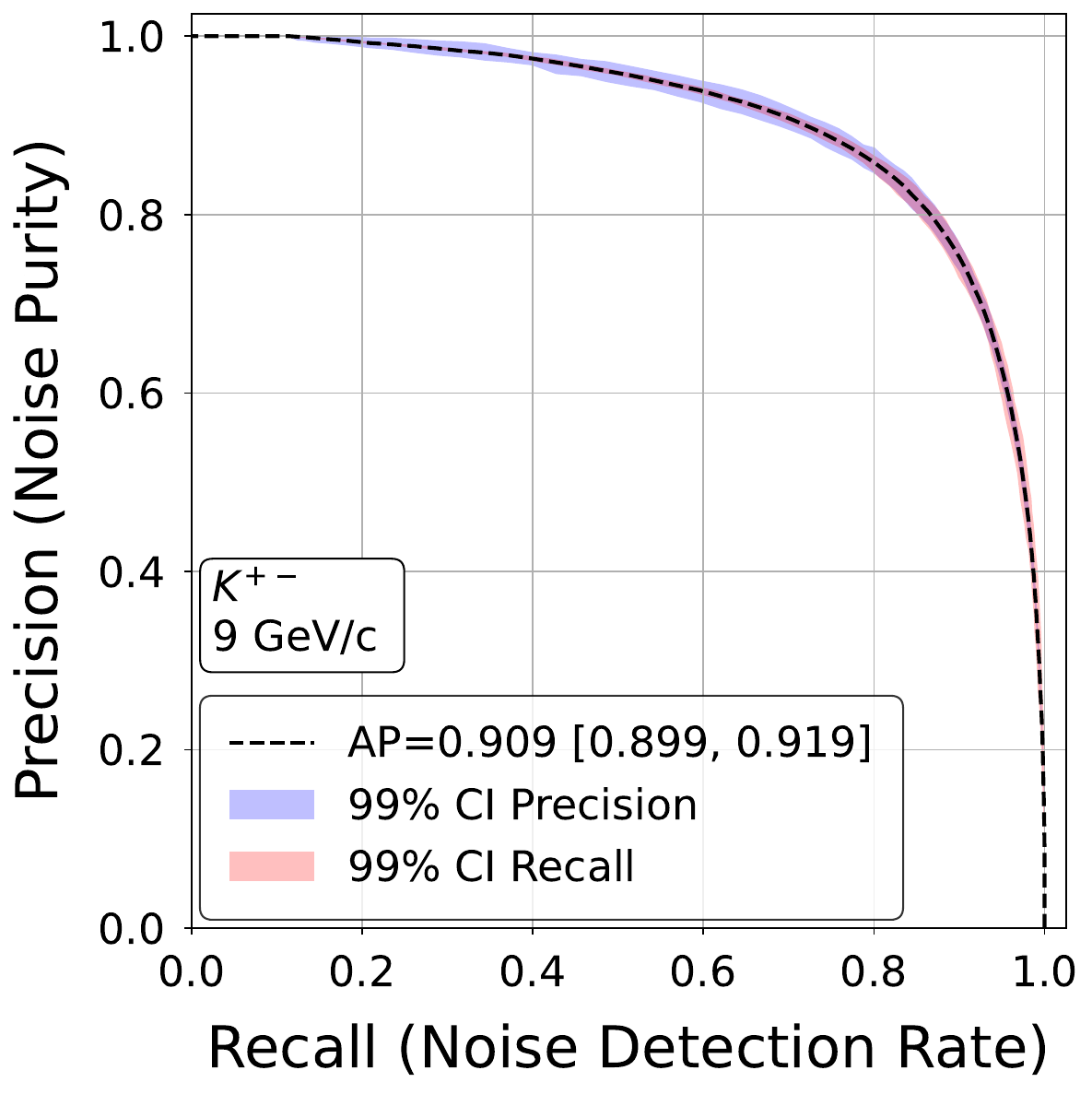}% 
    \includegraphics[width=0.49\textwidth]{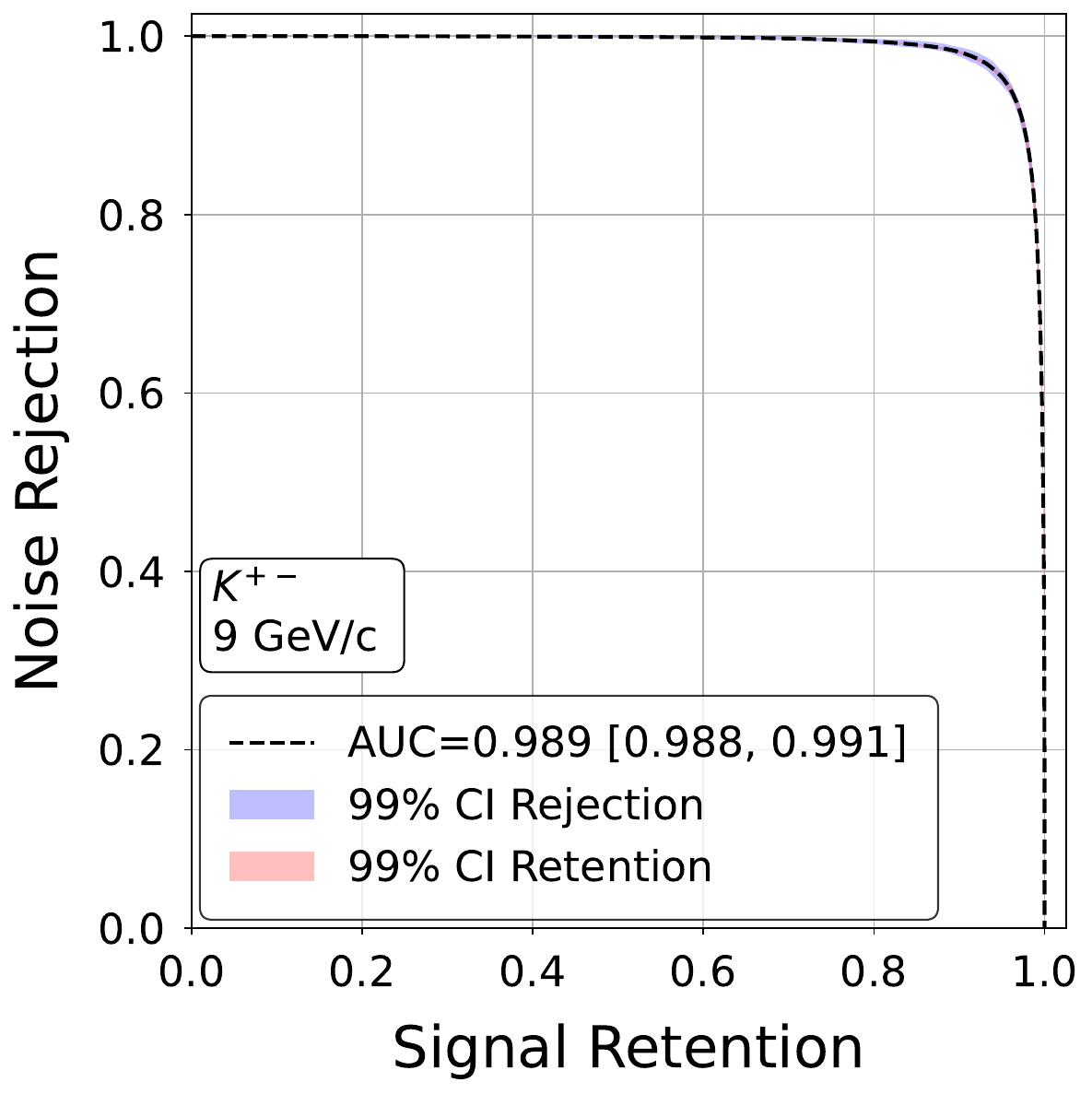} %
    \caption{Kaons}
    \end{subfigure} 
    \begin{subfigure}[b]{0.49\textwidth}
    \includegraphics[width=0.49\textwidth]{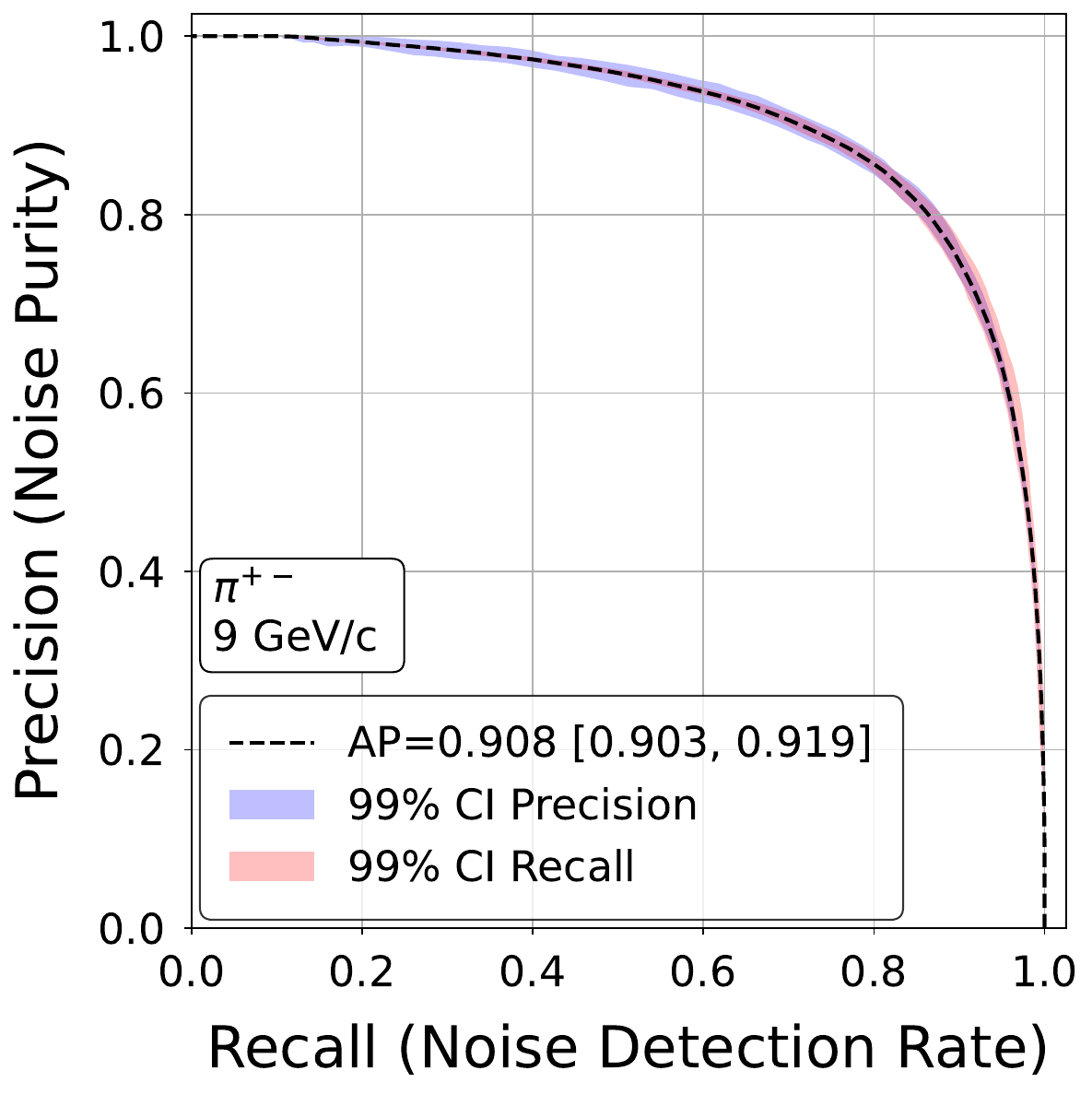}% \\
    \includegraphics[width=0.49\textwidth]{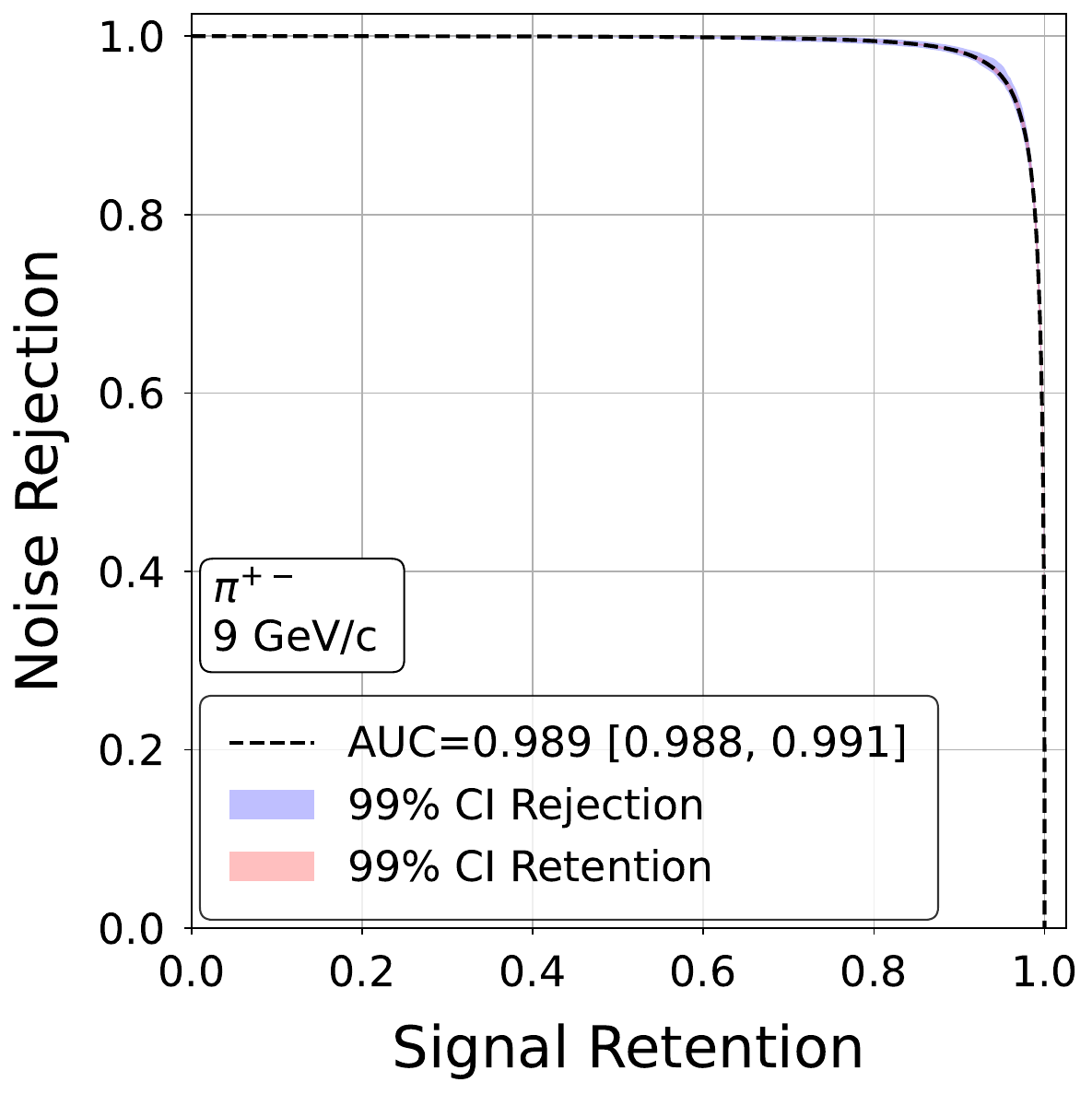} 
    \caption{Pions}
    \end{subfigure}
    \caption{
    \textbf{Noise Filtering at 9 GeV/c:} Noise filtering performance of our method at 9~GeV for (a) kaons and (b) pions, shown as precision-recall curves and noise rejection versus signal efficiency. The precision-recall curves are insensitive to class imbalance. The dark rate contributes approximately $8\text{--}10\%$ of total hits as noise.}
    \label{fig:Filtering_9GeV}
\end{figure}

\end{document}